\documentclass{article}

\usepackage[preprint]{neurips_2024}

\usepackage{amsmath,amsfonts,bm}
\def\eqref#1{equation~\ref{#1}}
\def\1{\bm{1}}

\def\vtheta{{\bm{\theta}}}

\def\vk{{\bm{k}}}

\def\vm{{\bm{m}}}

\def\vp{{\bm{p}}}

\def\vs{{\bm{s}}}

\def\vv{{\bm{v}}}

\def\vx{{\bm{x}}}

\def\mP{{\bm{P}}}

\def\mW{{\bm{W}}}
\def\mX{{\bm{X}}}

\DeclareMathAlphabet{\mathsfit}{\encodingdefault}{\sfdefault}{m}{sl}
\SetMathAlphabet{\mathsfit}{bold}{\encodingdefault}{\sfdefault}{bx}{n}

\def\gL{{\mathcal{L}}}

\def\gS{{\mathcal{S}}}

\def\gV{{\mathcal{V}}}

\def\sR{{\mathbb{R}}}

\usepackage{graphicx}
\usepackage{arydshln}  % 用于虚线
\usepackage{subcaption}
\usepackage{wrapfig}    % 让图片环绕文本

\usepackage{amsmath}
\usepackage{amssymb}
\usepackage{mathtools}
\usepackage{amsthm}

\usepackage{tabularx} % 在导言区加载tabularx包
\usepackage{array} % 加载array包
\usepackage{multirow}
\usepackage{makecell}
\usepackage{colortbl}  % 引入xcolor宏包，用于设置表格的颜色
\usepackage{subcaption}
\theoremstyle{plain}

\theoremstyle{definition}

\theoremstyle{remark}

\usepackage[textsize=tiny]{todonotes}

\usepackage[utf8]{inputenc} % allow utf-8 input
\usepackage[T1]{fontenc}    % use 8-bit T1 fonts
\usepackage{hyperref}       % hyperlinks
\usepackage{url}            % simple URL typesetting
\usepackage{booktabs}       % professional-quality tables
\usepackage{amsfonts}       % blackboard math symbols
\usepackage{nicefrac}       % compact symbols for 1/2, etc.
\usepackage{microtype}      % microtypography
\usepackage{xcolor}         % colors
\usepackage{comment}

\usepackage{mdframed} % 边框与背景设置
\usepackage[most]{tcolorbox}  % 导入 tcolorbox
\usepackage{xcolor}           % 颜色支持
\tcbset{
	observation/.style={
		colback=yellow!10,    % 背景色：20% 的黄色
		colframe=yellow!50!black, % 边框颜色
		boxrule=0pt,        % 边框粗细
		arc=2mm,              % 圆角大小
		left=4mm,             % 左内边距
		right=4mm,            % 右内边距
		top=1mm,              % 上内边距
		bottom=1mm,           % 下内边距
		before skip=6pt,      % 调整上方间距
		after skip=6pt,       % 调整下方间距
		enhanced              % 开启增强功能（必须）
	}
}

\newcommand{\ptf}{\vp_{\rm tf}}
\newcommand{\prnn}{\vp_{\rm rnn}}
\newcommand{\ptgt}{\vp_{\rm tgt}}
\newcommand{\ptheta}{\vp_{\vtheta}}
\newcommand{\htf}{\mathrm{H}(\vp_{\rm tf})}
\newcommand{\hrnn}{\mathrm{H}(\vp_{\rm rnn})}
\newcommand{\htgt}{\mathrm{H}(\vp_{\rm tgt})}
\newcommand{\htheta}{\mathrm{H}(\vp_{\vtheta})}
\newcommand{\kltf}{\mathrm{KL}(\vp_{\rm tgt} | \vp_{\rm tf})}
\newcommand{\klrnn}{\mathrm{KL}(\vp_{\rm tgt} | \vp_{\rm rnn})}

\newcommand{\kltheta}{\mathrm{KL}(\vp_{\rm tgt} | \vp_{\vtheta})}
\newcommand{\vDelta}{\mathbf{\Delta}}

\title{Revisiting Transformers through the Lens of \\ Low Entropy and Dynamic Sparsity}

\author{%
  Ruifeng Ren \\
	Gaoling School of Artificial Intelligence\\
	Renmin University of China\\
  	Beijing, China\\
  \texttt{renruifeng920@ruc.edu.cn} \\
  \And
  Yong~Liu\thanks{Corresponding Author} \\
  Gaoling School of Artificial Intelligence\\
  Renmin University of China\\
  Beijing, China\\
  \texttt{liuyonggsai@ruc.edu.cn} \\
}

\begin{document}

\maketitle

\begin{abstract}
	Compression has been a critical lens to understand the success of Transformers. 
	In the past, we have typically taken the target distribution as a criterion to evaluate a model’s compression performance.
	Nevertheless, it often remains challenging to precisely assess how well the model achieves compression and to compare the information content of the learned distribution with that of the target distribution during compression, as the target distribution is typically unknown and entropy computation often incurs exponential cost.
	In this work, we explore these issues under a controlled experimental setup. 
	We find that Transformers exhibit a unique inductive bias in data compression: beyond approaching the target distribution, they tend to favor learning lower-entropy distributions, with this tendency becoming more pronounced as the model size increases.
	This preference prevents Transformers from perfectly aligning with the target distribution, instead further compressing its information content.
	Furthermore, we show that the FFN module plays a critical role in driving this bias.
	In addition, while models remove informational redundancy from data during compression, they also exhibit redundancy within their parameters, which enables compression and can be characterized through dynamic sparsity.
	However, the dynamic sparsity patterns in Transformers, particularly in attention and FFN modules, demand further exploration.
	As for this, we show that larger Transformers show stronger preferences for bypassing attention computations via residual connections and have lower proportion of active neurons. 
	Interestingly, we also find that training instability in larger models strongly correlates with sudden increases in dead neurons.
	Our work contributes to a deeper understanding of Transformers from the lens of entropy and dynamic sparsity.
\end{abstract}

%	These issues remain challenging due to the typically unknown true distributions and the exponential computational complexity of model's learned distribution.

\section{Introduction}
Recently, compression has been a popular perspective for understanding the success of Transformer models~\citep{Ilyatalk2023, LMisCompression, CompressionIntelligenceLinearly, pan2025understanding}.
Depending on the compression objectives, it involves data compression and model compression.
Data compression focuses on removing redundant information from data.
From this perspective, Transformer can compress the large-scale data and model the underlying target distribution $\ptgt$ with significantly fewer parameters $\vtheta$.
\citet{LMisCompression} propose that there is a strong correlation between language modeling and compression.
During training, Transformer learns a parameterized distribution $\vp_{\vtheta}$ to maximize the log-likelihood, which is equivalent to minimizing the expected code length when using the model for compression.
According to Shannon's source coding theorem~\citep{shannon}, the minimum expected number of bits required to encode the data is precisely the entropy $\htgt$, which also serves as the limit of the model's compression performance.
Moreover, it is believed that compression can represent intelligence to some extent \citep{hutter2005universal, Ilyatalk2023, pan2025understanding}.
As for this, \citet{CompressionIntelligenceLinearly} show a linear relationship between the model's performance on downstream tasks and its compression efficiency.

In the past, we have typically taken the target distribution as a criterion and use metrics
such as cross-entropy loss to evaluate the performance of language modeling.
Nevertheless, it still remains difficult to say exactly how well the model achieves compression as the real target distribution is often unknown.
In addition, although theoretically the information content of the learned $\ptheta$ should be consistent with that of $\ptgt$ (i.e., $\htheta = \htgt$), it is also unknown whether the model can achieve this optimal scenario and what the relationship is between the learned $\htheta$ and the true $\htgt$ during the process of approaching $\ptgt$.
These issues are also difficult to explore in real-world scenarios. 
Apart from $\ptgt$ or its entropy being unseen, calculating $\htheta$ itself is also challenging: to compute the distribution of a sequence of length $n$, we typically need to compute $|\gV|^n$ discrete probabilities where $|\gV|$ is the vocabulary size and often reaches 60K or more in LLMs, resulting the impractical exponential computational cost to get $\htheta$.

Therefore, in this work, we turn our focus on a simpler and more controlled setting where we predefine a target distribution to generate data and train Transformers to investigate the aforementioned questions.
Interestingly, we find that Transformer exhibits its unique inductive bias in data compression: beyond aligning with the target distribution, it also show a preference for learning distributions with low entropy and this tendency will be stronger in larger Transformers (see Section~\ref{sec:low-entropy}).
Due to this inherent preference, Transformer often fails to perfectly match the target distribution (larger $\kltheta$) and instead learn $\ptheta$ whose entropy can be even lower than the true $\htgt$.
Furthermore, we investigate the roles of the attention module and the FFN module in this process and find that it is the FFN module that drives this preference of Transformers (see Section~\ref{sec:entropy-FFN}).

In the process of compressing data by eliminating informational redundancy within the data, models themselves also exhibit redundancy in their parameters providing a promising mechanism for model compression.
This redundancy in model parameters can be characterized through dynamic sparsity \citep{tutorial_dynamic_sparse}.
As two key components, the attention and FFN modules are essential for exploring dynamic sparsity in Transformers.
As for the attention module, previous works mainly focus on the dynamic sparsity of attention computation, where the outputs often depend only on limited key tokens \citep{sparse-TF, ent15max, adaptive_entmax}.
Instead of delving into this, inspired by \citet{attention_rank_collapse}, we treat each residual connection or each attention head as a path in the forward process and focus more on the sparsity of path selection in the attention module.
We find that after introducing a routing mechanism for these paths in the attention module, larger Transformers tend to favor residual connections to bypass attention computations (see Section~\ref{sec:attention}).

On the other hand, as for the FFN module, it has been shown that only a small subset of neurons in modern LLMs is selectively activated for different inputs, exhibiting a dynamic sparsity pattern \citet{neurons_dead}.
However, although we can collect large-scale corpora as inputs, it is still difficult to determine whether this pattern exists for all potentially unseen inputs. 
Moreover, the formation process of this sparsity pattern also requires further exploration.
By exhaustively traversing all possible inputs under our controlled setting, we find that this sparsity pattern is pervasive and larger Transformers will have a lower proportion of active neurons, including some never activated, which is consistent with findings in real scenarios \citep{neurons_dead}. 
Surprisingly, we also discover that this sparsity pattern enhances in a jump-like manner during training, and the loss spikes in larger models is highly correlated with sudden decreases in neuronal activity (see Section~\ref{sec:FFN}).

\section{Preliminary}\label{sec:pre}
The vector $\vs$ is used to represent a string (with a default length of $n$), where $\vs_i$ denotes the $i$-th character and $\vs_{<i}$ denotes the prefix of $\vs_i$.
We use the vector $\vp$ to represent a distribution and use the scalar \( p \) when emphasizing the probability of a specific string.
In addition, we use $\vDelta^{d}$ to denote the set of $(d-1)$-simplices, where all $d$ elements of a vector $\vp \in \sR^{d}$ sum to 1 if $\vp \in \vDelta^{d}$.

\begin{figure}[t]
	\centering
	\includegraphics[scale=0.60]{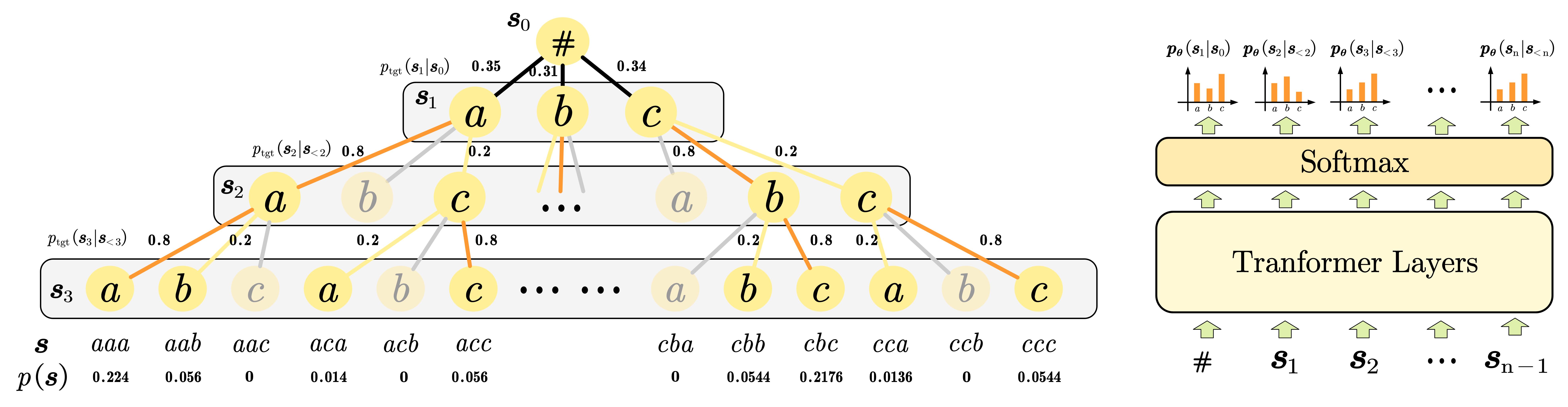}
	\caption{{\bf Left:} An example of generating a target distribution $\ptheta$ when $|\gV| = 3$ and length $n = 3$. We choose the vocabulary $V = \{a, b, c\}$.  
		We produce conditional probabilities to generate the final $p(\vs)$. 
		We introduce "\#"  as the the start symbol $\vs_0$ and we define $p(\vs_0) = p$(\#) = 1.
		First, $\vp(\vs_1 |\vs_0)$ is generated from a uniform distribution. Then, when generating the subsequent $ \vp(\vs_i | \vs_{<i}) $ for $ i\geq 2$, we randomly select two characters and assign them transition probabilities of 0.8 and 0.2, while the remaining characters have a probability of 0.
		{\bf Right:} 
		The sequence $\vs_{<i} = \vs_0\vs_1\dots\vs_{i-1}$ is fed into the model to generate $\vp(s_i | s_{<i})$ 
		and $p_{\vtheta}(\vs)$ can be calculated as $p_\vtheta(\vs) = \prod_{i=1}^{n} p_{\vtheta}(\vs_i|\vs_{<i})$.
		%where $\vs_0 = $"\#" serves as a trigger, allowing the model to output $\vp_{\vtheta}(\vs_1) = \vp_{\vtheta}(\vs_1|\vs_0)$.
	}
	\label{fig:setting}
\end{figure}

{\bf Language Modeling:}
A sequence $\vs = [\vs_i]_{i=1}^{n}$ of length $n$ is formed by $n$ concatenating characters from a vocabulary $\gV$, where we use $|\gV|$ to denote the size of $\gV$ and $\vs_i$ represents the $i$-th character.
We use $\gS_n$ to present the set of all sequences of length $n$, and the size of $\gS_n$ is denoted as $|\gS_n| = |\gV|^n$.
Then, the target distribution over the entire set $\gS_n$ can be denoted as $\ptgt(\gS) \in \vDelta^{|\gS_n|}$ where each element $p_{\rm tgt}(\vs)$ represents the probability of a specific sequence $\vs$.
We introduce an additional character "\#"  as the beginning of the sequence ($\vs_0$ is "\#") and we define $p(\vs_0) = p$(\#) = 1.
The probability $p_{\rm tgt}(\vs): \gS_n \to \sR$ can be expressed through $n$ conditional probabilities, that is, $p_{\rm tgt}(\vs) = \prod_{i=1}^{n} p_{\rm tgt}(\vs_i|\vs_{<i})$.
We also simply denote $\ptgt(\gS)$ as $\ptgt$. 

The aim of a language model is to learn a parameterized distribution $\ptheta$ and make it as close as possible to the target distribution $\ptgt$, which is typically achieved through auto-regressive loss, that is, $\gL(\vtheta) = \mathbb{E}_{\vs \sim \ptgt}\left[- \frac{1}{N} \sum_{i=1}^{n} \log p_{\vtheta} (\vs_i | \vs_{<i}) \right]$ where $\vs_{<i}$ represents the prefix of $\vs_i$.
The entropy of the learned $\ptgt$ can be respectively formulated as $\mathrm{H}(\ptheta) = - \sum_{\vs \in \gS_n} p_{\vtheta}(\vs) \log p_{\vtheta}(\vs)$ where $p_\vtheta(\vs) = \prod_{i=1}^{n} p_{\vtheta}(\vs_i|\vs_{<i})$.
The KL divergence between \( p_1 \) and the target distribution can also be computed as $\mathrm{KL}(\ptgt | \ptheta) = \sum_{\vs \in \gS_n} p_{\rm tgt}(\vs) \log[p_{\rm tgt}(\vs) / p_{\vtheta}(\vs)]$.
% [\log p_{\rm tgt}(\vs) - \log p_{\vtheta}(\vs)] = - \htgt + \chtheta $.

{\bf Controlled Experiments:} 
To obtain a known $\ptgt$, we fix the vocabulary size $|\gV|$ and sequence length $n$, and then generate the conditional distributions $\vp(\vs_i | \vs_{<i}) \in \vDelta^{|\gV|}$ through random sampling.
Specifically, we first sample $\vp(\vs_1|\vs_0)$ from a uniform distribution as the beginning and then follow the sparsity principle to generate subsequent $\vp(\vs_i|\vs_{<i})$, meaning that only a few characters are assigned high transition probabilities while the remaining ones are assigned zeros.
This sparsity principle is intuitive: in natural language, when the prefix is fixed, only a few words from the vocabulary are likely to follow. 
For example, adjectives are often followed by nouns, rather than verbs.
The final $\ptgt$ can be obtained through these generated conditional distributions, as illustrated in Figure \ref{fig:setting}.
We can alter $\htgt$ by controlling the degree of sparsity.

Since the generated $\ptgt$ is known, we can directly compute $\htgt$ and $\kltheta$. 
Furthermore, when generating the conditional probabilities, we randomly select the transition probabilities of two characters to be 0.8 and 0.2, with the remaining characters having a probability of 0.
More details of our settings can be seen in Appendix~\ref{app:base setting}.
In addition to generating distributions as described above, we also explore cases with different distributions and the results can be found in Appendix~\ref{app:more-dist}.
In the following, instead of using $\ptheta$ directly, we also use $\ptf$ and $\prnn$ to represent the distributions learned by Transformer and RNNs respectively.

\section{Transformers Prefer to Wander within Low-entropy Landscapes}

Transformer has been recognized as a powerful compressor \citep{LMisCompression, CompressionIntelligenceLinearly}. 
Traditionally, the target distribution has been used as a criterion with metrics like cross-entropy loss employed to evaluate language modeling performance. However, it remains challenging to precisely assess how well the model achieves compression as the true target distribution is often unknown. Furthermore, it also remains unclear whether the model can attain the optimal scenario (i.e., $\htheta = \htgt$) and what the relationship is between $\htheta$ and $\htgt$ as the model learns $\ptgt$.
In this section, we will investigate these under our controlled experimental settings where the target distribution and its scale can be predefined.

%Transformer has been recognized as a powerful compressor \citep{LMisCompression, CompressionIntelligenceLinearly}. 
%In the past, our focus has primarily been on the discrepancy between the distribution learned by Transformer and the target one, which is often indirectly measured by metrics such as the cross-entropy loss \citep{LMisCompression, CompressionIntelligenceLinearly}.
%This is intuitive as the goal of compression is to learn the underlying distribution of data. However, what would happen if we break free from the constraints of the target distribution and purely examine the information content (entropy) of the distribution modeled by Transformers? 
%In this section, we will investigate this under our controlled experimental settings.

\subsection{Transformers Tend to learn a Low-entropy Distribution}\label{sec:low-entropy}

In this part, we not only examine the KL divergence but also investigate the entropy of the learned distribution, comparing it with the true entropy to reveal that the Transformer tends to learn a lower-entropy distribution when approaching the target distribution.

{\bf Basic Experiment Setting:} 
First, we generate the target distribution $\ptgt$ according to Section \ref{sec:pre} and the true entropy of obtained $\ptgt$ is $\htgt = 3.571$\footnote{In fact, there may be a slight difference between the sampled distribution and the generated target distribution, even though the number of sampled sequences is sufficient. Therefore, in order to compare with the most realistic distribution, here we take the sampled distribution as the target distribution $\ptgt$.}.
Then, we sample sufficiently sequences from it and train Transformers in the auto-regressive manner using cross-entropy loss.
For Transformers, we use the decoder architecture similar in \citet{vaswani2017attention}. 
As a comparison with Transformers, we select two classic RNNs: GRU \citep{GRU} and LSTM \citep{LSTM}.
We compare the differences in entropy of distributions modeled by Transformer and RNNs, that is, $\htf$ and $\hrnn$.
The embedding size $d$ for all models is chosen from $\{8, 16, 32, 64\}$ and the hidden size is set as $d_h = 4d$.
The number of heads is fixed at $h=4$.
We use a single-layer configuration for all RNNs and the number of layers of Transformers is adjusted as $L = 5$ to ensure the model size comparable to RNNs with the same $d$.
We use Adam\citep{Adam} as the optimizer and train all models for 100 epochs, monitoring the model's entropy $\htheta$ as well as $\kltheta$ since $\ptgt$ is known.
More details can be seen in Appendix \ref{app:base setting}.

\begin{figure*}[t]
	\centering
	\begin{subfigure}[t]{0.22\linewidth}
		\centering
		\includegraphics[scale=.20]{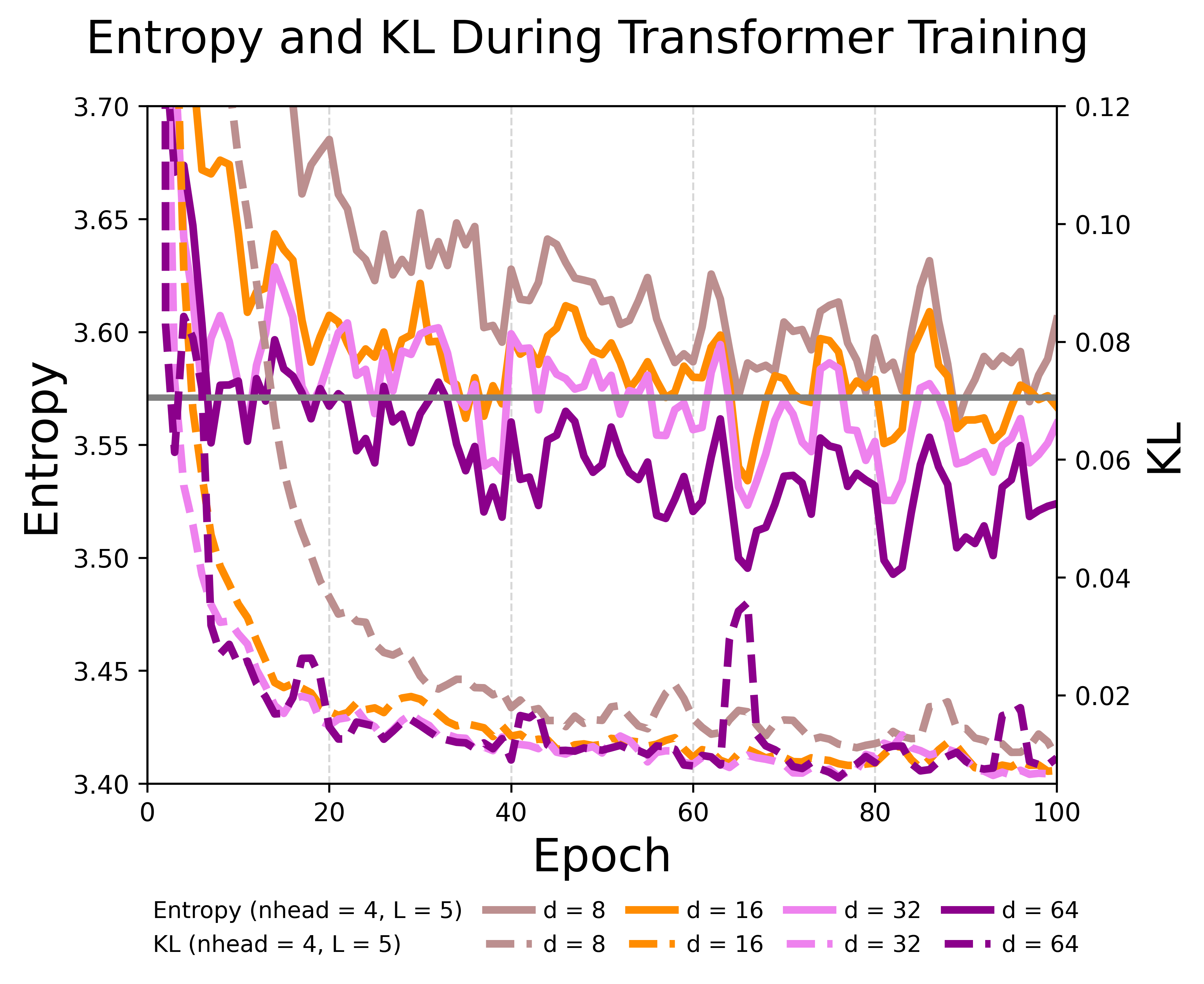}
		%		\subcaption{ }
		%		\label{fig:16-norm}
	\end{subfigure}
	\hspace{0.2cm}
	\begin{subfigure}[t]{0.22\linewidth}
		\centering
		\includegraphics[scale=.20]{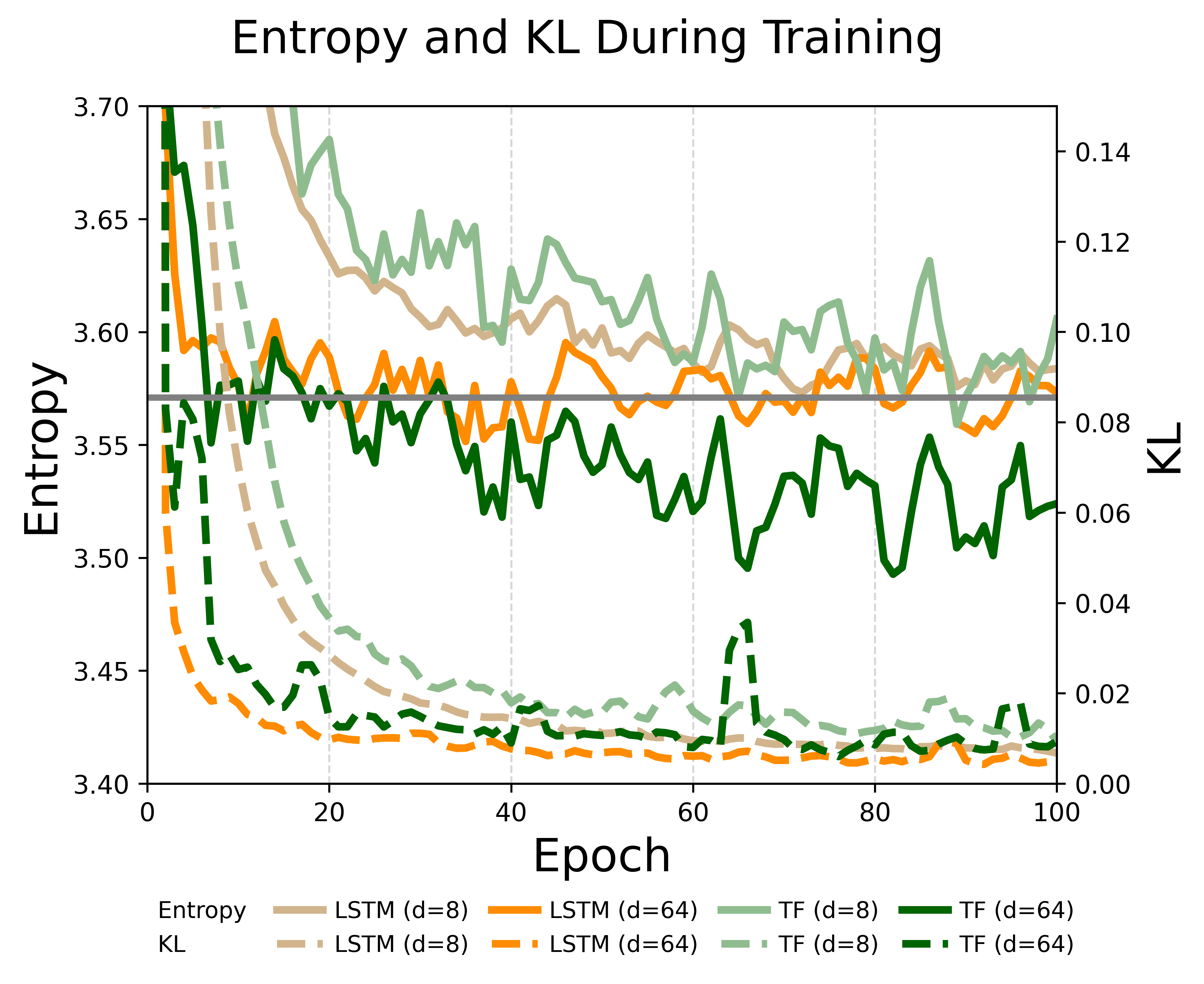}
		%		\subcaption{ }
		%		\label{fig:a}
	\end{subfigure}
	\hspace{0.2cm}
	\begin{subfigure}[t]{0.22\linewidth}
		\centering
		\includegraphics[scale=.202]{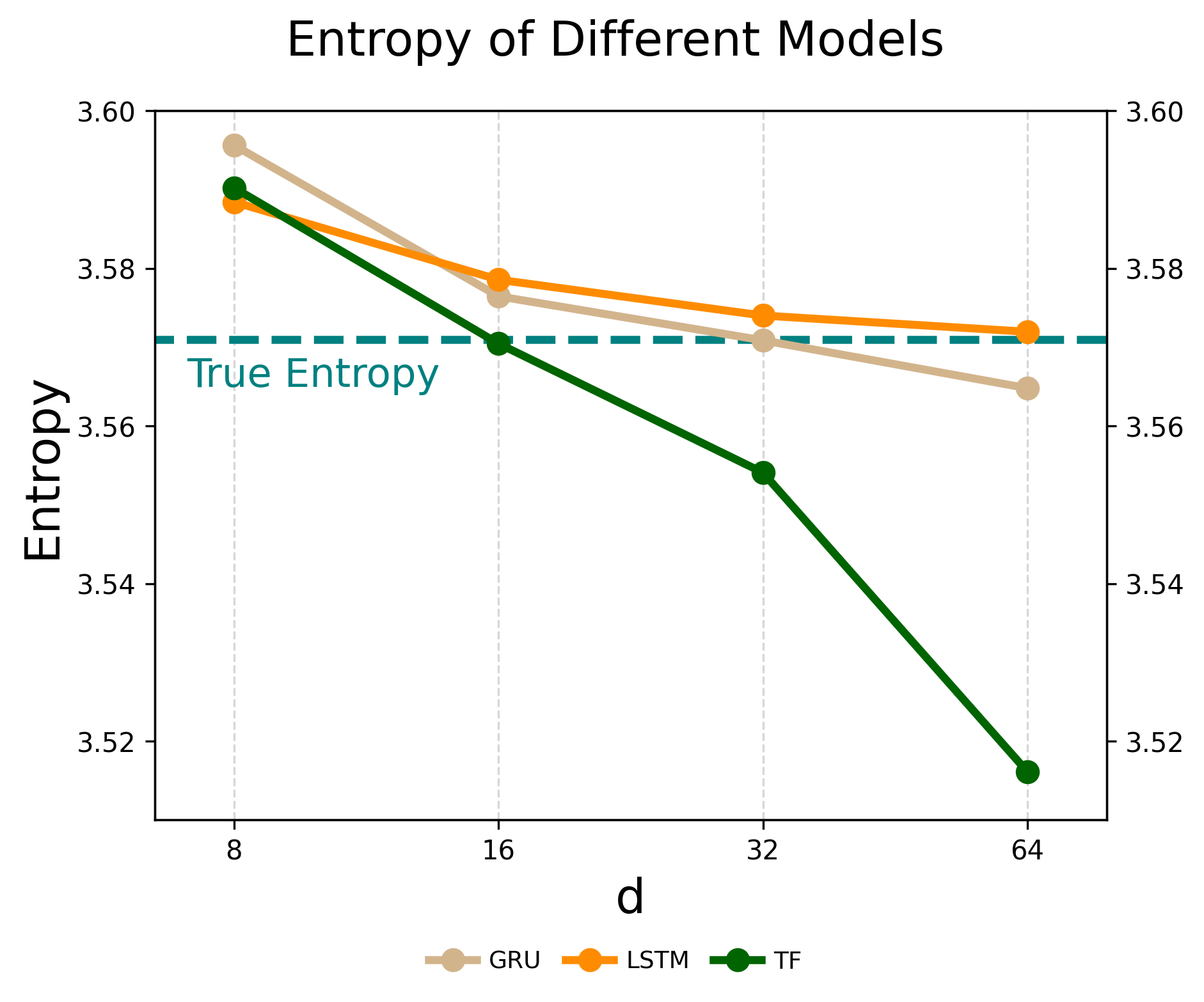}
		%		\subcaption{ }
		%		\label{fig:b}
	\end{subfigure}
	\hspace{0.2cm}
	\begin{subfigure}[t]{0.22\linewidth}
		\centering
		\includegraphics[scale=.205]{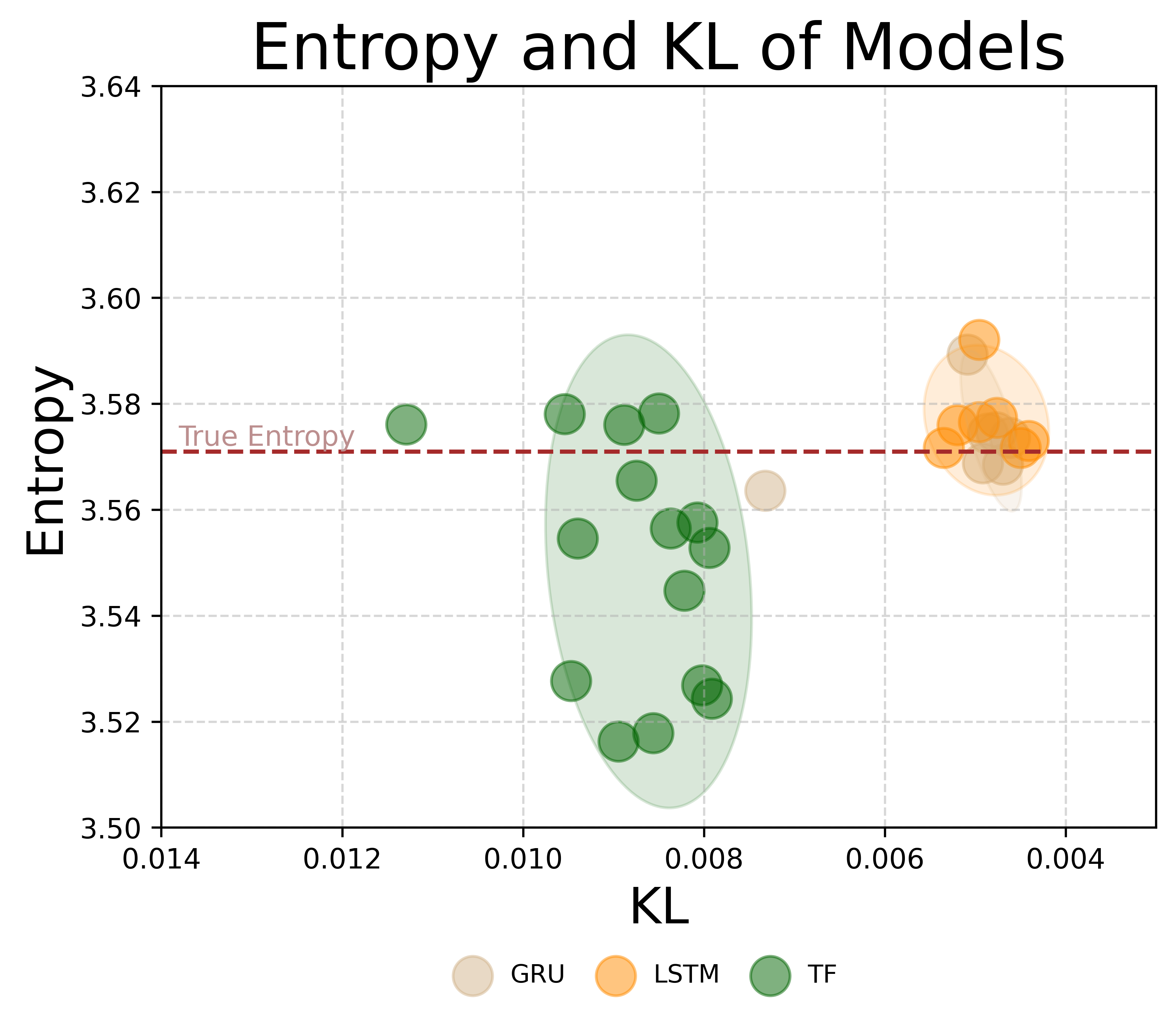}
		%		\subcaption{ }
		%		\label{fig:c}
	\end{subfigure}
	\vspace{-0cm}
	\caption{{\bf Left:} Entropy and KL During Training for Transformers of Different Sizes. {\bf Center Left:} Entropy and KL During Training for Transformer and LSTM when $d=8$ and $d=64$. {\bf Center Right:} The change in entropy with model size for GRU, LSTM and Transformer, averaged over the last 15 epochs. {\bf Right:} The relationship between entropy and KL for different model configurations (with each configuration averaged over the last 15 epochs).}
	\vspace*{-0cm}
	\label{fig:sec2-comp}
\end{figure*}

{\bf Larger Transformers can learn distributions with lower entropy:} 
Firstly, for Transformers, we monitor the changes in $\htf$ and $\kltf$ with different model sizes during training and the result is shown in the left part of Figure \ref{fig:sec2-comp}. 
It can be seen that as the model size increases, $\htf$ generally decreases. 
At first glance, this is not surprising: larger models often have stronger expressive power, making them more likely to achieve a distribution with lower entropy when approaching $\ptgt$ and the reduction in $\htf$~may be due to $\ptf \to \ptgt$~when enlarging the model size.
However, we notice that in the later stages of training, $\htf$~with larger models ($d = 64$) is often smaller than $\htgt = 3.571$ (denoted by a horizontal gray line).
This raises our speculation that: {\it Larger Transformers prefer to be learn $\ptf$~with lower entropy when approaching the target distribution $\ptgt$.}

{\bf Transformers prefer low-entropy landscape compared to RNNs:} 
Does RNNs exhibit similar behaviors when increasing the model size? Here we provide a negative answer.
As shown in the center left part of Figure \ref{fig:sec2-comp}, when the model size is small ($d=8$), there is no significant difference between $\hrnn$~and $\htf$, and meanwhile $\kltf$ is even higher than $\klrnn$.
However, when enlarging the model size($d=64$), $\hrnn$~fluctuates around $\htgt = 3.571$ while $\htf$~is significantly lower than $\hrnn$ even though Transformer does not model the target distribution as well as LSTM.
This can also be seen in the center right part of Figure \ref{fig:sec2-comp} where we compute the average entropy and KL divergence over the last 15 epochs for different model sizes.

To obtain a more general conclusion, we try various model settings (including changing the number of layers and the dimensions, see Appendix~\ref{app:base setting}) and also compute the average value over the last 15 epochs, with the result shown in the right part of Figure \ref{fig:sec2-comp}.
Overall, the entropy of the distributions modeled by Transformers is generally lower than RNNs, meanwhile it models the target distribution less well (with larger KL divergence).
In contrast, different RNN configurations consistently converge near the entropy of the target distribution ($\hrnn \approx \htgt$), causing the scatter points to eventually  converge in a small region, without showing the significantly lower entropy ($\htf < \htgt$) as seen in Transformers.
Thus, we arrive at our observation:
\begin{tcolorbox}[observation]
	{\bf Observation 1~}  {\it Transformers prefer to learn $\ptf$~with lower entropy when approaching the target distribution $\ptgt$.}
\end{tcolorbox}

{\bf Discussions: }
We interpret this preference of Transformers as an inductive bias introduced by model's architecture: during training, Transformers not only minimize the KL divergence with the target distribution but also intentionally reduces the entropy of the learned distribution $\ptheta$, which can be understood as an implicit regularization.
Mathematically, this can be formalized as:
\begin{equation}\label{eq:low-entropy}
	\min_{\vtheta} \gL(\vtheta) = \mathbb{E}_{\vs \sim \ptgt}\left[- \frac{1}{n} \sum_{i=1}^{n} \log p_{\vtheta} (\vs_i | \vs_{<i}) \right] + \alpha \htheta, 
\end{equation}
where  $\htheta = - \mathbb{E}_{\vs \sim \ptheta}\left[ \log p_{\vtheta}(\vs) \right]$ is a regularization term induced by the Transformer's architecture, and $\alpha$ is a coefficient related to the model size.
As seen in the right part of Figure~\ref{fig:sec2-comp}, this bias for lower entropy prevents Transformers from perfectly aligning with the target distribution, resulting in a larger KL divergence compared to RNNs.
This suggests that while compressing data, Transformers are not merely learning or memorizing the target distribution, but are further condensing its information content in their own way.
%Existing studies on the implicit bias of Transformers mainly focus on optimization algorithms and the properties of learned model parameters $\vtheta$ \citep{implicit_bias_LLMs, sheen2024implicit, vasudeva2024implicit}.
%We emphasize that the model architecture itself may also introduce the inductive bias from the lens of entropy, and this bias directly acts on the learned distribution $\ptheta$.
However, as previously mentioned, this bias in practice may be challenging to observe due to the exponentially growing cost of computing entropy.
Additionally, it should be also noted that Eq~(\ref{eq:low-entropy}) is given by intuition to help understanding, and we leave a more rigorous theoretical analysis for our future work.
\begin{wrapfigure}{r}{0.33\textwidth}  % r 表示靠右，0.4\textwidth 是图宽
	\centering
	\includegraphics[width=0.33\textwidth]{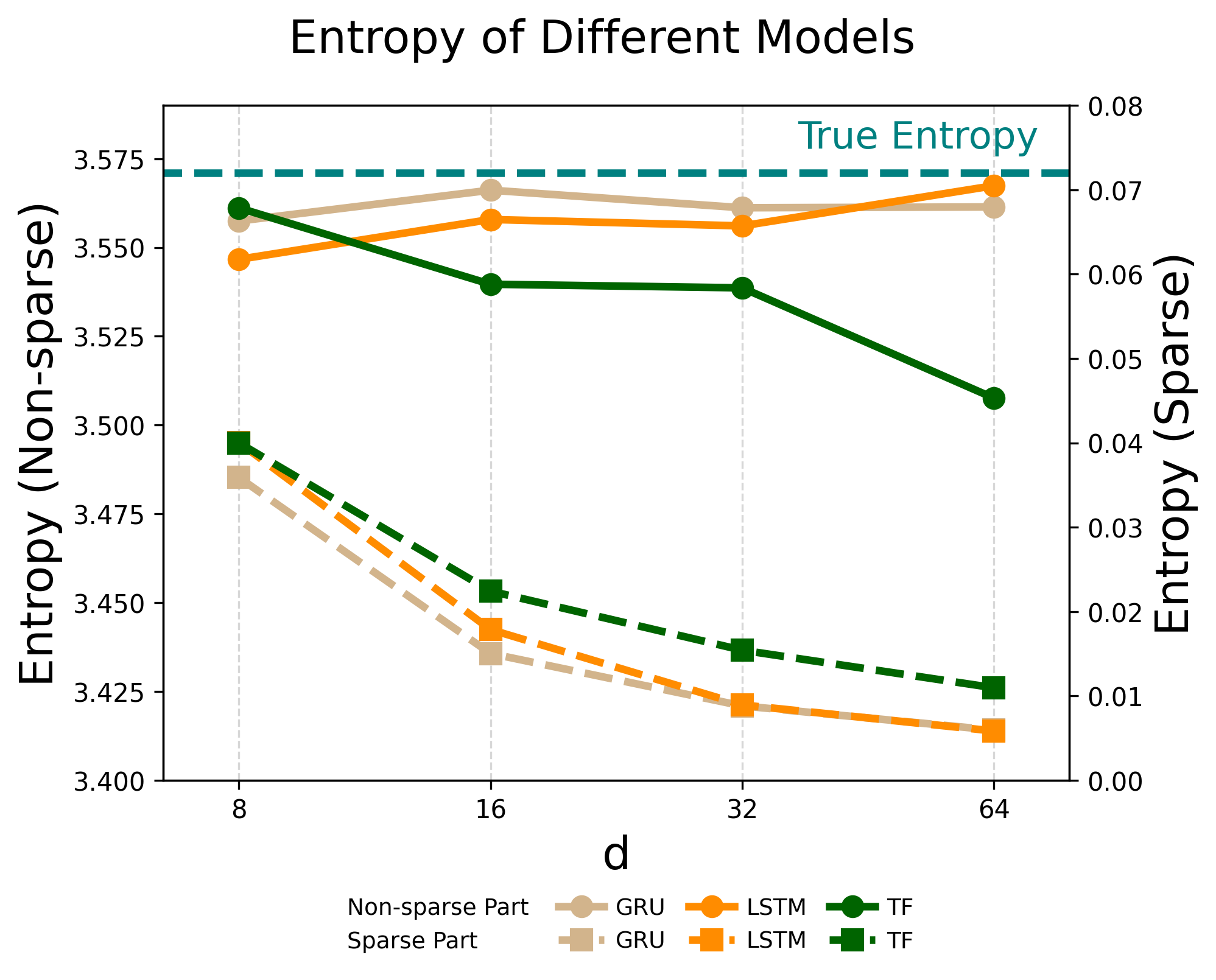}  % 你的图片文件
	\caption{The change in entropy of the sparse and non-sparse parts with model size.}
	\vspace{-0.5cm}
	\label{fig:sec2-part}
\end{wrapfigure}

\textbf{Larger Transformers tend to produce more deterministic distributions.}
Further, we examine the entropy of the following two parts of the modeled distribution:  
(1) {\bf Sparse part} includes sequences with a probability of $0$ in the target distribution and models should assign probabilities close to $0$ to them, contributing nearly $0$ to the total entropy;  
(2) {\bf Non-sparse part} includes the rest sequences with nonzero probabilities and models should approximate the true probabilities for these sequences, which contributes the majority of the entropy.
The main results can be seen in Figure~\ref{fig:sec2-part}.
As the model size increases, the probabilities assigned by Transformers to the sparse part become lower, leading to a smaller entropy in this part, which suggests that larger Transformers are better at learning zero probabilities. 
This also means $\ptf$ approximates $\ptgt$ more closely in the sparse part.
At the same time, the entropy of the non-sparse part decreases, indicating that larger Transformers tends to learn more deterministic distributions.
This results in $\ptf$ deviating from $\ptgt$ in the non-sparse part, which appears slightly contradictory to its behavior in the sparse part.
In contrast, RNNs gradually approach the true entropy in non-sparse part as model size increases.
Therefore, we conclude that larger Transformers generally achieve lower entropy by yielding more deterministic distributions. 

\subsection{The FFN Module Seeks the Low-entropy Optimization Landscape}\label{sec:entropy-FFN}

In the above part, we show that Transformers tend to favor lower-entropy distributions during training. 
The next question is: {\it Which components of Transformers drive this preference for low-entropy optimization landscapes?} 
Here, we focus on the attention modules and the FFN modules, whose alternating stacking forms the fundamental backbone of Transformers.
To analyze the impact of each component, we modify the original Transformer as following:
(1) {\bf Attention-only:} All FFN modules are removed, retaining only the attention modules.  
(2) {\bf Attention-main:} Only the FFN module in the final layer is retained, while all other FFN modules are removed.  
(3) {\bf FFN-main:} Only the attention module in the first layer is retained, while the remaining attention modules are removed.
Note that we don't consider the {\bf FFN-only} setting as a purely stacked FFN module can not effectively model sequences. 
The main results are shown in Figure~\ref{fig:sec2-ablation} and more details can be seen in Appendix~\ref{app:base setting}.

{\bf FFN prefers exploring low-entropy regions:} 
We compare the modified models and present the case of $d=64$ in the left part of Figure~\ref{fig:sec2-ablation} (more cases can be found in Figure~\ref{app:fig:entropy_kl_attention_ffn}). 
It can be observed that compared to Attention-only and Attention-main, FFN-main exhibits lower entropy comparable to that of original Transformer and it is also lower than the true entropy $\htgt = 3.571$ in the later stages of training.
Furthermore, we also try more settings of all models and also compute the average entropy, KL divergence, and loss over the final 15 epochs.
As shown in the center left and right parts of Figure \ref{fig:sec2-ablation}, the entropy of Attention-only is basically higher than the true entropy while after adding one layer of FFN module, the entropy of Attention-main decreases and fluctuates around the true entropy.
For FFN-main and original Transformers, which retain all FFN modules, their entropy is mostly lower than the true entropy.
Therefore, we draw our conclusion as following:
\begin{tcolorbox}[observation]
	{\bf Observation 2~}  {\it It is the FFN module that drives the Transformer to seek the distribution  $\ptf$ with lower entropy when approaching the target distribution $\ptgt$.}
\end{tcolorbox}

{\bf Can loss reflect the effectiveness in learning the target distribution?}
The objective of the cross-entropy loss is to enable the model to learn the conditional probability of the target distribution given prefixes of varying lengths.
Undoubtedly, in the ideal case, when the model learns each true conditional distribution given different lengths, the final joint distribution it learns would match the target distribution.
However, when the model's conditional probability learning is not optimal, what is the relationship between the model's loss and the KL divergence?
Under our controlled experiments, since the target distribution is known, we can measure this relationship for different variants, as shown in the right part of Figure~\ref{fig:sec2-ablation}.
It can be observed that for each variant, there exists a clear linear relationship between the KL divergence and the loss, which is consistent with the findings of \citet{CompressionIntelligenceLinearly,emergent-loss}, where they also discover that the model's downstream task performance exhibits a strong linear correlation with the loss.
However, it should be noted that different models with identical loss values may have different KL divergences.

\begin{figure*}[t]
	\centering
	\begin{subfigure}[t]{0.22\linewidth}
		\centering
		\includegraphics[scale=.20]{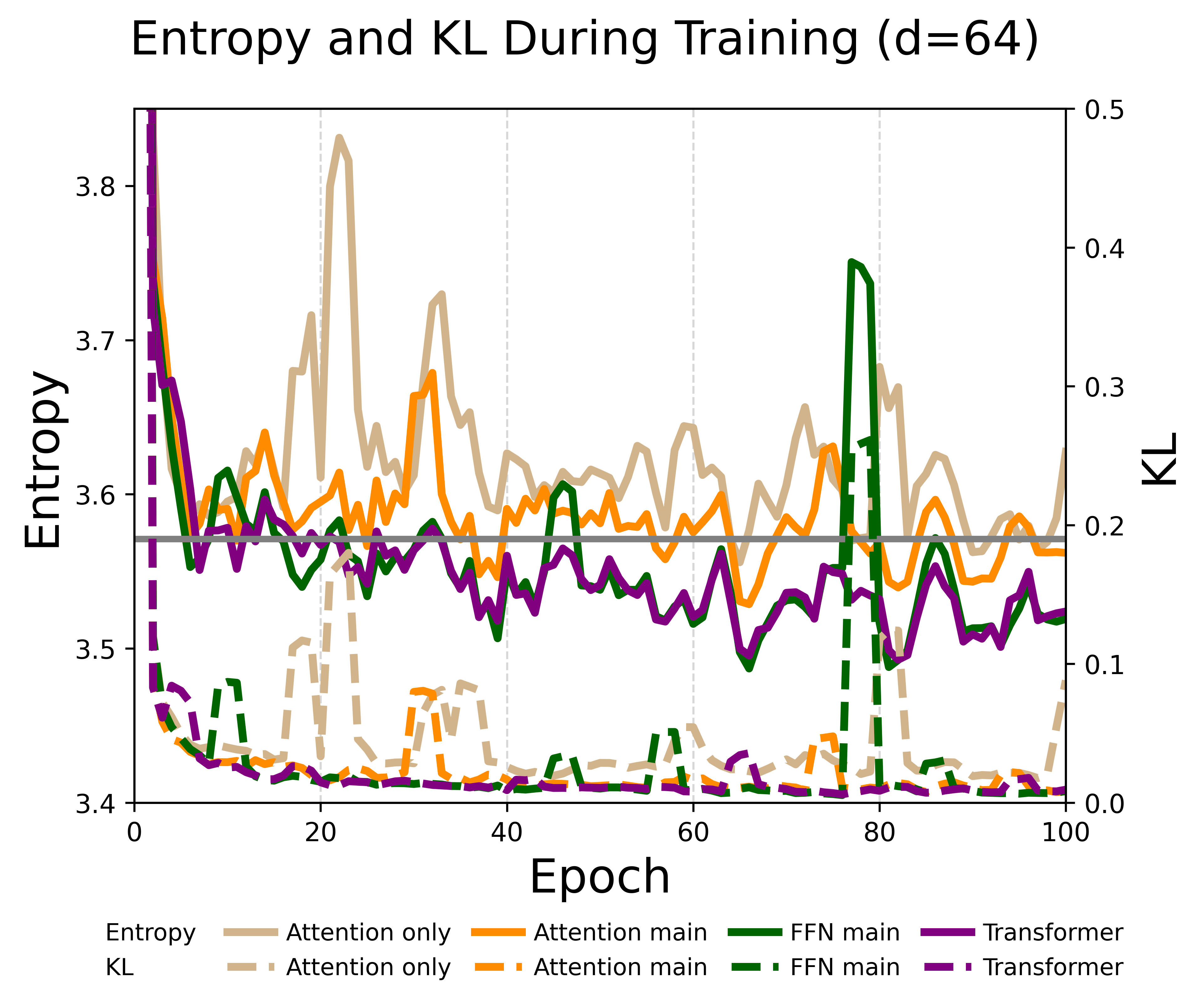}
		%		\subcaption{ }
		%		\label{fig:16-norm}
	\end{subfigure}
	\hspace{0.2cm}
	\begin{subfigure}[t]{0.22\linewidth}
		\centering
		\includegraphics[scale=.205]{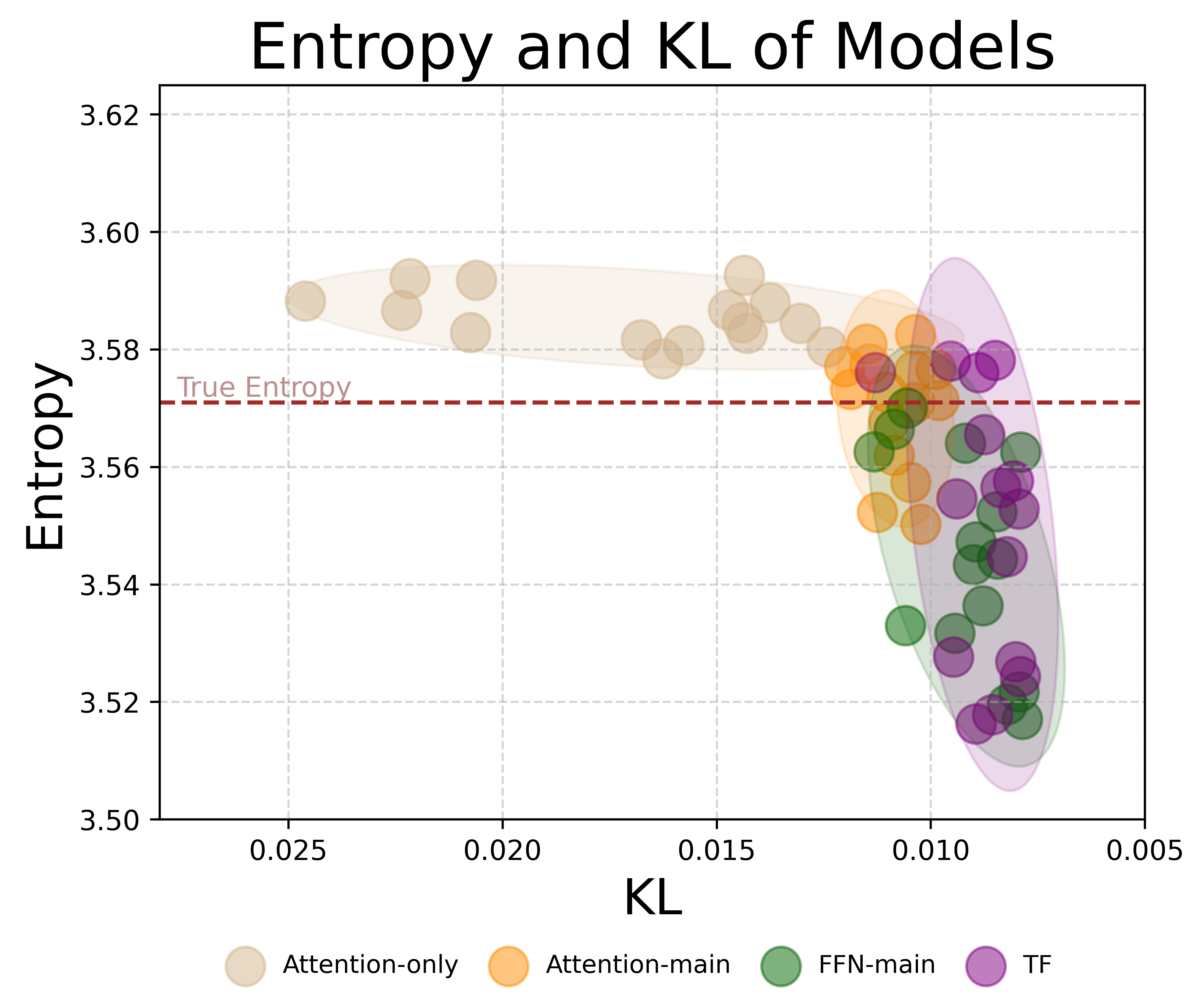}
		%		\subcaption{ }
		%		\label{fig:a}
	\end{subfigure}
	\hspace{0.2cm}
	\begin{subfigure}[t]{0.22\linewidth}
		\centering
		\includegraphics[scale=.205]{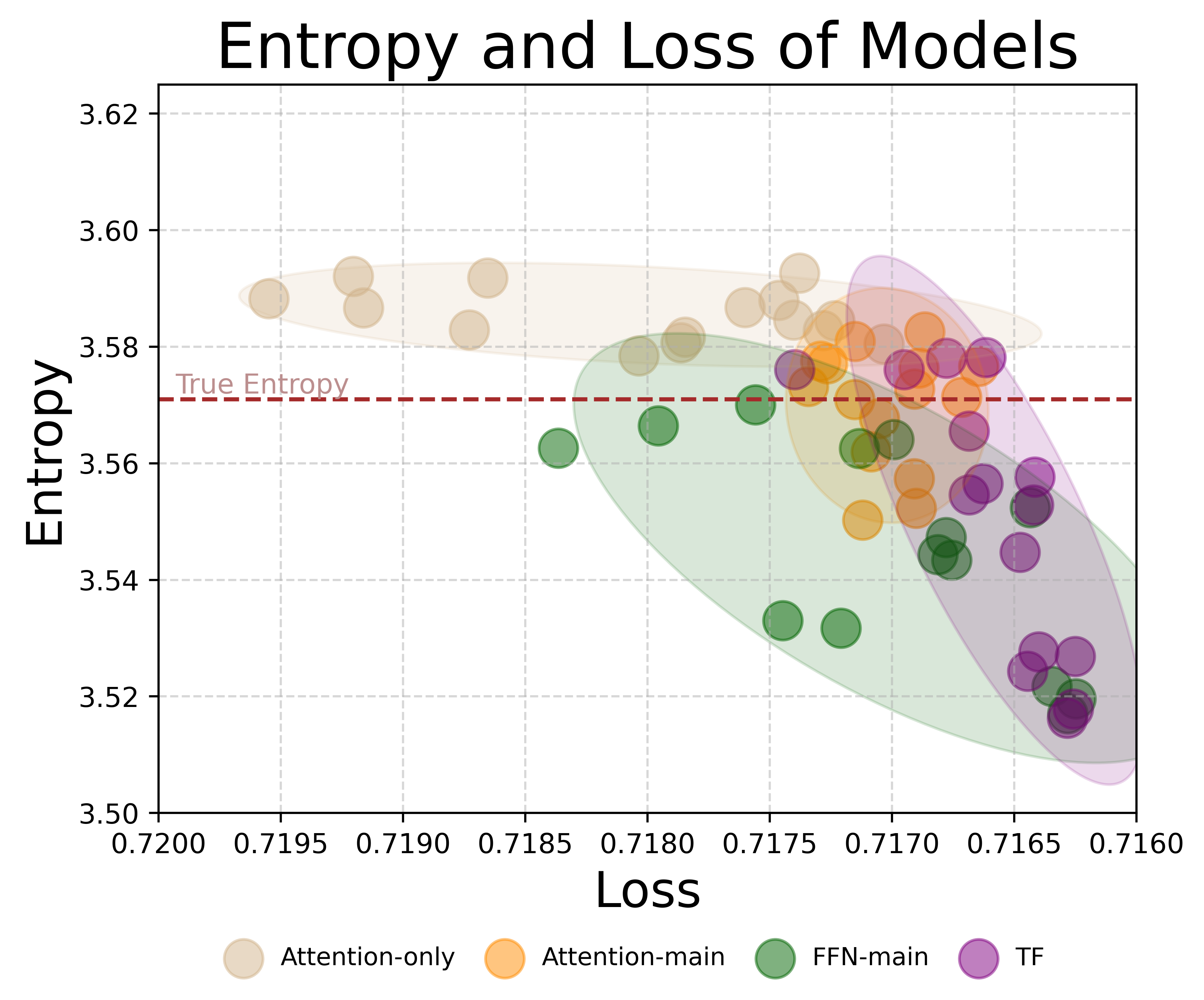}
		%		\subcaption{ }
		%		\label{fig:b}
	\end{subfigure}
	\hspace{0.2cm}
	\begin{subfigure}[t]{0.22\linewidth}
		\centering
		\includegraphics[scale=.205]{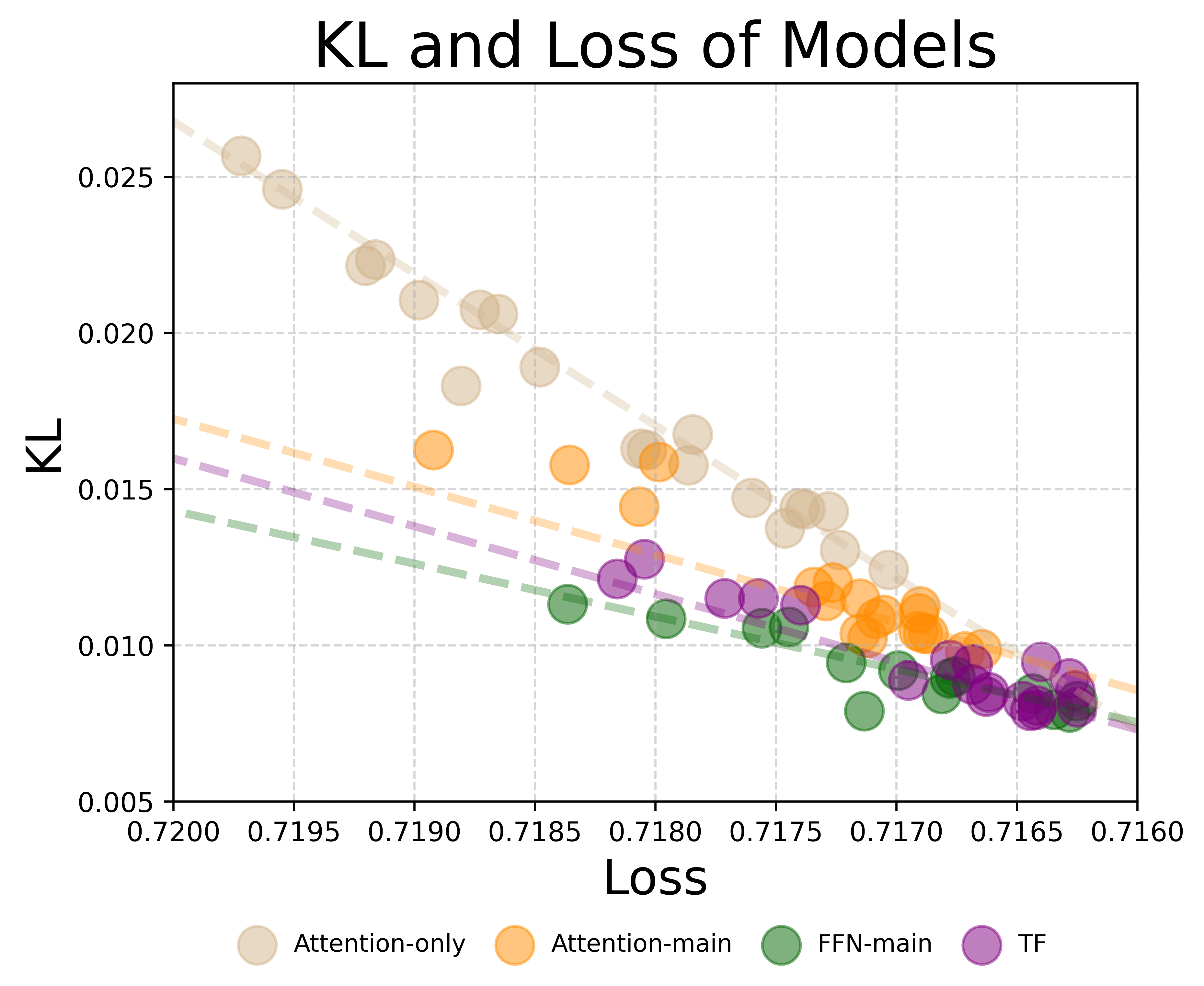}
		%		\subcaption{ }
		%		\label{fig:c}
	\end{subfigure}
	\vspace{-0cm}
	\caption{
		{\bf Left:} Entropy and KL during training for different Transformer variants. 
		{\bf Center Left and Center Right:} Relationship between KL/Loss and Entropy for different variants. FFN-main can achieve lower entropy compared to Attention-only and Attention-main.
		{\bf Right:} Relationship between KL and Loss for different variants.
	}
	\vspace*{-0cm}
	\label{fig:sec2-ablation}
\end{figure*}

\section{Dynamic Sparsity in Transformers}
\begin{figure}[t]
	\centering
	\includegraphics[scale=0.40]{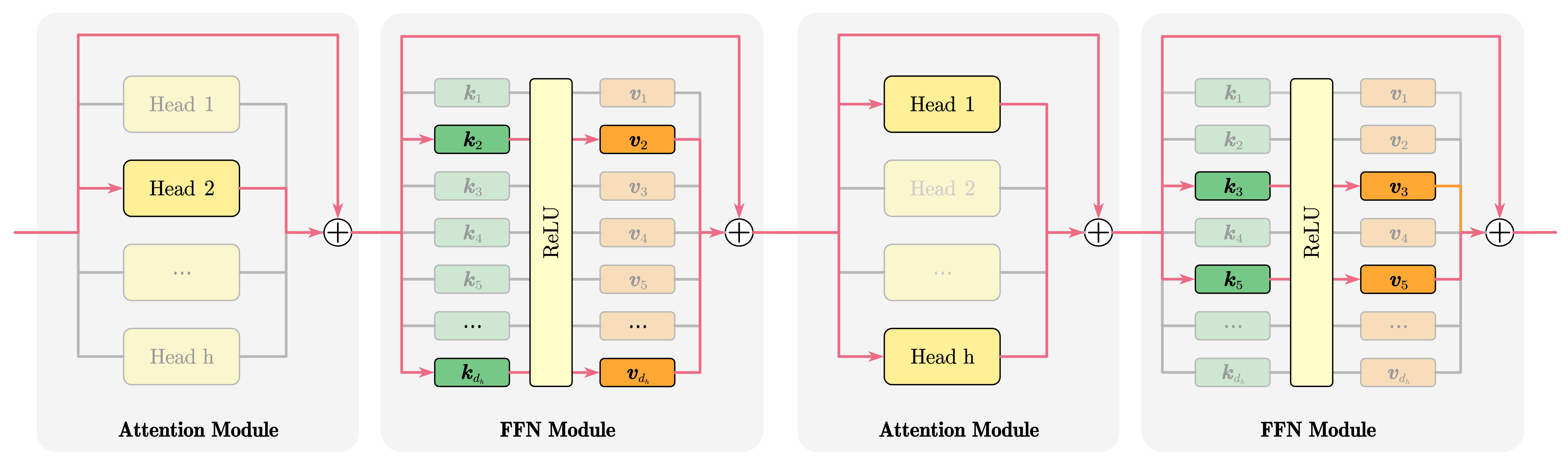}
	\caption{
	Explanation of dynamic sparsity in the attention and FFN modules of Transformers. As for the Attention modules, each attention head or residual connection can be viewed as a forward path and	larger Transformers prefer residual connections to bypass attention head computations. 
	As for the FFN modules, the parameters of the first and second layer can be viewed as key and value vectors respectively and each neuron is a key-value pair.
	Larger Transformers will have a lower proportion of active neurons (i.e., key-value pairs).
	}
	\label{fig:dynamic_sparse}
\end{figure}

In the above, we reveal the low-entropy preference exhibited by Transformers when compressing data.
Data compression emphasizes the removal of redundant information in the data, while in this section, we shift our focus to the parameter redundancy of models, which can be characterized by dynamic sparsity. 
We concentrate on the two most critical components of Transformers: the Attention module and the FFN module, to illustrate that larger Transformers often have a stronger preference for dynamic sparsity.
An overview of this section can be seen in Figure \ref{fig:dynamic_sparse}.

\subsection{Dynamic Sparsity in the Attention Module}\label{sec:attention}
The interaction between the attention layer and the residual connection is considered one of the key factors behind the success of Transformers\citep{attention_rank_collapse,no_residual,simplify_TF}. \citet{attention_rank_collapse} regard attention heads in each layer as optional paths and theoretically show that simply stacking attention layers will lead to a phenomenon known as rank collapse, where the model will lose expressive power and output uniform representations for different tokens.
However, introducing residual connections can address this issue by providing the model with potentially shorter forward pathways.
{\it Despite this, it remains unclear whether the model implicitly prefers these shorter pathways when processing inputs.}
Therefore, to observe this, we slightly modify the original attention module by incorporating a dynamic routing mechanism, that is, for the $i$-th token $\vx_i$, the output of the attention module will be
\begin{equation}
\begin{aligned}
	&\mathrm{Atten}(\vx_i) = \sum_{k=1}^{h} f_k(\vx_i) \mathrm{Head}^{(k)}(\vx_i) + f_{h+1}(\vx_i) \vx_{i}, \\
	\mathrm{Head}^{(k)}(&\vx_i)  = \mP^{(k)} \mW_V^{(k)} \mX\mathrm{Softmax}\left( \frac{1}{\sqrt{d}}\left(\mW_K^{(k)} \mX \right)^T\mW_Q^{(k)} \vx_{i} + \vm_{i} \right),
\end{aligned}
\end{equation}
where $\mathrm{Head}^{(k)}$ is the output of $k$-th attention head, $\vm_i$ is the casual mask for the $i$-th token, $f(\vx) = \mathrm{Softmax}(\mW_2 \sigma(\mW_1 \vx)) \in \sR^{h+1}$ is the routing function dynamically computed based on the input $\vx$ and $f_k(\vx)$ represent the weight for the $k$-th head path.
Specially, we use $f_{h+1}(\vx)$ to denote the weight for the residual connection path. 
After training such Transformers as in Section~\ref{sec:low-entropy}, we iterate over all possible sequences as inputs and analyze the weight distribution of all paths. 
We set $h=4$ in Transformers by default, resulting in a total of $h+1=5$ paths per layer when considering the additional residual connection. 
We equally partition the interval $[0,1]$ into 30 subintervals and calculate the proportion of weight values falling into each one.
%The main results are shown in the left and center left parts of Figure \ref{fig:sec3-atten}.

{\bf Larger Transformers prefer residual connections:} First, we vary the model size and show the weight distribution of all paths in the left part of Figure \ref{fig:sec3-ds}.
It can be observed that when the model size is small, the weights of all paths are concentrated around $\frac{1}{h+1} = 0.2$, indicating that all paths in each layer contribute to the forward computation.
However, as the model size grows, the proportion of weights with smaller values begins to increase.
Notably, when $d=64$, more than half of the weights are assigned values close to 0, meaning that they are almost ignored in forward computation. 
Furthermore, we show the distribution of weights assigned to residual connections in the center left of Figure \ref{fig:sec3-ds}.
As the model size increases, the proportion of higher weight values begins to rise and when $d=64$, nearly 30\% of the weight values are close to 1.
This indicates that larger Transformers have a stronger preference for residual connections.
Therefore, we conclude as follows:
\begin{tcolorbox}[observation]
	{\bf Observation 3~}  {\it Larger Transformers have a stronger preference for residual connections, bypassing the attention computations and leading to a more sparse computation pattern in the attention module.}
\end{tcolorbox}

\subsection{Dynamic Sparsity in the FFN Module}\label{sec:FFN}

The FFN module, which accounts for nearly two-thirds of the model’s parameters, is also a crucial component of the Transformer.
In this part, we turn our focus to the dynamic sparsity of the FFN layer.
Previous works \citep{FFN_key_value, neurons_dead, softmax_relu_tf} treat the FFN layer as key-value memories. 
\begin{figure*}[t]
	\centering
	\begin{subfigure}[t]{0.22\linewidth}
		\centering
		\includegraphics[scale=.21]{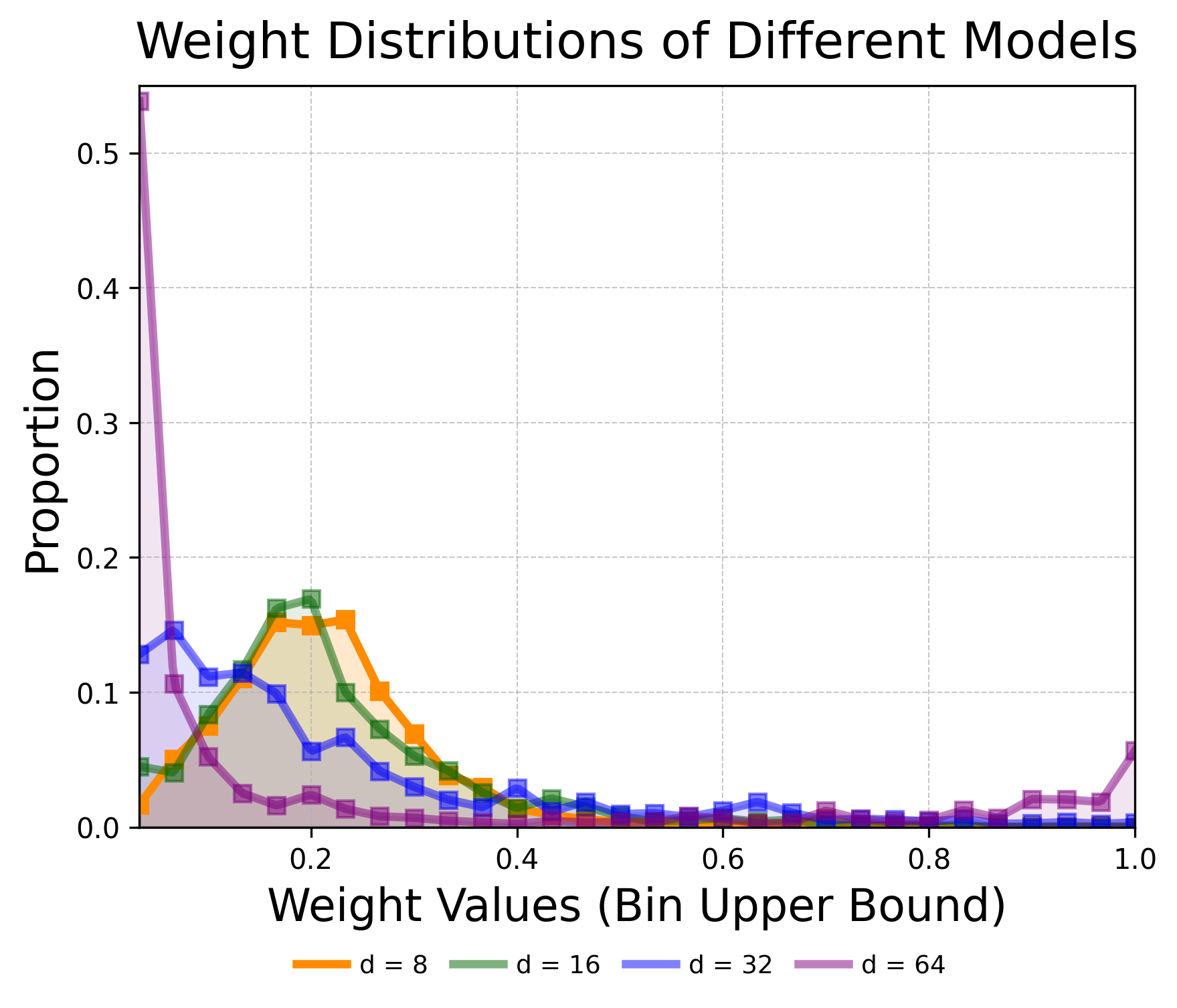}
		%		\subcaption{ }
		%		\label{fig:16-norm}
	\end{subfigure}
	\hspace{0.1cm}
	\begin{subfigure}[t]{0.22\linewidth}
		\centering
		\includegraphics[scale=.21]{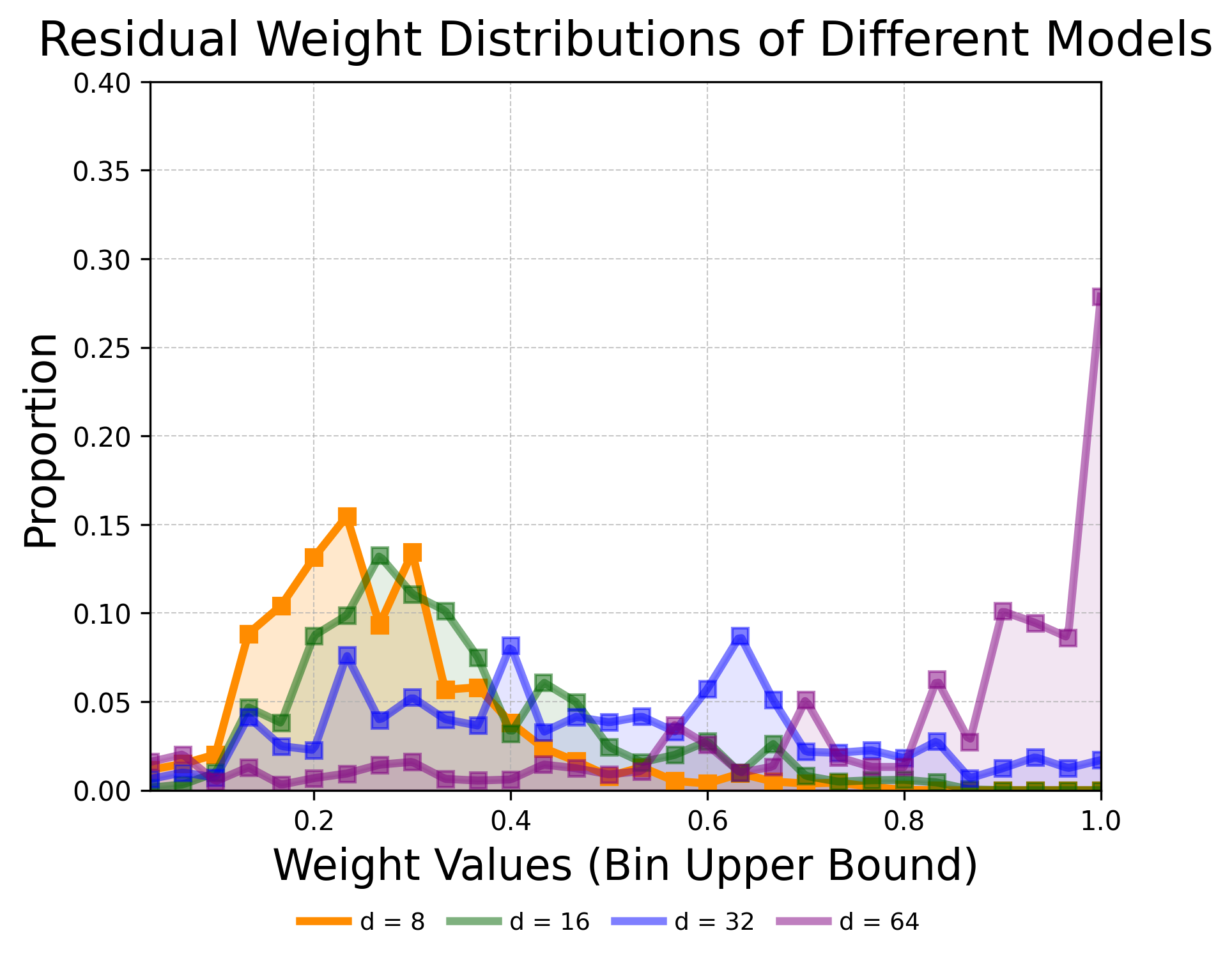}
		%		\subcaption{ }
		%		\label{fig:a}
	\end{subfigure}
	\hspace{0.1cm}
	\begin{subfigure}[t]{0.22\linewidth}
		\centering
		\includegraphics[scale=.21]{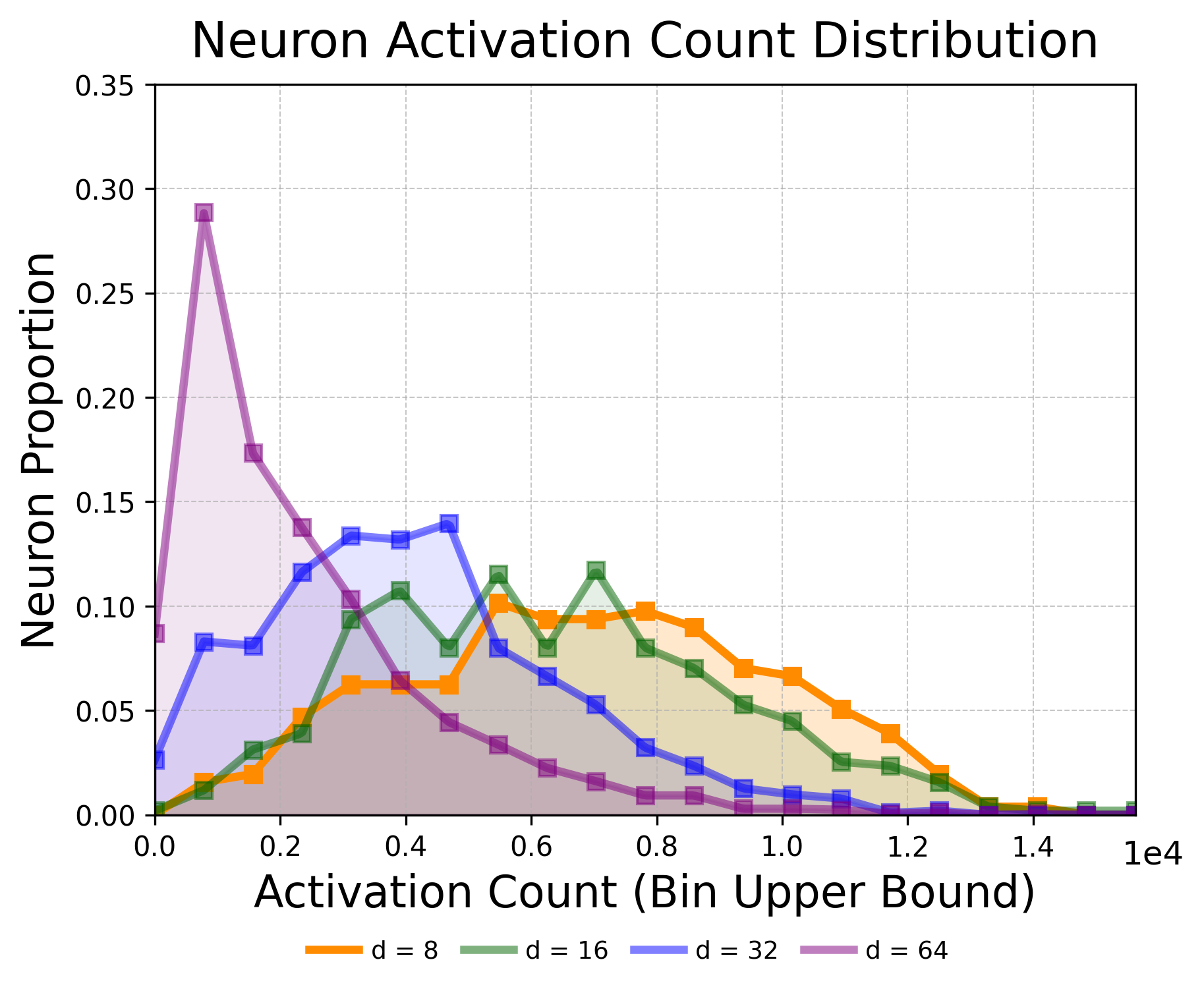}
		%		\subcaption{ }
		%		\label{fig:b}
	\end{subfigure}
	\hspace{0.1cm}
	\begin{subfigure}[t]{0.22\linewidth}
		\centering
		\includegraphics[scale=.21]{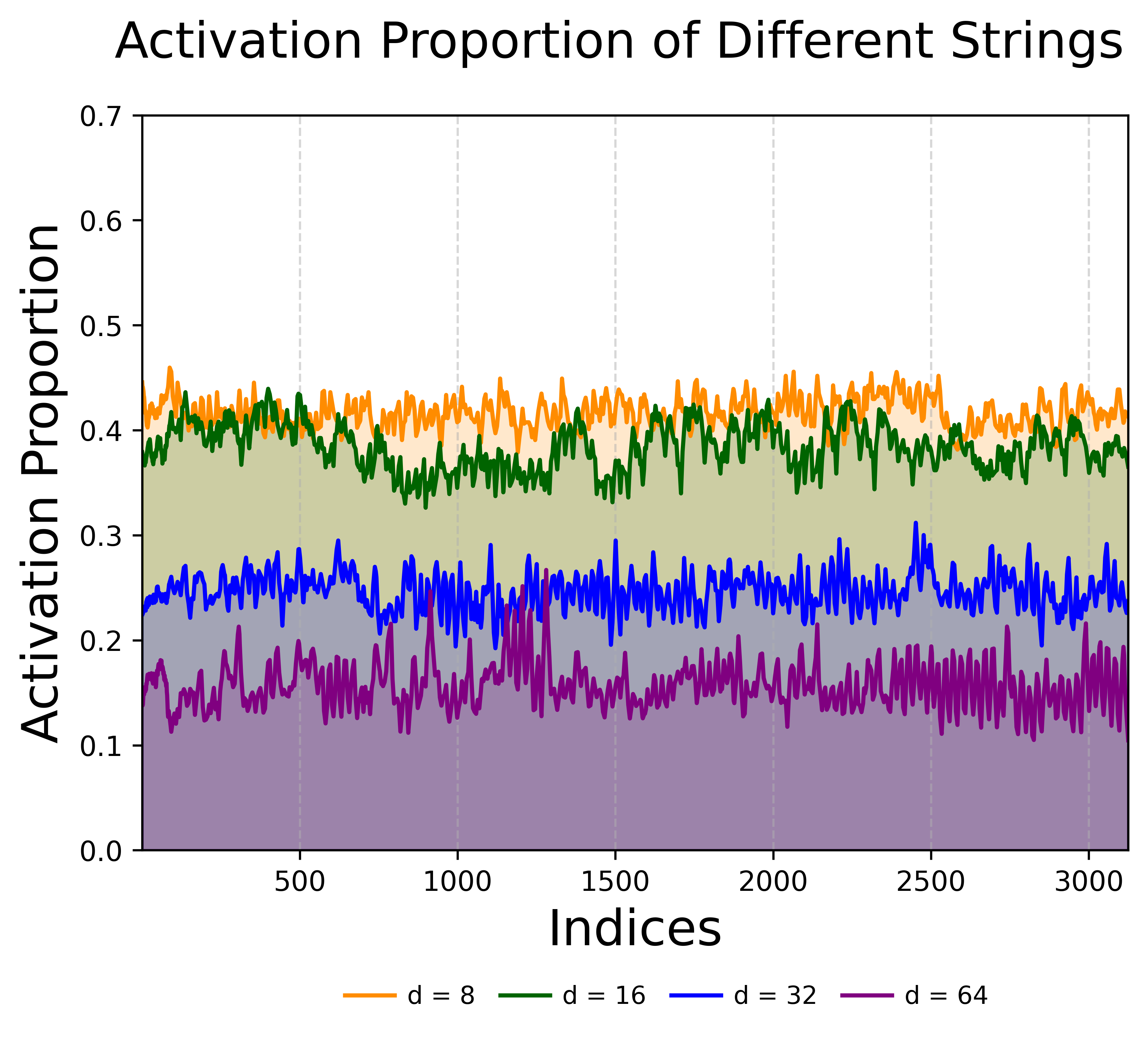}
		%		\subcaption{ }
		%		\label{fig:c}
	\end{subfigure}
	\vspace{-0cm}
	\caption{
		{\bf Left:} The weight distribution of all paths for different model sizes.
		{\bf Center Left:} The weight distribution of only residual paths for different model sizes.
		{\bf Center Right:} The distribution of neuron activation counts for Transformers of different sizes.
		{\bf Right:} The proportion of neurons activated across different input sequences.
	}
	\vspace*{-0cm}
	\label{fig:sec3-ds}
\end{figure*}
Specifically, the first-layer parameters $\mW_1 = [\vk_1, \vk_2, \dots, \vk_{d_h}] \in \sR^{d_h \times d}$ of the FFN consist of $d_h$ key vectors, while the second-layer $\mW_2 = [\vv_1, \vv_2, \dots, \vv_{d_h}]^T \in \sR^{d \times d_h}$ can be viewed as $d_h$ value vectors.
For a given input $\vx \in \sR^d$, it first computes similarities with the key vectors, passes through an activation function (e.g., ReLU) for filtering, and then applies a weighted sum to selected value vectors, which can be formulated as:
\begin{equation}
	\mathrm{FFN}(\vx) = \mW_2 \mathrm{ReLU}(\mW_1 \vx) = \sum_{i=1}^{d_h} \max(\vx^T \vk_i, 0) \vv_i .
\end{equation}
In this process, the FFN exhibits dynamic sparsity, that is, for a given $\vx$, only a subset of neurons is selectively activated and really contributes to the computation of output.
\citet{neurons_dead} analyze neuron activations by feeding diverse datasets into LLMs and demonstrate the presence of dynamic sparsity in LLMs.
For example, they show that there are certain key-value pairs that are never activated in the OPT model family\citep{opt} of various sizes.
\begin{wrapfigure}{r}{0.33\textwidth}  % r 表示靠右，0.4\textwidth 是图宽
	\centering
	\vspace{-0.0cm}
	\includegraphics[width=0.33\textwidth]{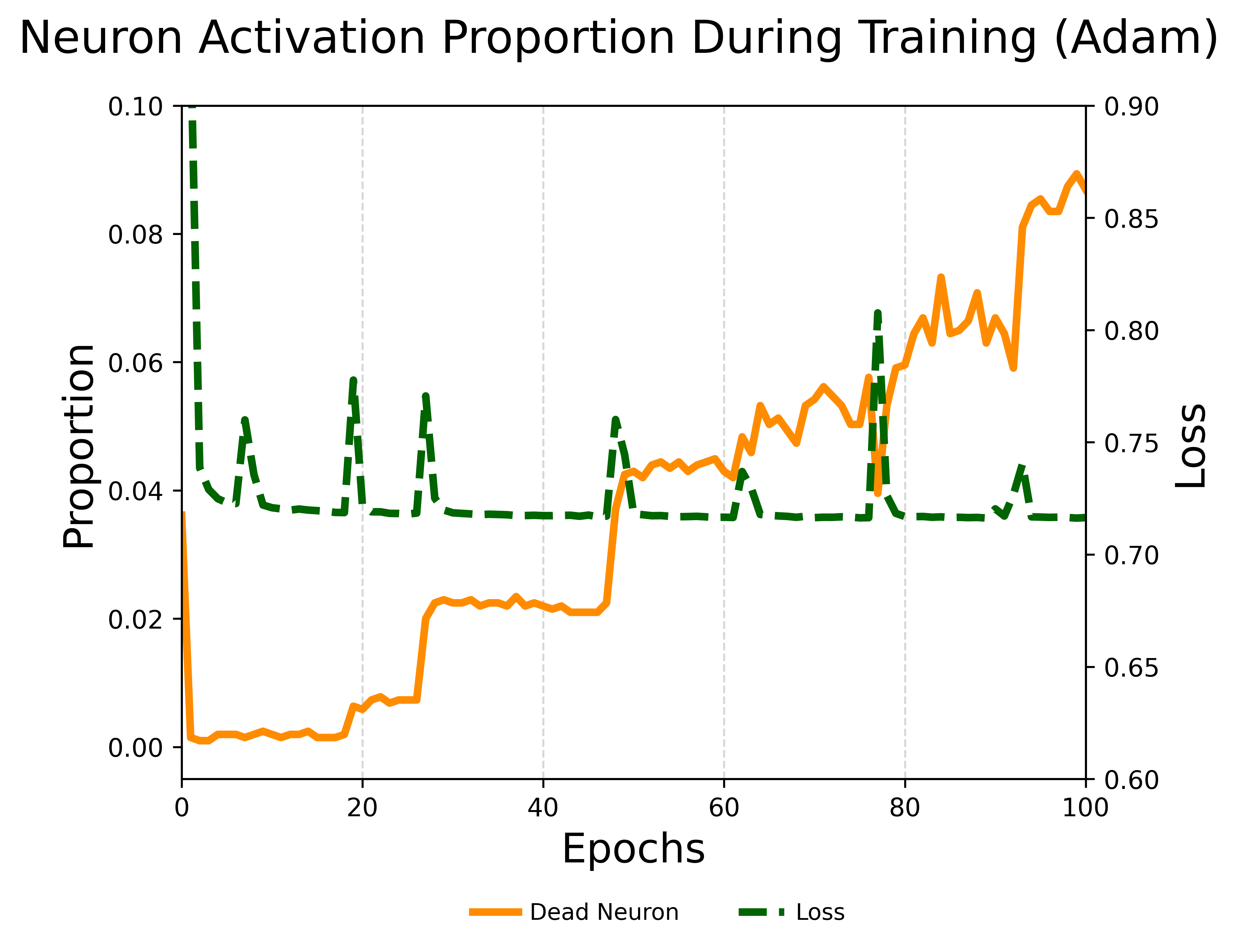}  % 你的图片文件
	\caption{Loss and proportion of dead neurons during training when $d = 64$.}
	\vspace{-0.5cm}
	\label{fig:FFN-monitor}
\end{wrapfigure}

However, exploring this sparsity in real-world scenarios faces following challenges: 
(1) Although we can collect large-scale corpora as inputs, it still remains difficult to determine whether this sparse pattern persists in unseen potential inputs. 
(2) The formation process of this sparse pattern during training still requires further exploration, and monitoring it on large-scale corpora entails substantial computational costs.
Therefore, instead of directly focusing on real-world scenarios, we explore these issues under controlled settings.
Our experimental setup is basically consistent with that in Section \ref{sec:low-entropy} and more details can be found in Appendix \ref{app:FFN setting}. 
Here, all possible inputs consist of $|\gV|^{n} = 3125$ sequences of length $n=5$, allowing us to examine the activation of each FFN neuron across 15,625 tokens, which is also the maximum activation count $N_{\max}$.
We equally divide $(0, N_{\max}]$ into 20 subintervals and count the proportion of neurons whose activation counts fall into each one.
Additionally, we specially count the proportion of dead neurons (i.e., never activated). 

{\bf Larger Transformers exhibit more pronounced dynamic sparsity:} 
In the cneter right part of Figure \ref{fig:sec3-ds}, we show the distribution of neuron activation counts across different model sizes. 
We find that smaller models have a higher proportion of active neurons while in larger models, neurons tend to be more inactive.
Notably, when $d = 64$, over 25\% of neurons have activation counts within $(0, 5\%N_{\max}]$.
Meanwhile, nearly 10\% of the neurons remain completely inactive across all possible inputs, indicating that they do not contribute to the final output at all.
These findings are similar to those of \citet{neurons_dead} where they test OPT families \citep{opt} with different corpora.
The difference is that here we traverse all possible $|\gV|^{n}$ sequences as inputs under our controlled settings, which is often difficult to achieve in real scenarios because $|\gV|$ is typically very large.
In addition, in the center left part of Figure \ref{fig:sec3-ds}, we show the proportion of neurons activated by each input sequence.
It can be seen that larger models activate a lower proportion of neurons on average for each input sequence.
When $d=64$, only about 15\% of neurons are activated on average for each input sequence, whereas for $d=8$, this number exceeds 40\%.
Thus, we draw our conclusion as following:
\begin{tcolorbox}[observation]
	{\bf Observation 4~}  {\it 
		Larger Transformers exhibit stronger dynamic sparse patterns in the FFN module when faced with all possible inputs.}
\end{tcolorbox}

{\bf The FFN module achieves sparsity in a jump-like manner:}
One may ask that: {how is the dynamic sparse pattern of the FFN module formed?}
To further investigate this, we monitor neuron activations throughout the training process.
In Figure~\ref{fig:FFN-monitor}, we present the proportion of dead neurons during training when $d = 64$.
Surprisingly, we find that this dynamic sparsity enhances in a jump-like manner rather than continuously.
Moreover, the enhancement timing closely coincides with the occurrence of loss spikes, where the loss rapidly increases and then returns to its previous level.
This suggests that loss spikes are highly correlated with the sudden decrease in neuronal activity.
During training, the model discards certain neurons in phases and re-optimizes the remaining ones, gradually exhibiting dynamic sparsity.
Notably, loss spikes also commonly occur in LLMs training \citep{loss_spike_phenomenon, loss_spike_adam} and in our controlled experimental setup, we also observe that loss spikes occur less frequently in smaller models (see Figure \ref{app:fig:loss_spike_model_size} in Appendix).
As illustrated above, this could be because the proportion of inactive neurons is lower in smaller models, thereby having a limited impact on training stability.

\begin{figure*}[t]
	\centering
	\begin{subfigure}[t]{0.22\linewidth}
		\centering
		\includegraphics[scale=.20]{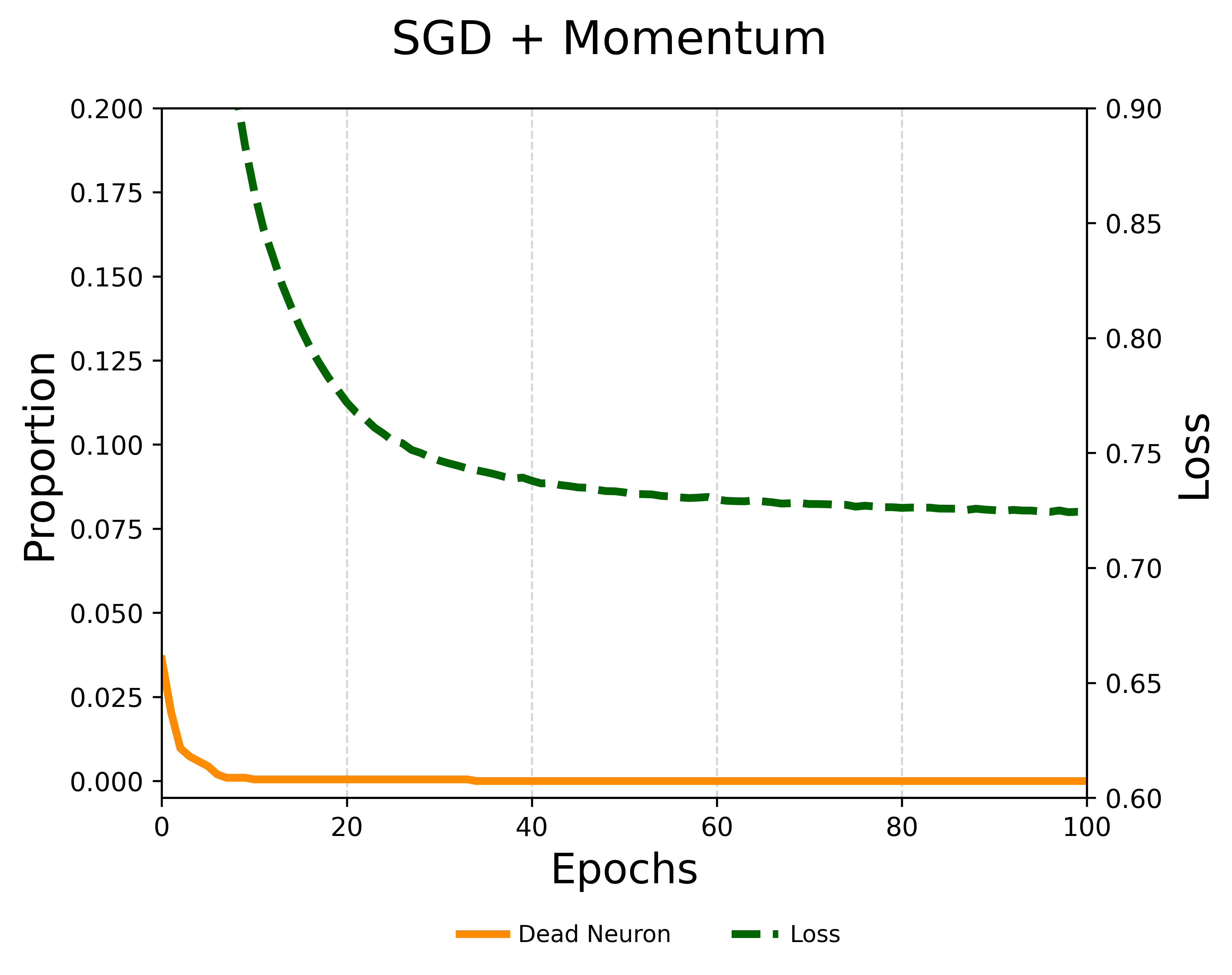}
		%		\subcaption{ }
		%		\label{fig:16-norm}
	\end{subfigure}
	\hspace{0.15cm}
	\begin{subfigure}[t]{0.22\linewidth}
		\centering
		\includegraphics[scale=.20]{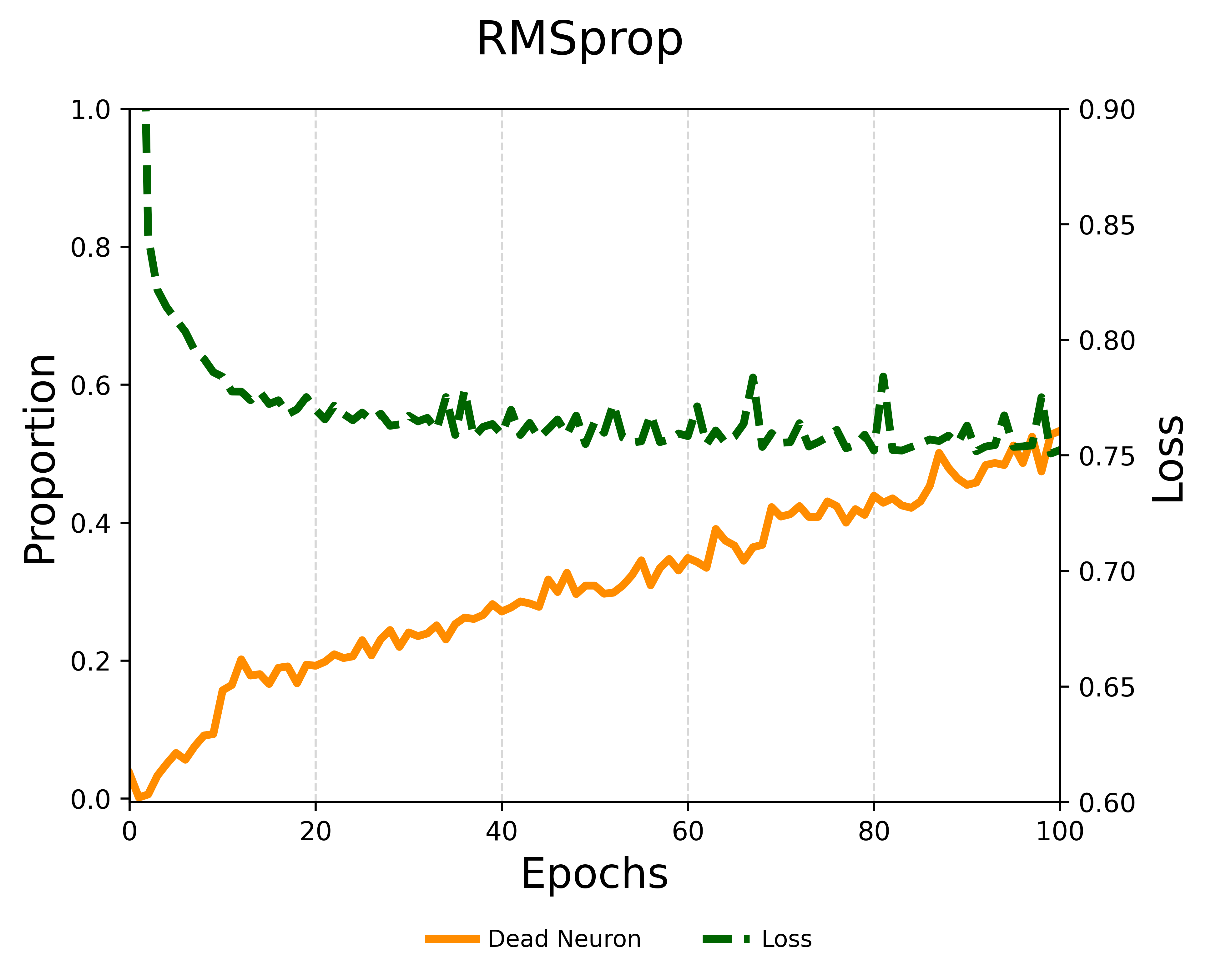}
		%		\subcaption{ }
		%		\label{fig:b}
	\end{subfigure}
	\hspace{0.15cm}
	\begin{subfigure}[t]{0.22\linewidth}
		\centering
		\includegraphics[scale=.20]{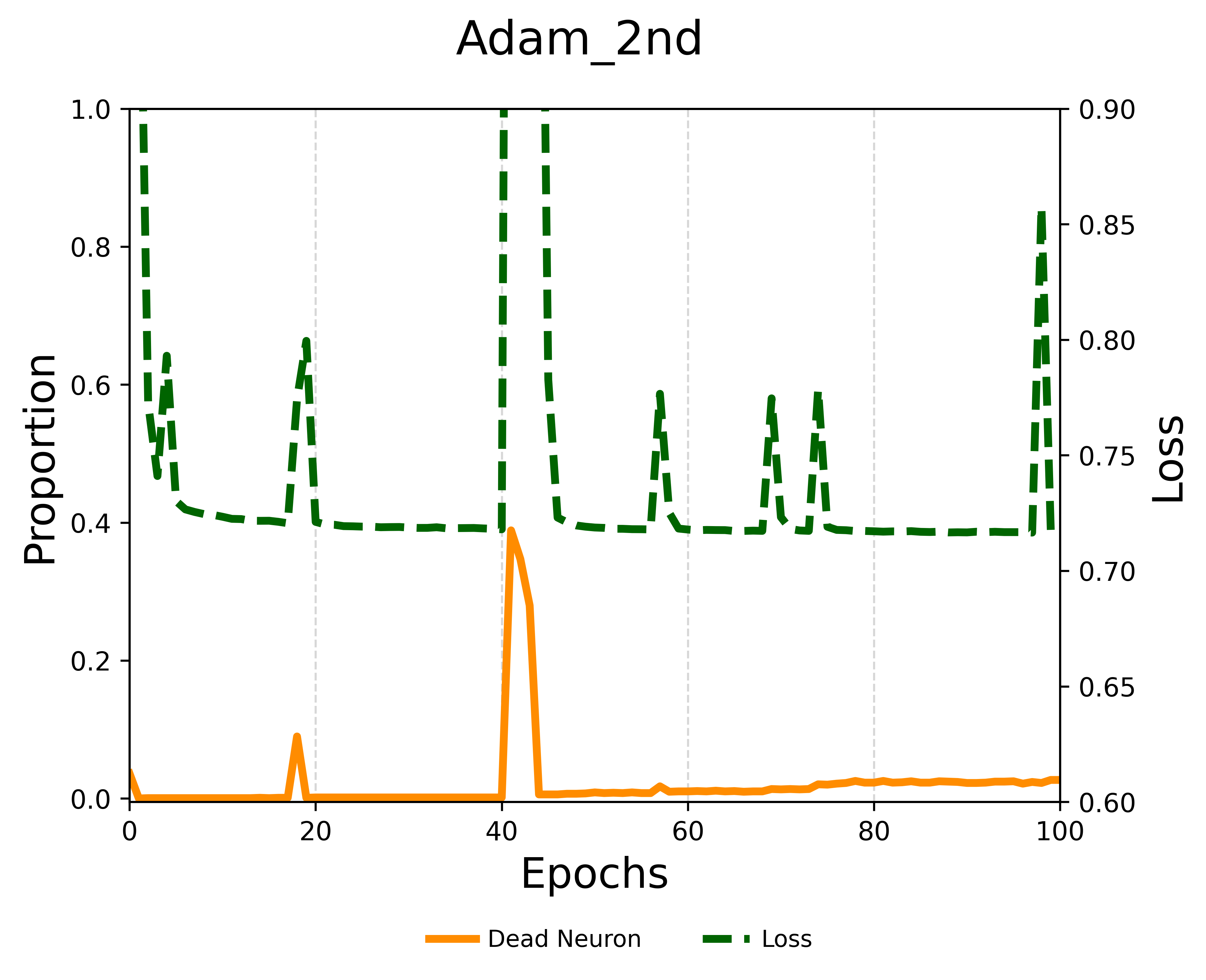}
		%		\subcaption{ }
		%		\label{fig:a}
	\end{subfigure}
	\hspace{0.15cm}
	\begin{subfigure}[t]{0.22\linewidth}
		\centering
		\includegraphics[scale=.20]{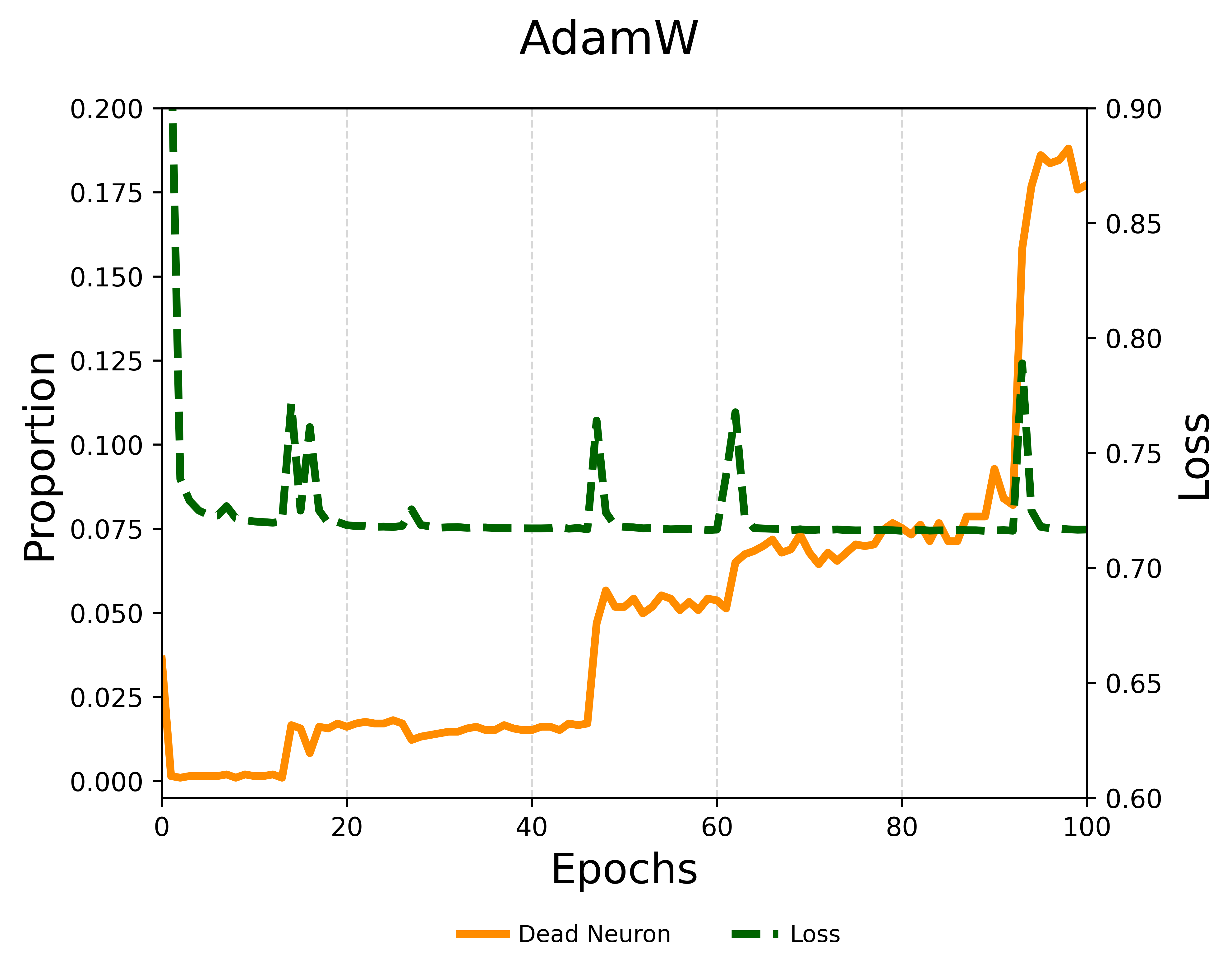}
		%		\subcaption{ }
		%		\label{fig:c}
	\end{subfigure}
	\vspace{-0cm}
	\caption{Loss and the proportion of dead neurons during training under different optimizers. {\bf SGD with momentum} does not cause loss spikes or an increase in the proportion of dead neurons. {\bf RMSprop} exhibits unstable training and a higher proportion of dead neurons. {\bf Adam primarily relying on second-order gradients} causes more severe loss spikes and greater fluctuations in the dead neurons proportion compared to regular Adam while {\bf AdamW} shows behavior similar to Adam.}
	\vspace*{-0cm}
	\label{fig:sec3-loss_spike}
\end{figure*}

{\bf Second-Order Gradient Information Plays a Crucial Role in Dynamic Sparsity:}
Previous work \citep{loss_spike_adam} theoretically analyze that the occurrence of loss spikes is related to some special state entered by the Adam optimizer. 
To further explore this, we compare different optimizers (noting that our default setting is Adam).
Noting that our default Adam setting are $\beta_1 = 0.9, \beta_2 = 0.999$, to align as closely possible with Adam and ensure training performance, we set the following optimizers:
(a) {SGD + Momentum} \citep{SGD} ($\beta_1 = 0.9$);
(b) {RMSprop} \citep{RMSprop} ($\beta_2 = 0.99$);
(c) {Adam\_2nd} ($\beta_1 = 0.01, \beta_2 = 0.999$);
(d) {AdamW} \citep{AdamW}.
Optimizer (a) mainly relies on first-order gradients, while (b) and (c) make more use of second-order gradient information.
Optimizer (d) is similar to Adam but applies weight decay directly to parameters.
The main results can be seen in Figure~\ref{fig:sec3-loss_spike}.
It can be observed that SGD + Momentum does not exhibit loss spikes or dynamic sparsity.
In contrast, optimizers dominated by second-order information (RMSprop and Adam\_2nd) exacerbate training instability and induce sparse patterns.
However, we also notice that the enhancement of sparsity in Adam\_2nd becomes unstable when lacking the guidance of first-order gradient momentum.
Additionally, AdamW exhibits a behavior similar to Adam.
Therefore, we give our conclusion as following:
\begin{tcolorbox}[observation]
	{\bf Observation 5~}  {\it 
	The dynamic sparsity in FFN modules is formed in a jump-like manner: during training, Transformers selectively discard part of the neurons in stages and re-optimize the remaining ones, which is strongly associated with to the appearance of loss spikes.
	Moreover, the triggering of this process is closely related to the second-order gradient information.}
\end{tcolorbox}

{\bf More discussions:} 
Although we have shown the correlation between dynamic sparsity in FFNs and loss spikes, it remains to be further explored how Transformers guide the gradual enhancement of this dynamic sparsity.
Additionally, it should be emphasized that FFN modules are not the sole factor contributing to loss spikes-such spikes also occur in Attention-only models (see Figure~\ref{app:fig:entropy_loss_attention_ffn}). 
In addition, we also find that as the model size increases, Transformers are more prone to loss spikes compared to RNNs (see Figure~\ref{app:fig:entropy_loss_comparison}).
These findings suggest that the occurrence of loss spikes is related to not only the second-order information of the optimizer, but also the special model architecture.
We leave these aspects for our future work.

%{\bf Regarding Loss Spikes:} During training process of our controlled experiments, we also observe the occurrence of loss spikes, a phenomenon that also appears in real large-scale model pretraining\citep{ loss_spike_phenomenon,loss_spike_adam}. 
%We find that as the model size increases, Transformers are more prone to loss spikes compared to RNNs. 
%On the other hand, when a loss spike occurs, the KL divergence does not necessarily exhibit a significant abrupt change (see Figure~\ref{fig:loss_spike} in Appendix).
%We leave the investigation of this for future work.

%\newpage

\section{Conclusion}
In this paper, we examine Transformers from two key compression aspects: data compression and dynamic sparsity. We find that beyond approximating the target distribution, Transformers tend to explore low-entropy landscapes, with the FFN module facilitating this process. 
Additionally, beyond eliminating redundancy in the data, Transformers also exhibit parameter redundancy, where larger models favor residual connections to bypass attention computations and have a lower proportion of active neurons.
We also find that loss spikes in training instability of larger models is linked to the sudden decrease in neuronal activity, with second-order gradient information exacerbating this effect.

\bibliographystyle{plainnat}
\bibliography{example_paper}

%%%%%%%%%%%%%%%%%%%%%%%%%%%%%%%%%%%%%%%%%%%%%%%%%%%%%%%%%%%%

\appendix

\newpage
\section{Related Work}
\textbf{Data Compression and Transformers:}
Recently, data compression has emerged as an important perspective for understanding the success behind Transformer-based LLMs.
\citet{LMisCompression} establish a connection between the maximum likelihood training objective of LLMs and arithmetic coding, proposing that LLMs are powerful lossless compressors. 
%Furthermore, they empirically verify that foundation models can serve as general-purpose compressors.
Moreover, a prevailing belief is that compression can lead to intelligence \citep{hutterprize,Ilyatalk2023,pan2025understanding}. 
To quantitatively explore this, \citet{CompressionIntelligenceLinearly} take knowledge and commonsense, coding, and mathematical reasoning as indicators of intelligence and observe the strong linear relationship between compression performance and downstream task performance.
%evaluate a wide range of LLMs on corresponding datasets. 
Previous works focus more on evaluating compression quality by measuring the discrepancy between the model’s learned distribution and the target distribution. 
In contrast, our work lifts the constraint of the target distribution and instead focuses on the variation in the information content (entropy) of the learned distribution itself.
Another line of work  orthogonal to our study is those focusing more on understanding the architecture of Transformers from a compression perspective \citep{whiteboxTF, whiteboxTF2, attention-only}, whose core idea is to establish an equivalence between the forward process of Transformers and iterative algorithms related to data compression objective. \citet{whiteboxTF,whiteboxTF2} view Transformers as an iterative process toward a objective called sparse rate reduction and propose an interpretable, white-box Transformer architecture. 
\citet{attention-only} interprets attention layers as compressing noisy token representations into low-dimensional subspaces.

\textbf{Mechanisms of Attention and FFN Modules:}
Previous work advances our understanding of the interplay between attention modules and residual connections. 
\citet{attention_rank_collapse} show that stacking pure self-attention layers will lead to the collapse of output representations into a rank-1 matrix, while residual connections effectively prevent this. 
\citet{attention_collapse_signal} analyze rank collapse from the perspective of signal propagation and propose mitigating it by scaling the residual branch which are adopted by \citet{BN_skip, no_residual, simplify_TF}. 
Following \citet{attention_rank_collapse}, our work views each attention head or skip connection as a path and introduces a routing mechanism to explore the dynamic sparsity of attention modules.
For the FFN module, \citet{FFN_key_value} interpret the feed-forward layer as a key-value store and empirically show that the keys capture input patterns while the values contribute to the output distribution\citep{ffn_predict}. The work most closely related to ours is \citet{neurons_dead}, which investigates neuron activations in the FFNs of the OPT\citep{opt} family by collecting a large corpus, finding that many neurons in the early part of models are inactive (dead).
However, despite the large-scale test corpus, it remains unclear whether such sparsity pattern holds on all underlying unseen inputs, and how the pattern forms is still an open question. 
Our work further explore these problems under a controlled experimental setup.

\textbf{Dynamic Sparsity of Transformers:}
Dynamic sparsity is widely present in the structure of foundation models\citep{tutorial_dynamic_sparse}, including attention computation\citep{ent15max, adaptive_entmax, oren2024transformers}, MoE routing \citep{switchTF, artetxe2021efficient, MoE_ScalingLaw}, and sparse layers\citep{Depth-adaptiveTF, MoD}, among others. For the attention module, previous work primarily focuses on sparsity in the attention map, that is, a token attends to only a small set of key tokens.
However, inspired by \citet{attention_rank_collapse}, our work focuses on sparsity in the path selection between attention heads and residual connections.
As for the FFN module, current foundation models often adopt the MoE structure \citep{switchTF, survey_MoE}, where different inputs typically activate only a limited number of experts. 
Thus, studies on dynamic sparsity are usually conducted at the granularity of expert networks. 
In contrast to MoE-based approaches, our work emphasizes the dynamic sparsity within the FFN module itself, exploring sparsity at the granularity of individual neurons similar to \citet{neurons_dead}.

\section{More details of Experiment Setting}\label{app:base setting}
{\bf More details of data generating:}
To trigger the model to output the probability of the first character, we introduce a special symbol "\#" as the start of the sequence, just similar to <BOS>.
We define $\vs_0 =$"\#" and $\vp_{\vs_0} = 1$.
Then, we sample from a uniform distribution to obtain $\vp(\vs_1|\vs_0)$ for generating the first character.
When generating the subsequent conditional probabilities $\vp(\vs_i|\vs_{<i})$, we follow the sparsity principle: that is, we randomly select two characters from the vocabulary and assign them probabilities of 0.8 and 0.2 respectively, while the probabilities of other characters are all set to 0.
Additionally, it should be noted that the complexity of calculating the learned $\ptheta$ is exponential, that is, $|\gV|^n$. 
Therefore, the vocabulary size and sequence length cannot be set too large. 
In the main body of this paper, we fix the vocabulary size $|\gV| = 5$ (not including "\#") and the sequence length $n=5$, which results in a total of $ |\gV|^n = 3125 $ possible sequences. 
Then, under the approach of the sparsity principle, only $ |\gV| \cdot 2^{n - 1} =80 $ sequences will have non-zero probabilities. 
These sequences form the non-sparse part, while the remaining constitute the sparse part as in Section \ref{sec:low-entropy}.
It should be noted that there may be slight differences between the sampled true distribution and the generated target distribution. 
Therefore, to better compare with the real distribution, we use the sampled distribution as the target distribution $\ptgt$.
In our calculations of entropy $\htgt$ and KL divergence $\kltheta$, the target distributions all refer to this sampled distribution.
In the main body, we have $\htgt = 3.571$.

In addition to generating the target distribution as described above, we also test different distributions by adjusting the values of conditional probabilities: 
{\bf (a) Lower-entropy distribution:} we still sample from a uniform distribution to generate $\vp(\vs_1|\vs_0)$ while when generating subsequent conditional distributions, we randomly select two characters and assigned them probabilities of 0.9 and 0.1 while the remaining are set to 0, resulting in $\ptgt = 2.864$;
{\bf (b)  Higher-entropy distribution:} we randomly selected three characters and assigned them probabilities of 0.6, 0.3, and 0.1 respectively, which ultimately resulted in $\ptgt = 5.160$.
Our conclusions still hold for these cases. 
The experimental results can be found in Appendix \ref{app:dist91} and \ref{app:dist631} respectively.

{\bf More details of model settings:} 
We select two classic RNN models: GRU \citep{GRU} and LSTM \citep{LSTM}, both implemented using PyTorch's default modules (torch.nn.GRU and torch.nn.LSTM).
For RNNs, we find that a single layer is sufficient while stacking too many layers may lead to optimization issues under our controlled experimental setup.
Therefore, we use a single-layer configuration for all RNNs by default.
In addition, the hidden size is set to be $d_h = 4d$.
For Transformers, we use the decoder-only architecture similar in \citet{vaswani2017attention}.
More specifically, we use the cosine absolute position encoding and add it to the token embeddings.
In the attention module, we fix the number of attention heads at 4.
In the FFN module, we use ReLU as the activation function, and the hidden size $d_h$ is set to four times the size of model dimension, that is, $d_h = 4d$.
In both the attention module and the FFN module, we apply layer normalization after the residual connection.
The input size to the embedding layer is $|\gV| + 1$ as there is an additional symbol "\#" while for the final projection head, the output size of the linear layer is exactly $|\gV|$, which is followed by Softmax to output the probabilities over $|\gV|$ characters.

When a comparison is needed, noting that the scale of the task is relatively small, the embedding size is chosen from $\{8,16,32,64\}$ meanwhile we adjust the number of layers of Transformer ($L = 5$) to make the model size comparable to RNNs under the same model dimension $d$ (in center left and center right part of Figure~\ref{fig:sec2-comp} and Figure~\ref{fig:sec2-part} in section~\ref{sec:low-entropy}.
In addition, when more model settings are required, we change the model dimension and the number of layers of the model.
More specifically, in the right part of Figure~\ref{fig:sec2-comp}, for RNNs, the model dimension $d$ is selected from $\{16, 20, 24, 28, 32, 64\}$ while for Transformers, the model dimension $d$ is selected from $\{16,32,64\}$ and the number of layers is from $\{3,4,5,6,8\}$.
The settings of Transformer variants in the center left and right parts of Figure \ref{fig:sec2-ablation} also follow the above.
The cases $d = 8$ are all excluded above for clear presentation as the learned distributions are relatively far from the target distribution when models are small and these scatter points often become outliers relative to the main region.
This does not affect our conclusions because the low-entropy preference tends to gradually emerge in larger Transformers.
However, when the model is small, there is still a strong linear relationship between KL and Loss. Therefore, we retained the case of $d = 8$ in the right part of Figure~\ref{fig:sec2-ablation}.
Moreover, in Section~\ref{sec:attention} and \ref{sec:FFN}, we set the layers $L = 8$ to better observe the role of residual connections and ensure a sufficient number of FFN neurons. 
However, we find that our conclusions still hold when $L = 5$.

{\bf More details of training settings:}
For all models, we sample 65,536 sequences $\vs$ from the generated target distribution $\ptgt$ as our training data. 
During training, we set the batch size to 512 and set the dropout rate to 0.1 for RNNs and the FFN modules in Transformers.
All models are trained for 100 epochs using Adam \citep{Adam}, where we set the learning rate $\mathrm{lr} = 0.001$, $\beta_1 = 0.9$ and $\beta_2 = 0.999$.
As for settings of Figure~\ref{fig:sec3-loss_spike} in Section \ref{sec:FFN}, the parameters of each optimizer are as follows: (a) SGD + Momentum: lr = 0.001 and Momentum = 0.9; (b) RMSprop: lr = 0.0001, $\alpha$($\beta_2$) = 0.99 and weight decay = 0.01; (c) Adam\_2nd: lr = 0.0005, $\beta_1$ = 0.01 and $\beta_2$ = 0.999; (d) AdamW: lr = 0.001, $\beta_1$ = 0.01, $\beta_2$ = 0.999 and weight decay = 0.01. 
We observe that during training, entropy and KL exhibit more pronounced fluctuations compared to loss.
To more clearly present the trends and the relative levels between different models, we smooth the curves of entropy and KL changes during training by taking an average with a window size of 3. 
However, to better illustrate the occurrence of loss spikes, we do not apply similar smoothing to the loss curve. 
Additionally, when drawing scatter plots in Figure~\ref{fig:sec2-comp} and \ref{fig:sec2-ablation}, for each model configuration, we average the values of the last 15 epochs where we exclude outliers caused by the occurrence of loss spikes especially for larger models.

%\newpage
\section{More Experiment Details}\label{app:FFN setting}

\begin{figure*}[!htbp]
	\centering
	\begin{subfigure}[t]{0.22\linewidth}
		\centering
		\includegraphics[scale=.20]{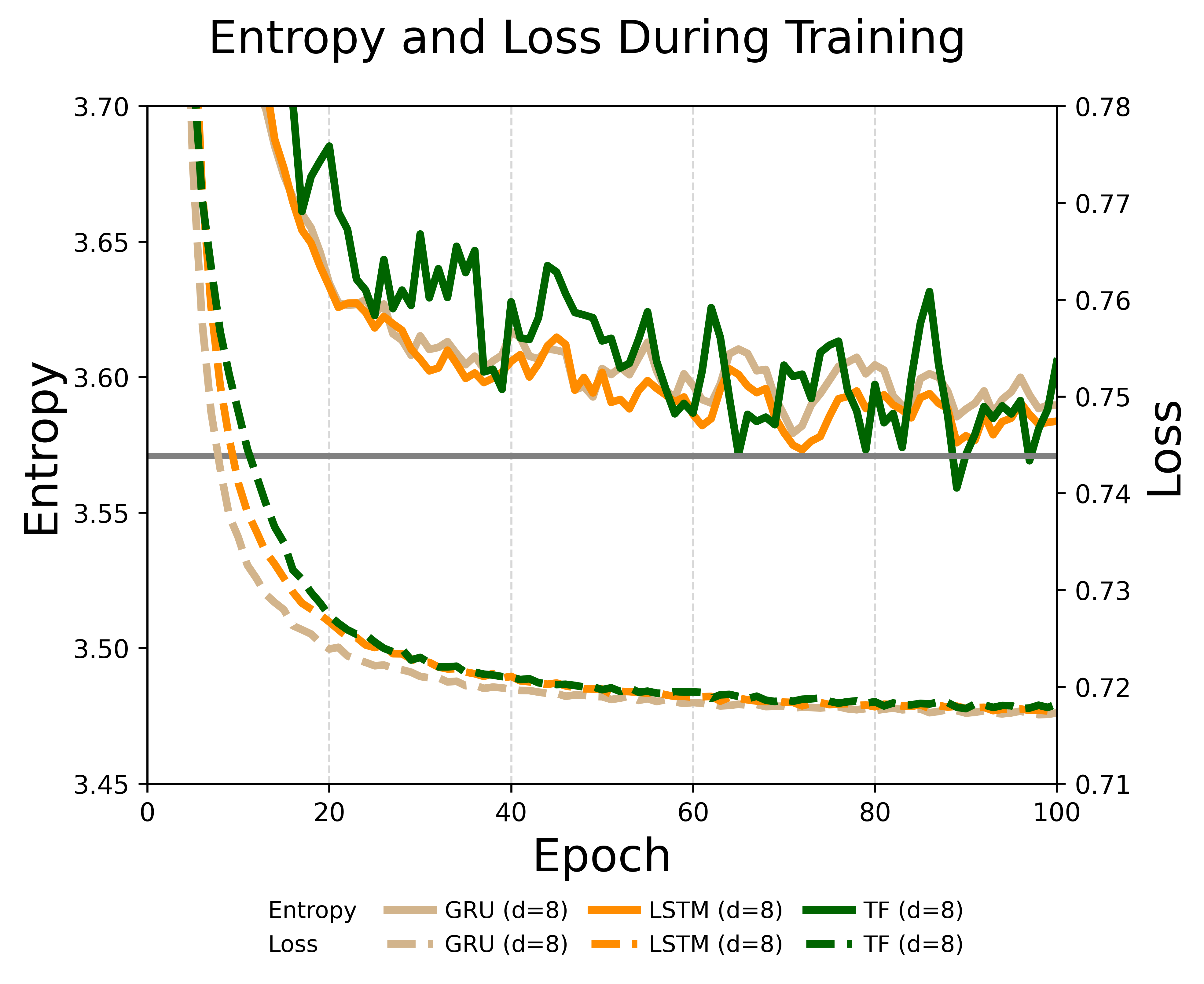}
		%		\subcaption{ }
		%		\label{fig:a}
	\end{subfigure}
	\hspace{0.05cm}
	\begin{subfigure}[t]{0.22\linewidth}
		\centering
		\includegraphics[scale=.20]{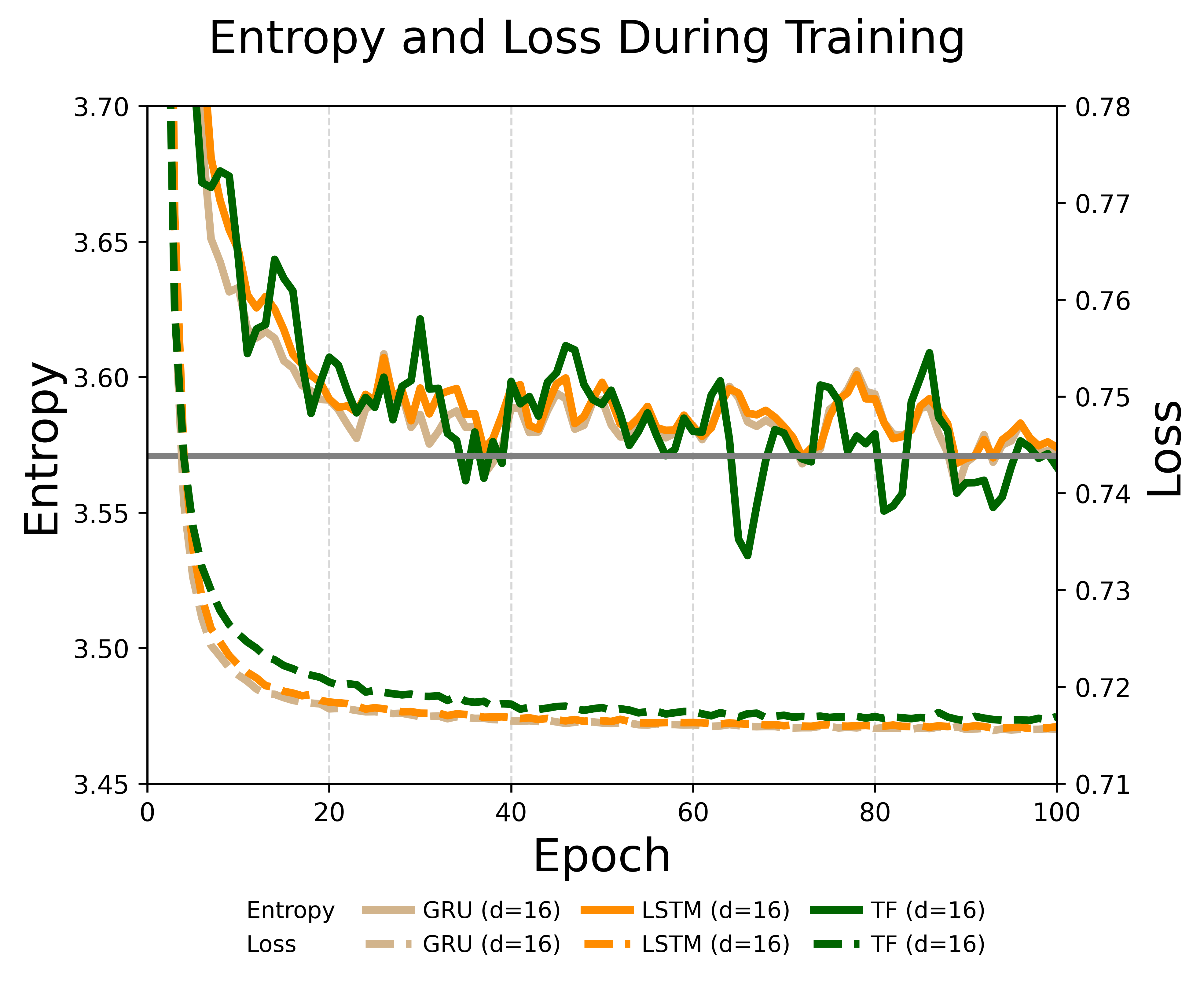}
		%		\subcaption{ }
		%		\label{fig:b}
	\end{subfigure}
	\hspace{0.05cm}
	\begin{subfigure}[t]{0.22\linewidth}
		\centering
		\includegraphics[scale=.20]{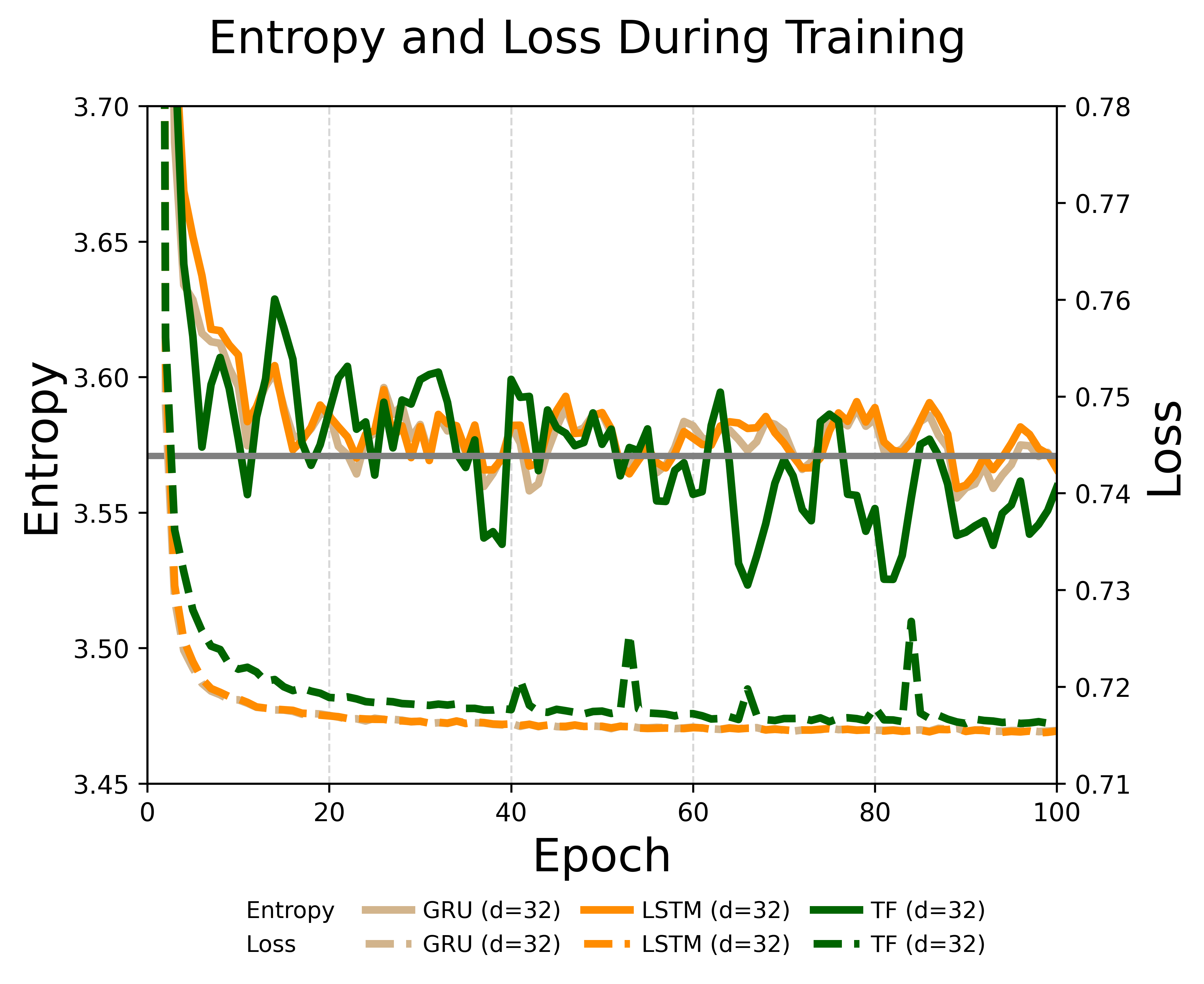}
		%		\subcaption{ }
		%		\label{fig:c}
	\end{subfigure}
	\hspace{0.05cm}
	\begin{subfigure}[t]{0.22\linewidth}
		\centering
		\includegraphics[scale=.20]{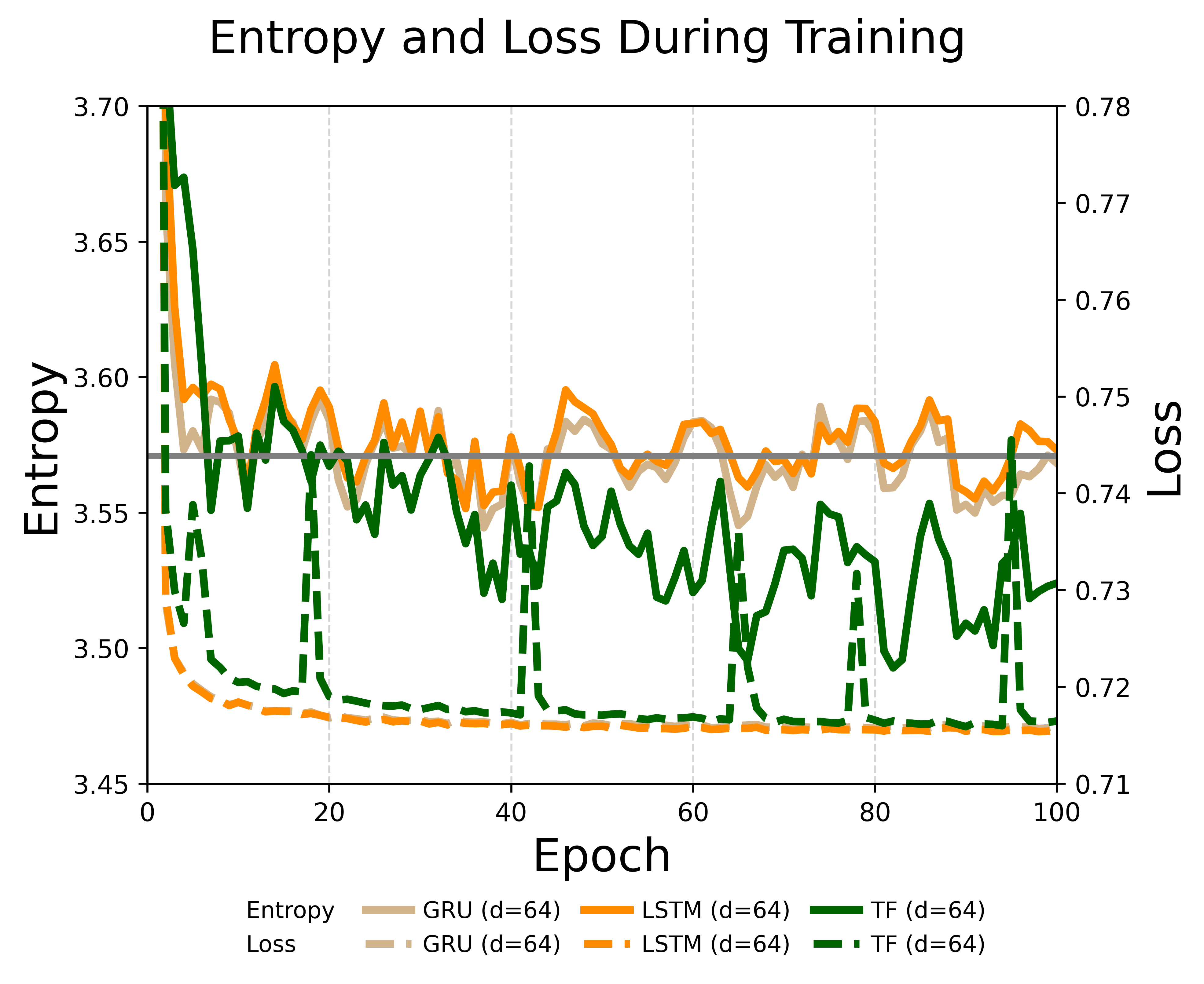}
		%		\subcaption{ }
		%		\label{fig:a}
	\end{subfigure}
	\vspace{-0cm}
	\caption{Entropy and KL during training for GRU, LSTM and Transformer when $d = 8,16,32,64$.}
	\vspace{-0cm}
	\label{app:fig:entropy_kl_comparison}
\end{figure*}

\begin{figure*}[!htbp]
	\centering
	\begin{subfigure}[t]{0.22\linewidth}
		\centering
		\includegraphics[scale=.20]{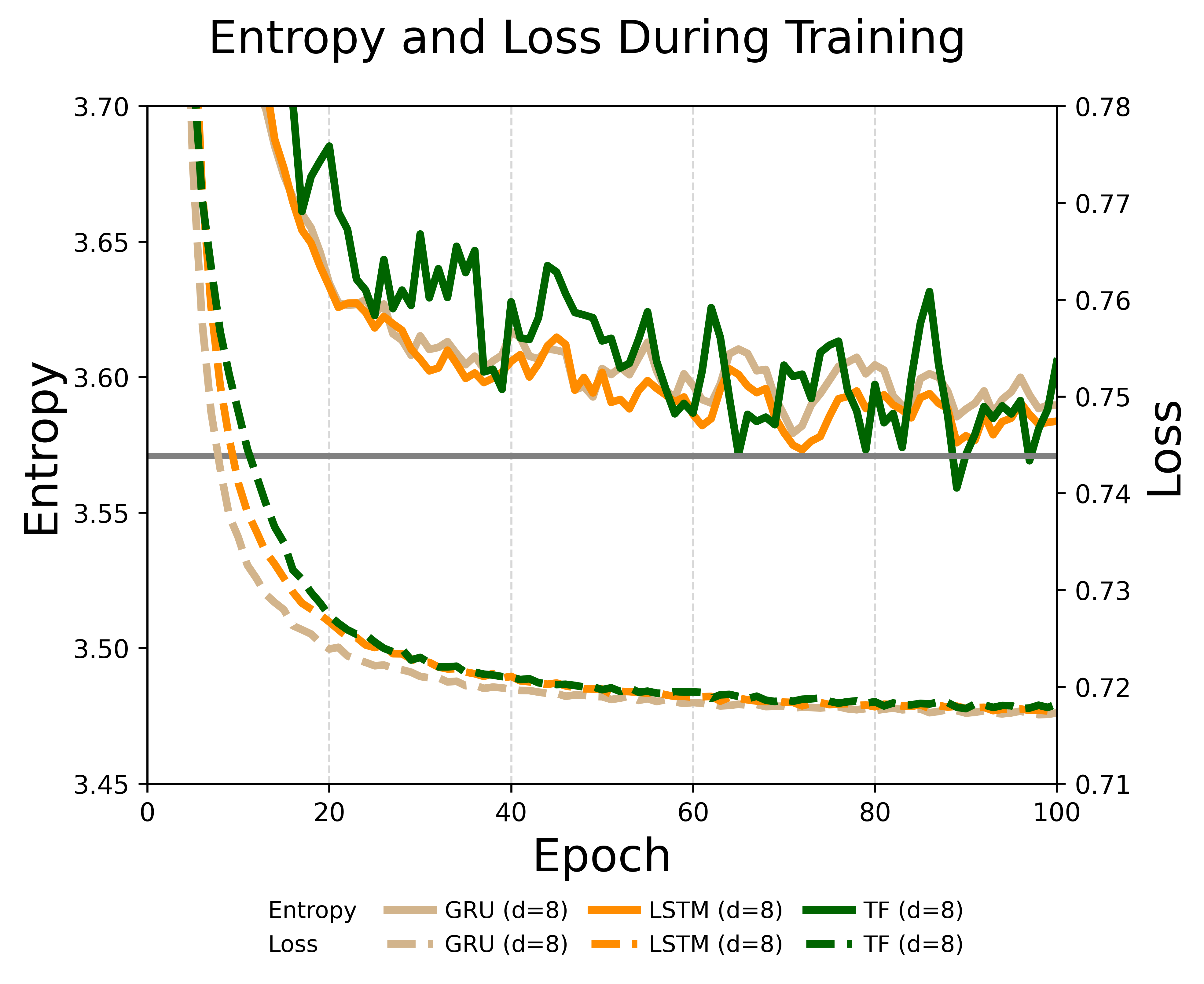}
		%		\subcaption{ }
		%		\label{fig:a}
	\end{subfigure}
	\hspace{0.05cm}
	\begin{subfigure}[t]{0.22\linewidth}
		\centering
		\includegraphics[scale=.20]{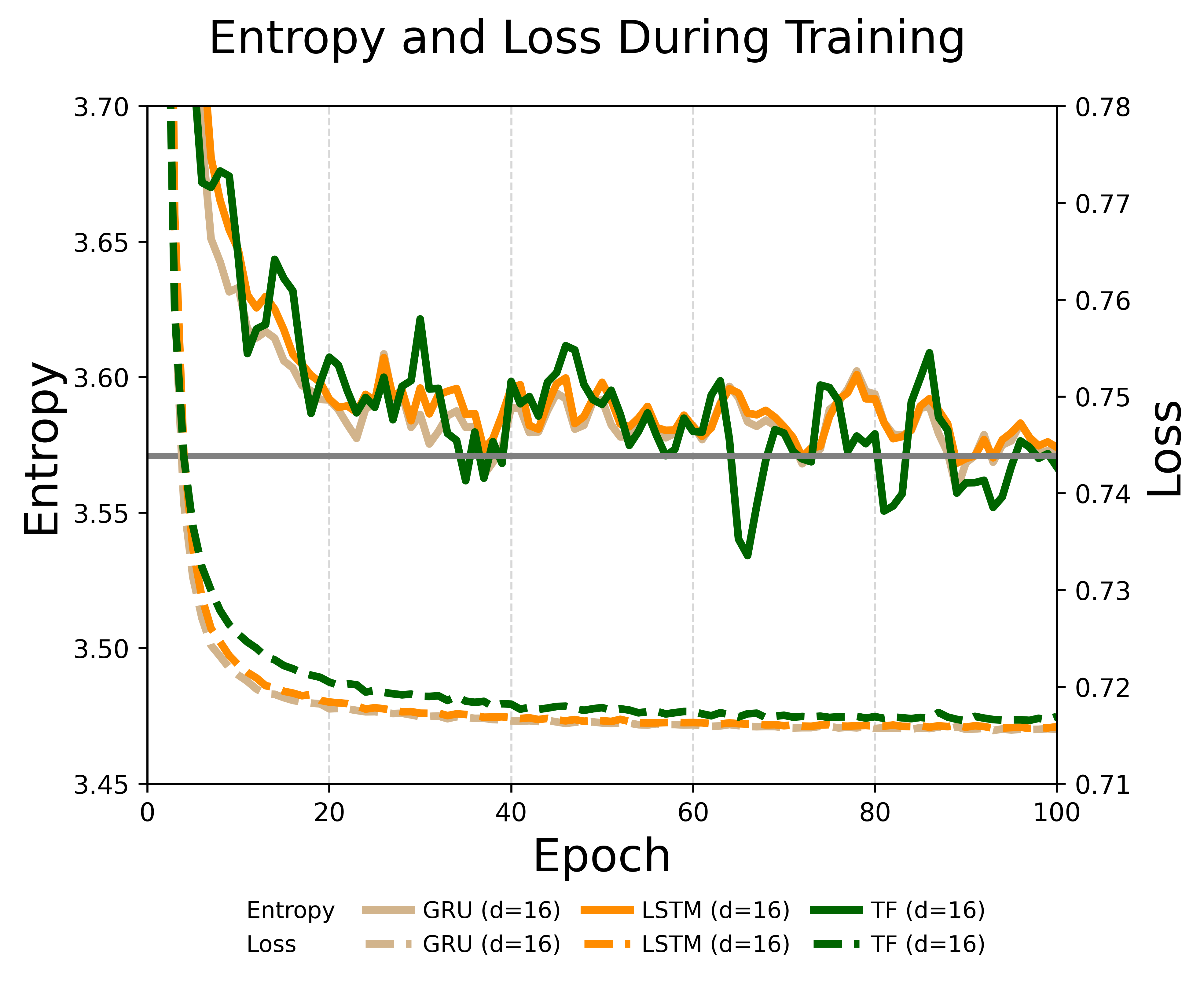}
		%		\subcaption{ }
		%		\label{fig:b}
	\end{subfigure}
	\hspace{0.05cm}
	\begin{subfigure}[t]{0.22\linewidth}
		\centering
		\includegraphics[scale=.20]{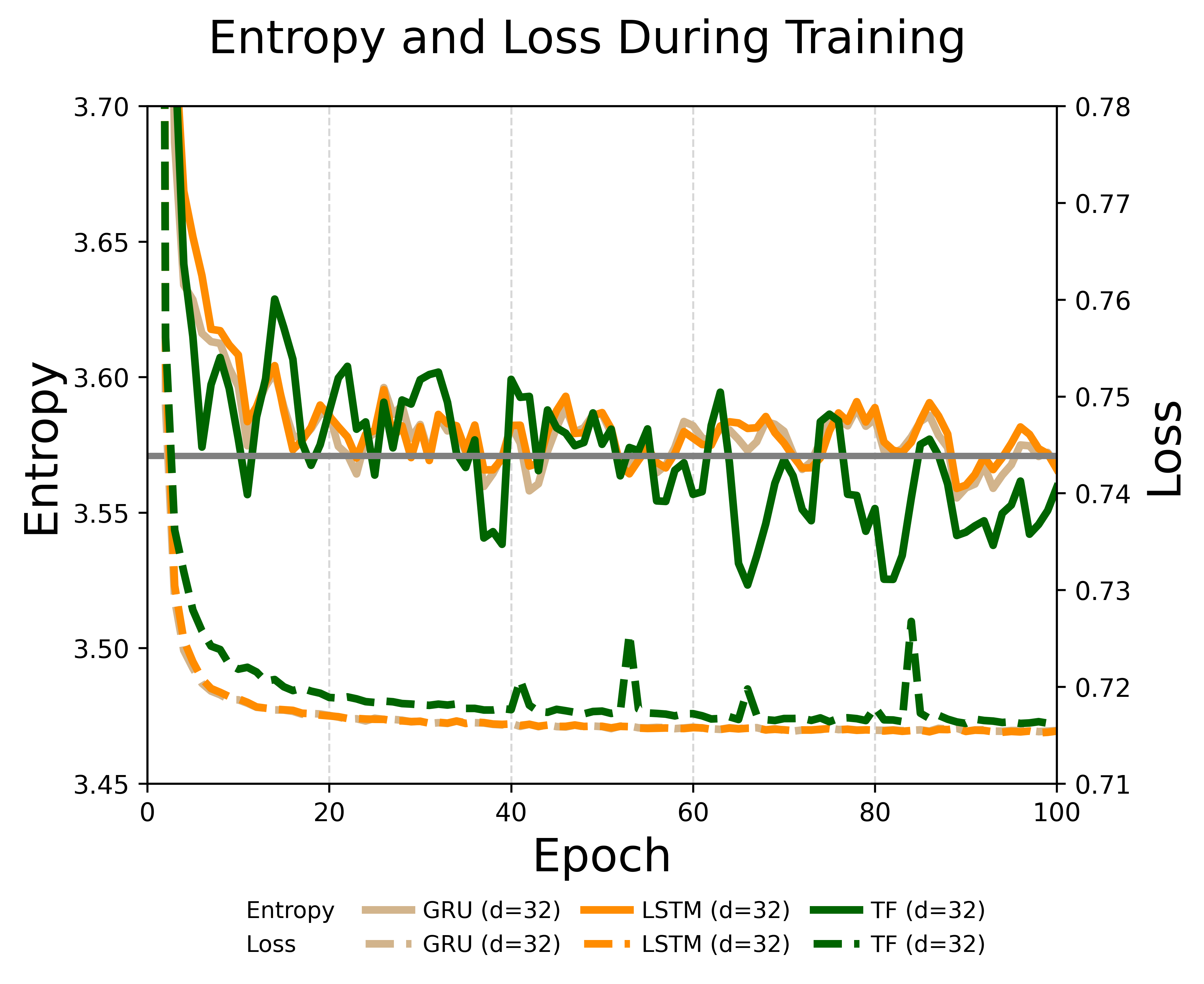}
		%		\subcaption{ }
		%		\label{fig:c}
	\end{subfigure}
	\hspace{0.05cm}
	\begin{subfigure}[t]{0.22\linewidth}
		\centering
		\includegraphics[scale=.20]{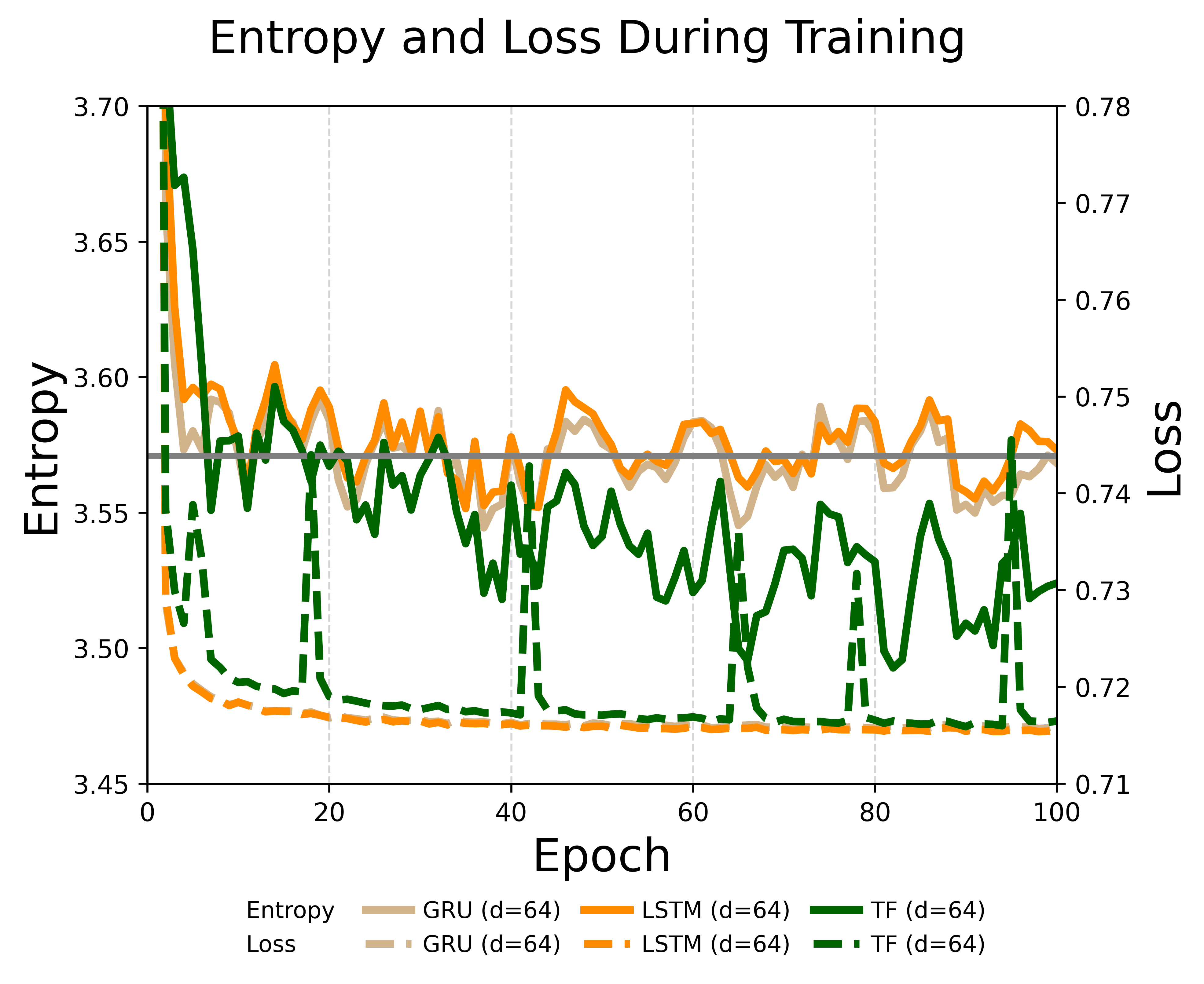}
		%		\subcaption{ }
		%		\label{fig:a}
	\end{subfigure}
	\vspace{-0cm}
	\caption{Entropy and Loss during training for GRU, LSTM and Transformer when $d = 8,16,32,64$.}
	\label{app:fig:entropy_loss_comparison}
	\vspace{-0cm}
\end{figure*}

\begin{figure*}[!htbp]
	\centering
	\begin{subfigure}[t]{0.22\linewidth}
		\centering
		\includegraphics[scale=.20]{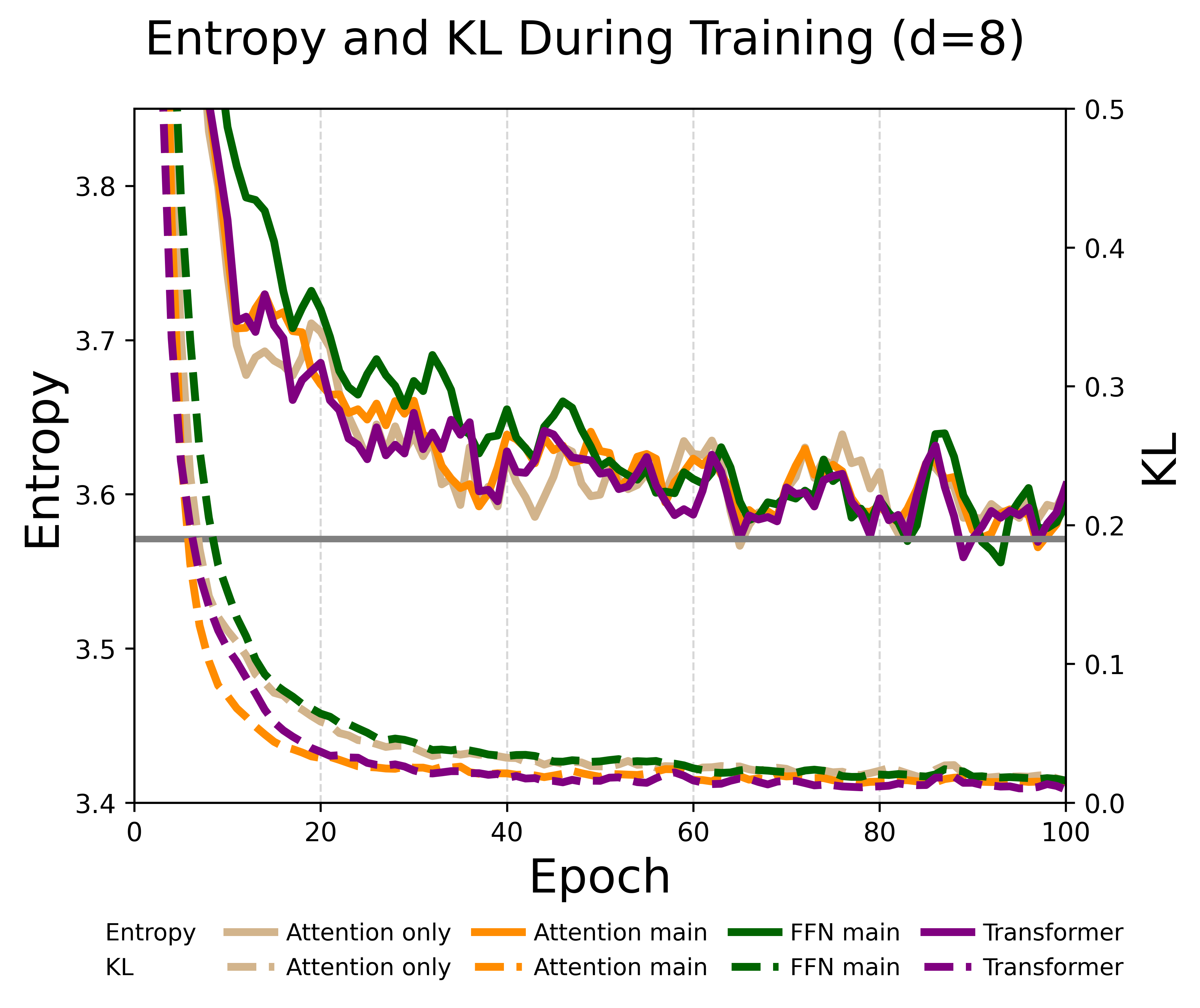}
		%		\subcaption{ }
		%		\label{fig:16-norm}
	\end{subfigure}
	\hspace{0.05cm}
	\begin{subfigure}[t]{0.22\linewidth}
		\centering
		\includegraphics[scale=.20]{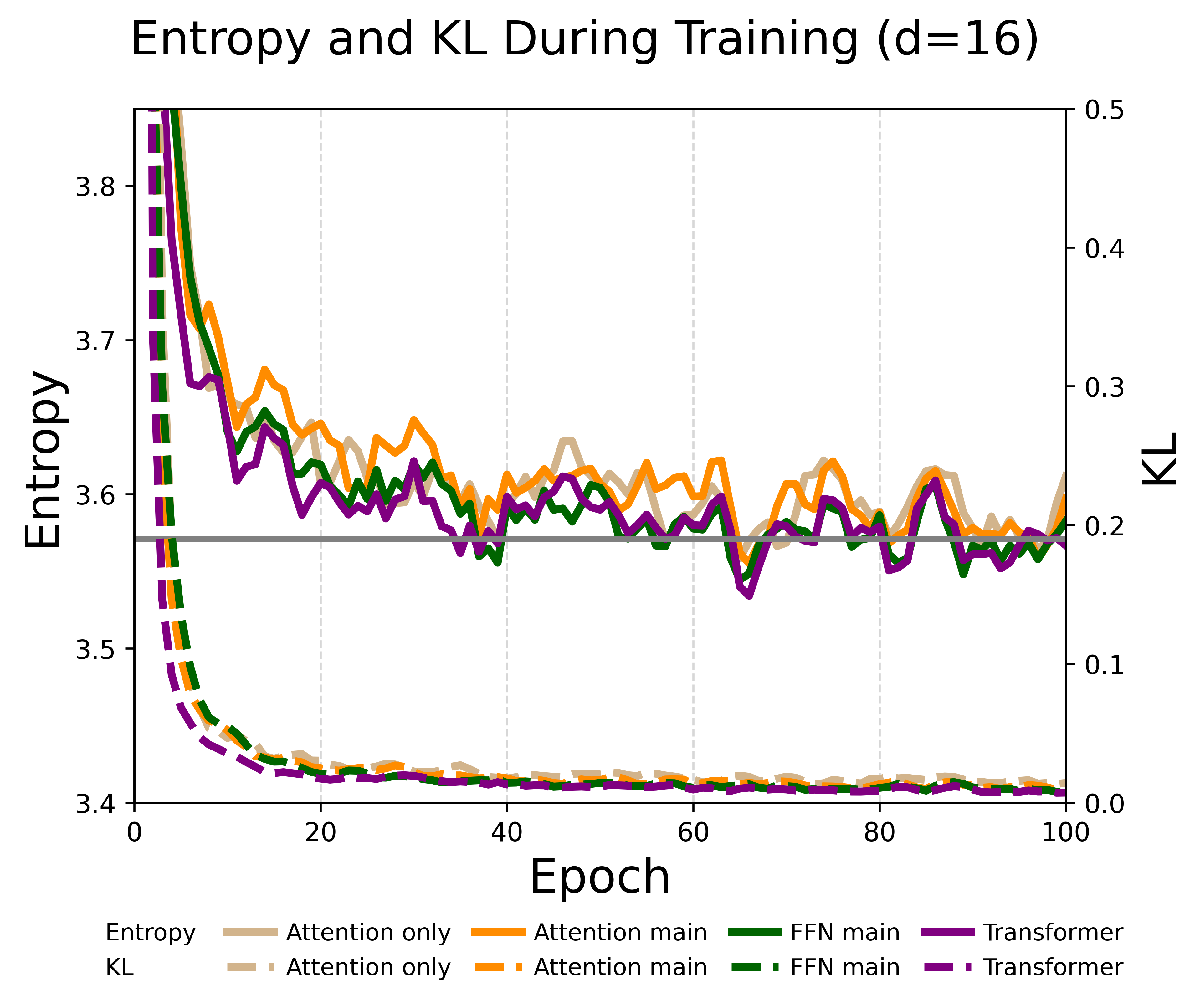}
		%		\subcaption{ }
		%		\label{fig:a}
	\end{subfigure}
	\hspace{0.05cm}
	\begin{subfigure}[t]{0.22\linewidth}
		\centering
		\includegraphics[scale=.20]{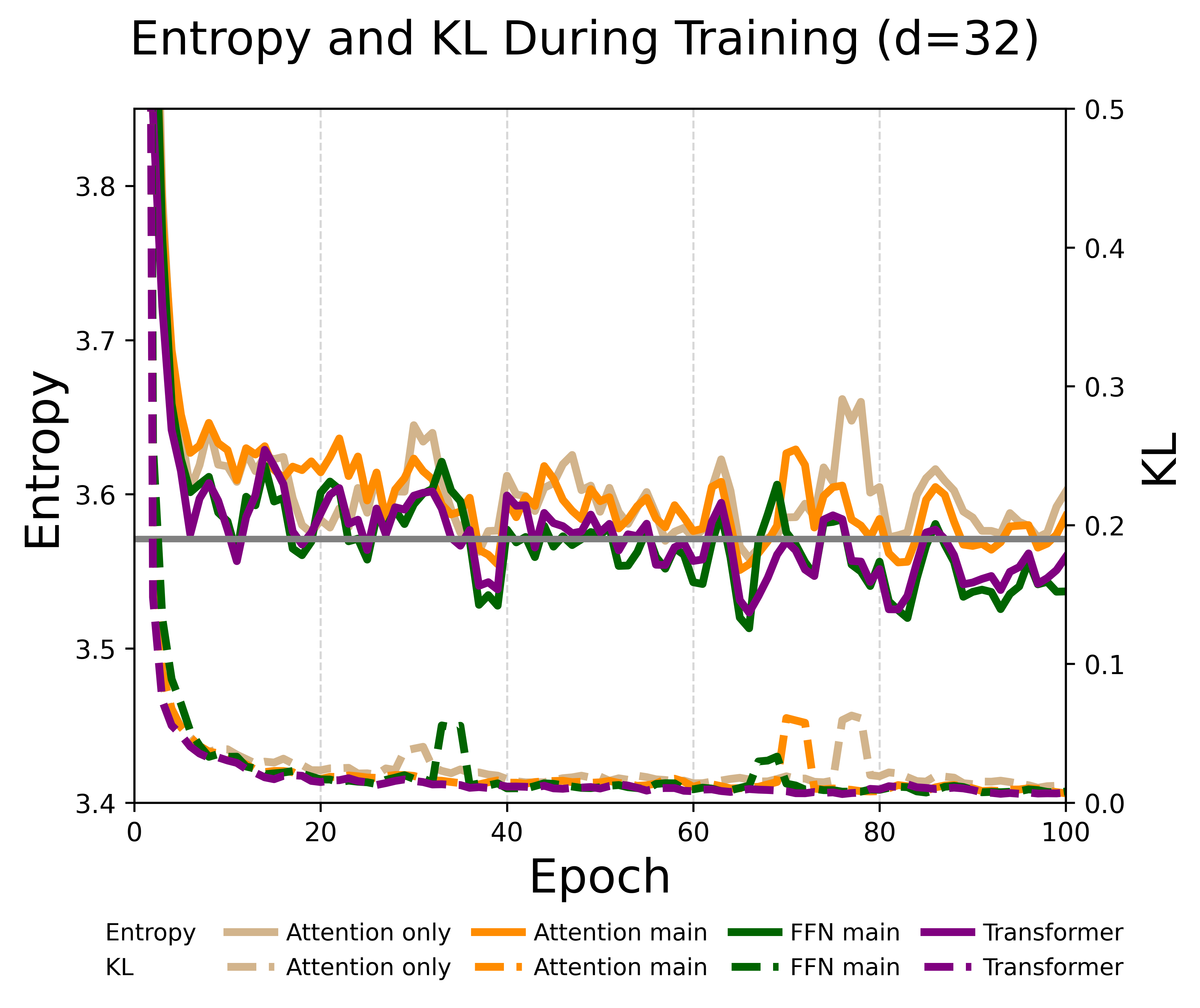}
		%		\subcaption{ }
		%		\label{fig:b}
	\end{subfigure}
	\hspace{0.05cm}
	\begin{subfigure}[t]{0.22\linewidth}
		\centering
		\includegraphics[scale=.20]{pics/Entropy_KL_Attention_FFN_TF_64.png}
		%		\subcaption{ }
		%		\label{fig:c}
	\end{subfigure}
	\vspace{-0cm}
	\caption{Entropy and KL during training for different Transformer variants when $d = 8,16,32,64$.}
	\vspace{-0cm}
	\label{app:fig:entropy_kl_attention_ffn}
\end{figure*}

\begin{figure*}[!htbp]
	\centering
	\begin{subfigure}[t]{0.22\linewidth}
		\centering
		\includegraphics[scale=.20]{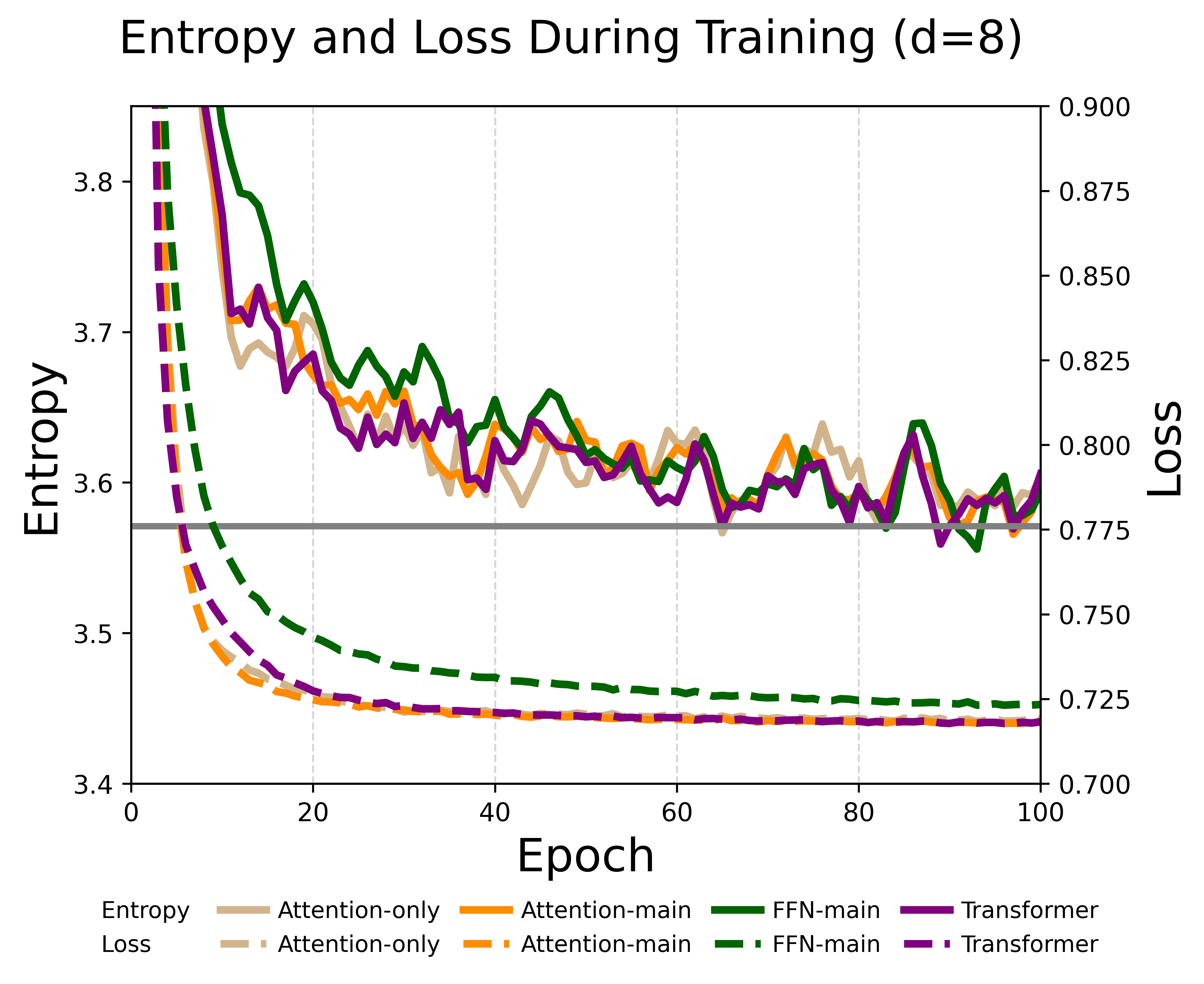}
		%		\subcaption{ }
		%		\label{fig:16-norm}
	\end{subfigure}
	\hspace{0.05cm}
	\begin{subfigure}[t]{0.22\linewidth}
		\centering
		\includegraphics[scale=.20]{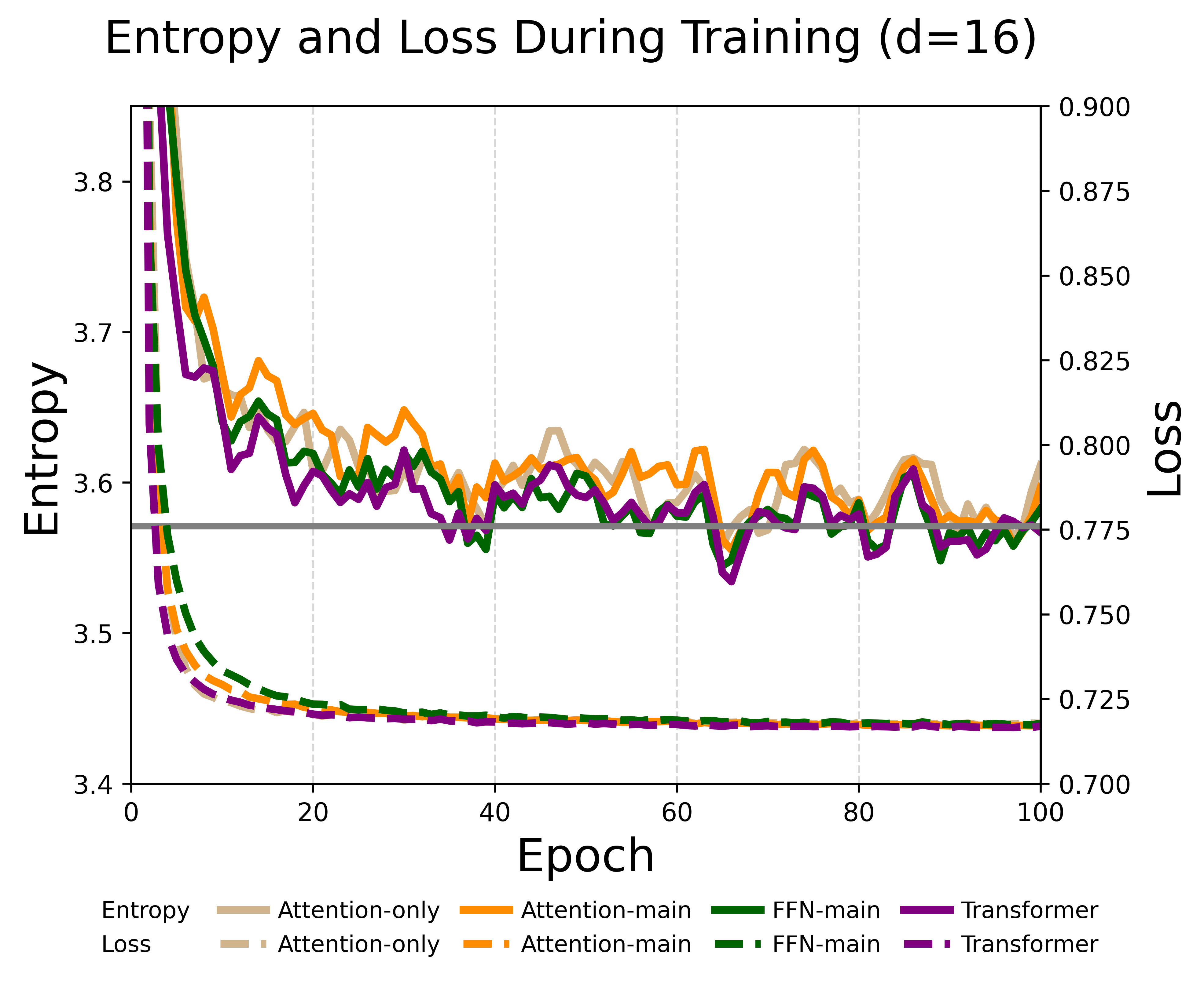}
		%		\subcaption{ }
		%		\label{fig:a}
	\end{subfigure}
	\hspace{0.05cm}
	\begin{subfigure}[t]{0.22\linewidth}
		\centering
		\includegraphics[scale=.20]{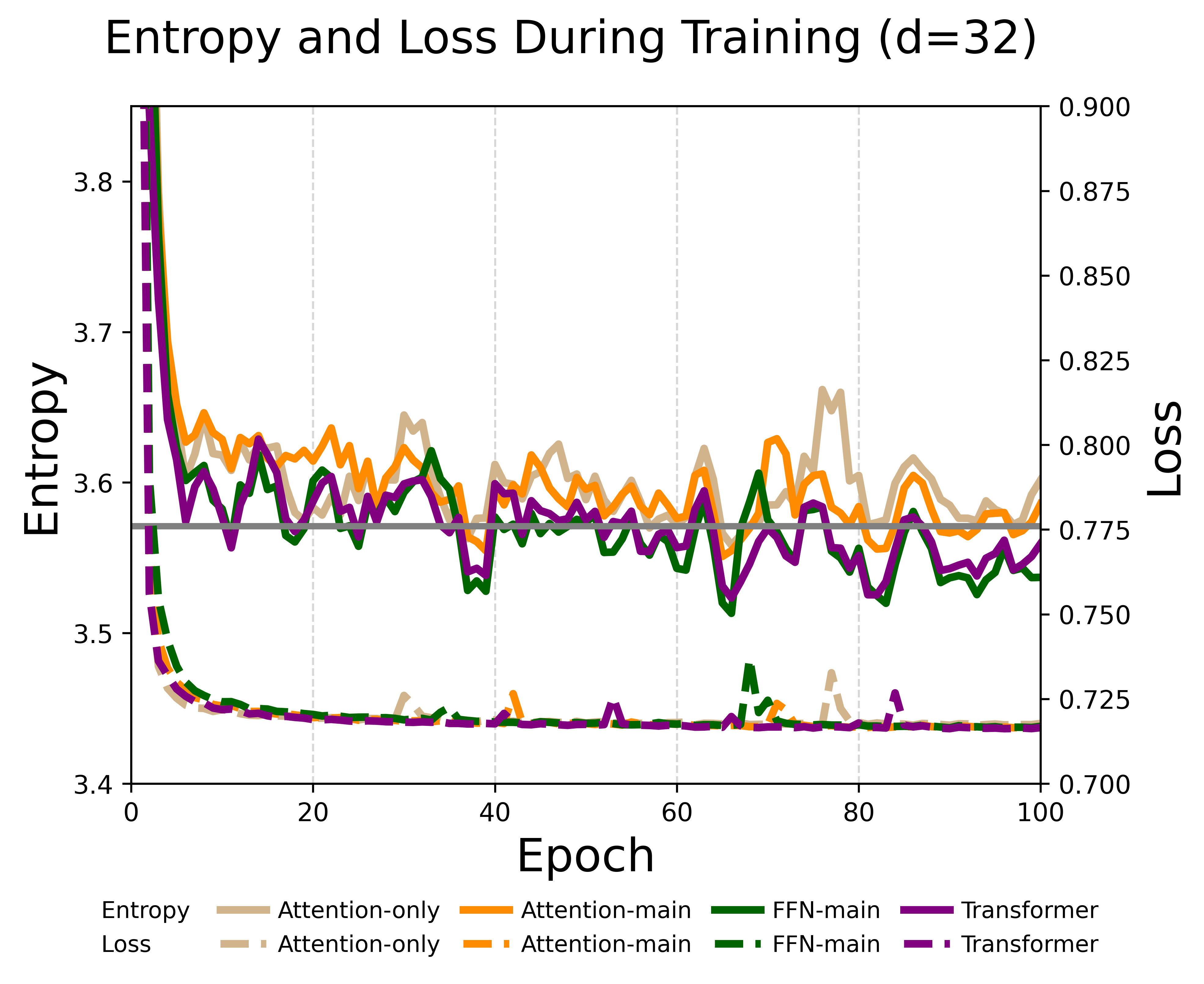}
		%		\subcaption{ }
		%		\label{fig:b}
	\end{subfigure}
	\hspace{0.05cm}
	\begin{subfigure}[t]{0.22\linewidth}
		\centering
		\includegraphics[scale=.20]{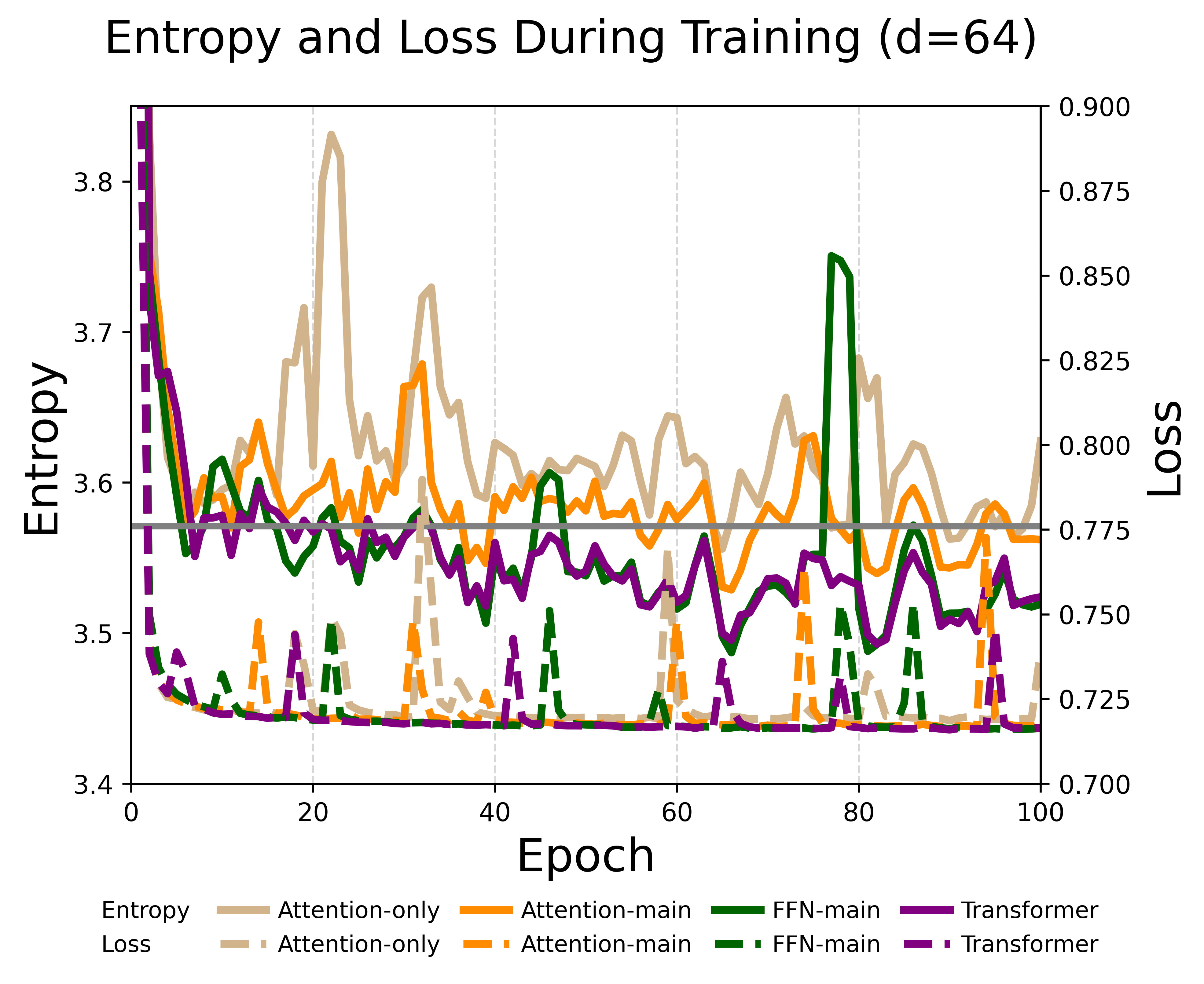}
		%		\subcaption{ }
		%		\label{fig:c}
	\end{subfigure}
	\vspace{-0cm}
	\caption{Entropy and Loss during training for different Transformer variants when $d = 8,16,32,64$.}
	\vspace{-0cm}
	\label{app:fig:entropy_loss_attention_ffn}
\end{figure*}

\begin{figure*}[!htbp]
	\centering
	\begin{subfigure}[t]{0.22\linewidth}
		\centering
		\includegraphics[scale=.18]{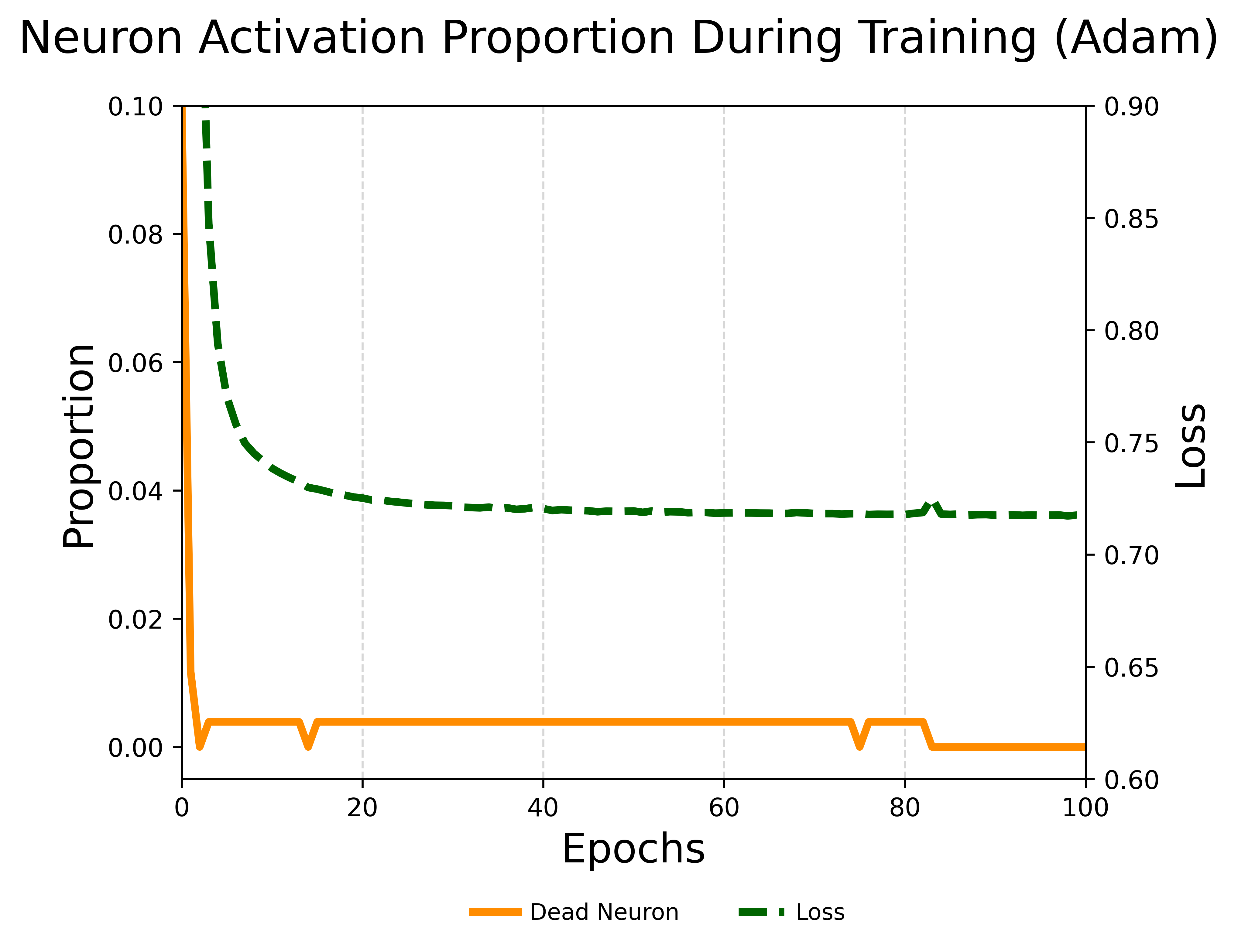}
		%		\subcaption{ }
		%		\label{fig:16-norm}
	\end{subfigure}
	\hspace{0.05cm}
	\begin{subfigure}[t]{0.22\linewidth}
		\centering
		\includegraphics[scale=.18]{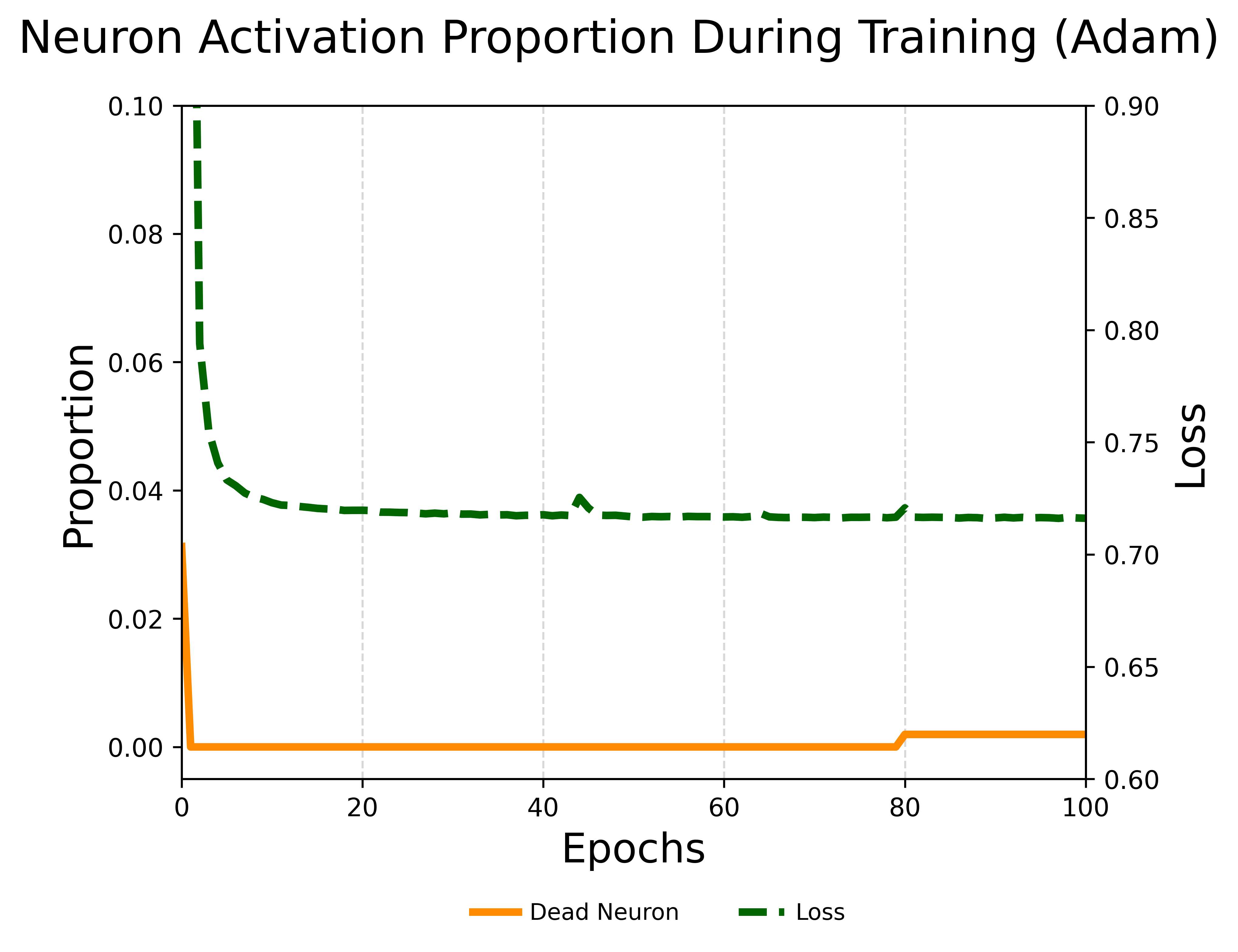}
		%		\subcaption{ }
		%		\label{fig:b}
	\end{subfigure}
	\hspace{0.05cm}
	\begin{subfigure}[t]{0.22\linewidth}
		\centering
		\includegraphics[scale=.18]{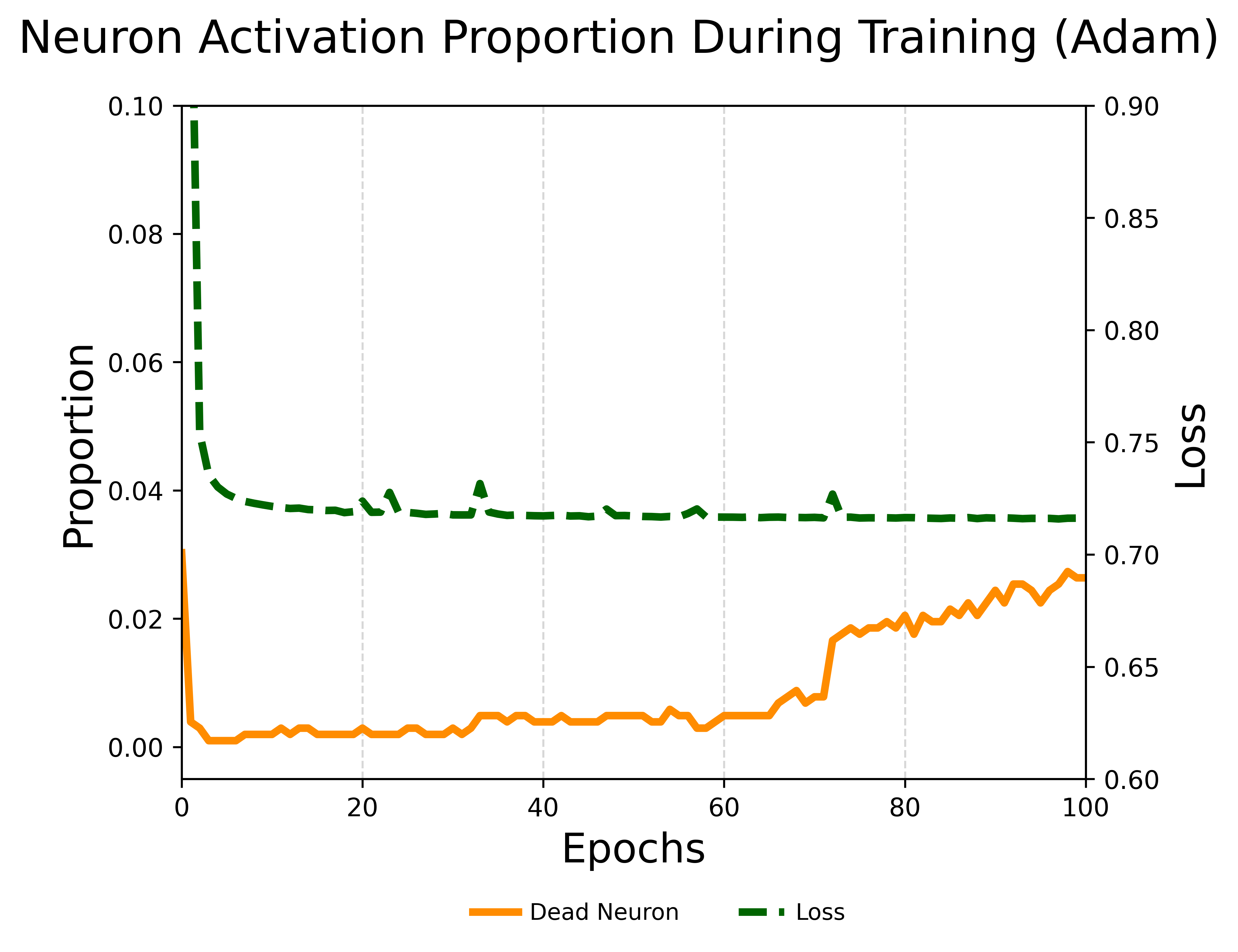}
		%		\subcaption{ }
		%		\label{fig:a}
	\end{subfigure}
	\hspace{0.05cm}
	\begin{subfigure}[t]{0.22\linewidth}
		\centering
		\includegraphics[scale=.18]{pics/Adam_Transformer_64_4_8_simple.png}
		%		\subcaption{ }
		%		\label{fig:c}
	\end{subfigure}
	\vspace{-0.cm}
	\caption{Loss and proportion of dead neurons during training when $d = 8,16,32,64$ (left to right).}
	\vspace{-0cm}
	\label{app:fig:loss_spike_model_size}
\end{figure*}

\newpage
\section{More Results on Different Distributions}\label{app:more-dist}
\subsection{More results on the distribution with lower entropy}\label{app:dist91}
\begin{figure*}[!htbp]
	\centering
	\begin{subfigure}[t]{0.18\linewidth}
		\centering
		\includegraphics[scale=.165]{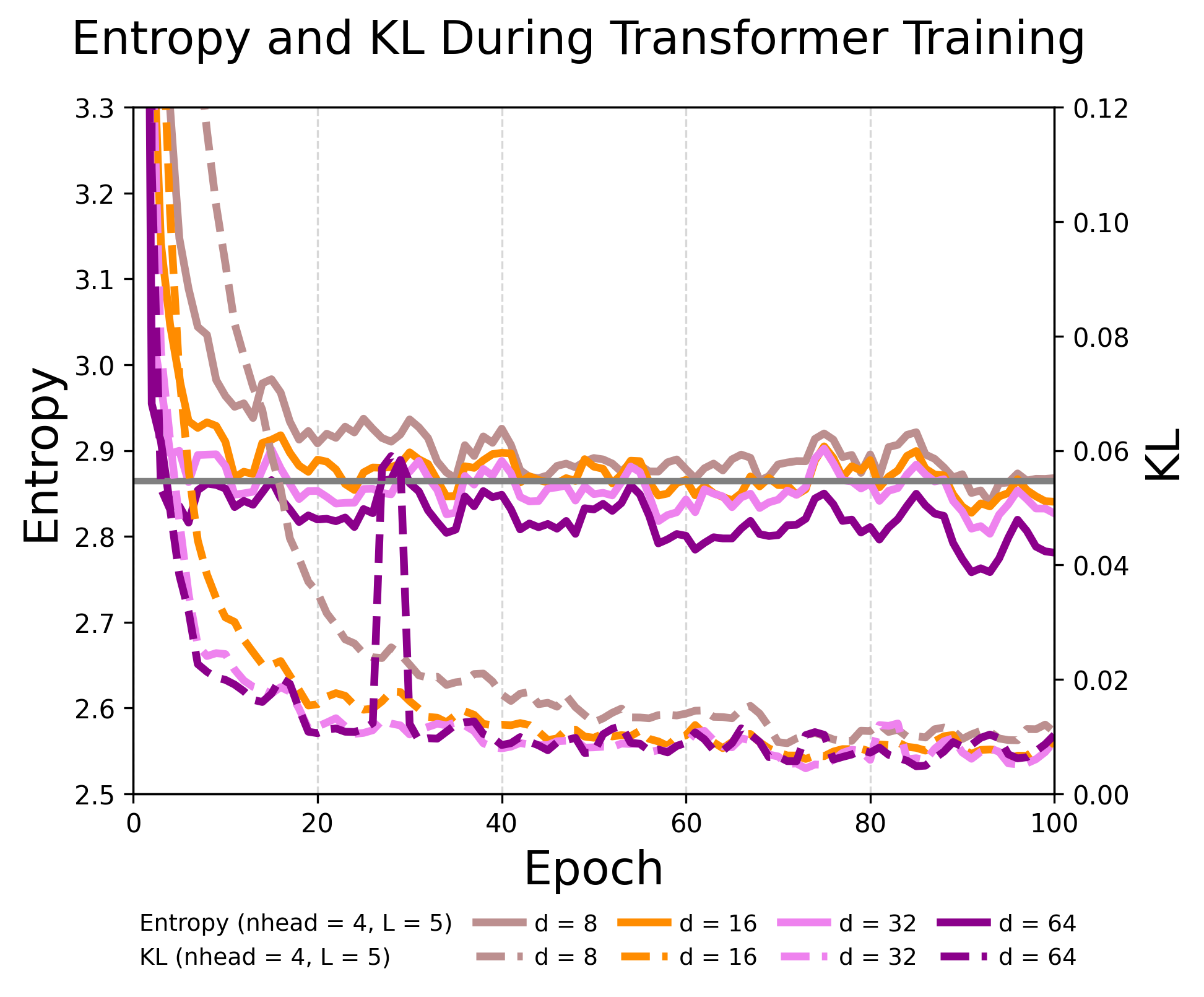}
		%		\subcaption{ }
		%		\label{fig:16-norm}
	\end{subfigure}
	\hspace{0.05cm}
	\begin{subfigure}[t]{0.18\linewidth}
		\centering
		\includegraphics[scale=.165]{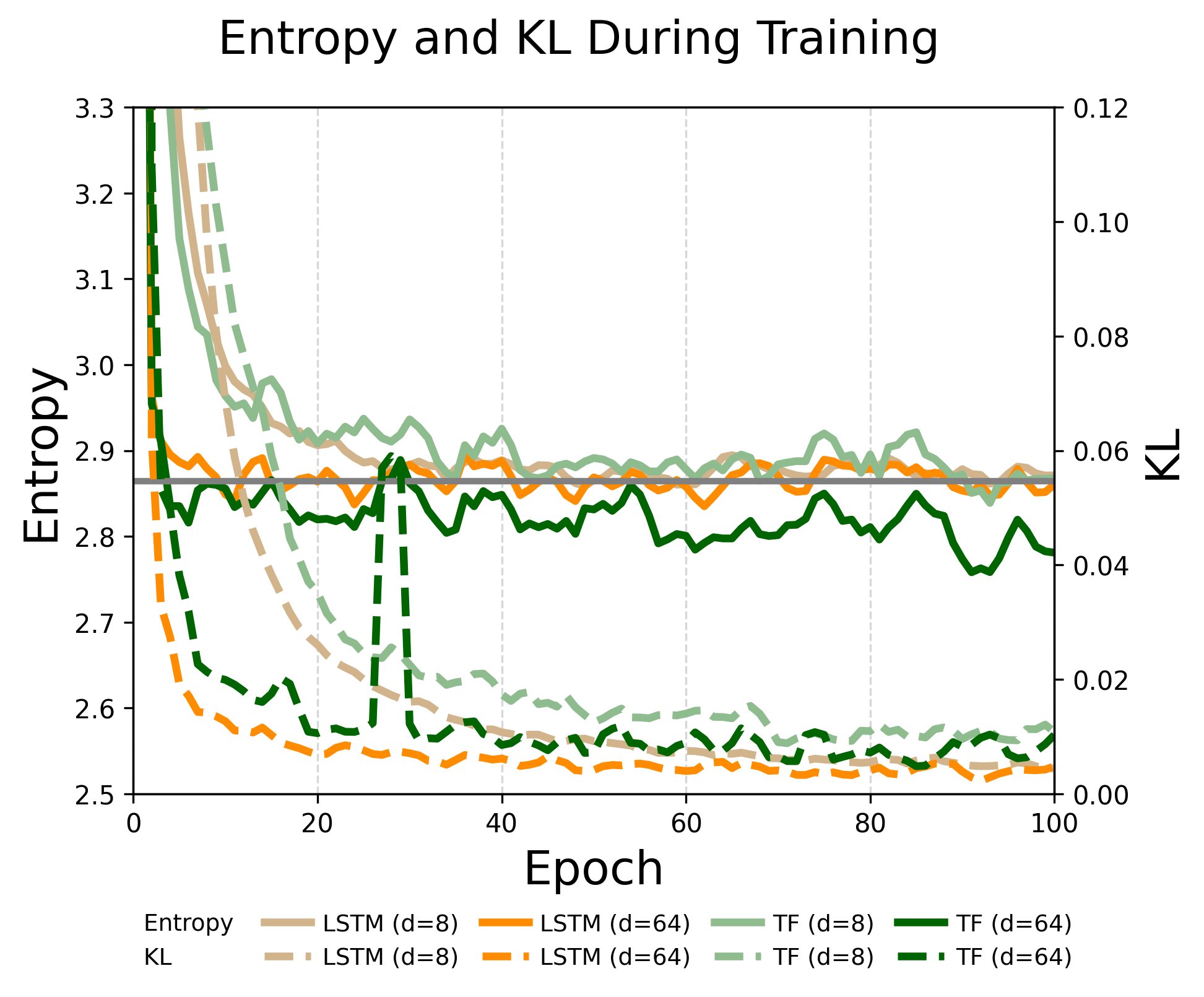}
		%		\subcaption{ }
		%		\label{fig:a}
	\end{subfigure}
	\hspace{0.05cm}
	\begin{subfigure}[t]{0.18\linewidth}
		\centering
		\includegraphics[scale=.17]{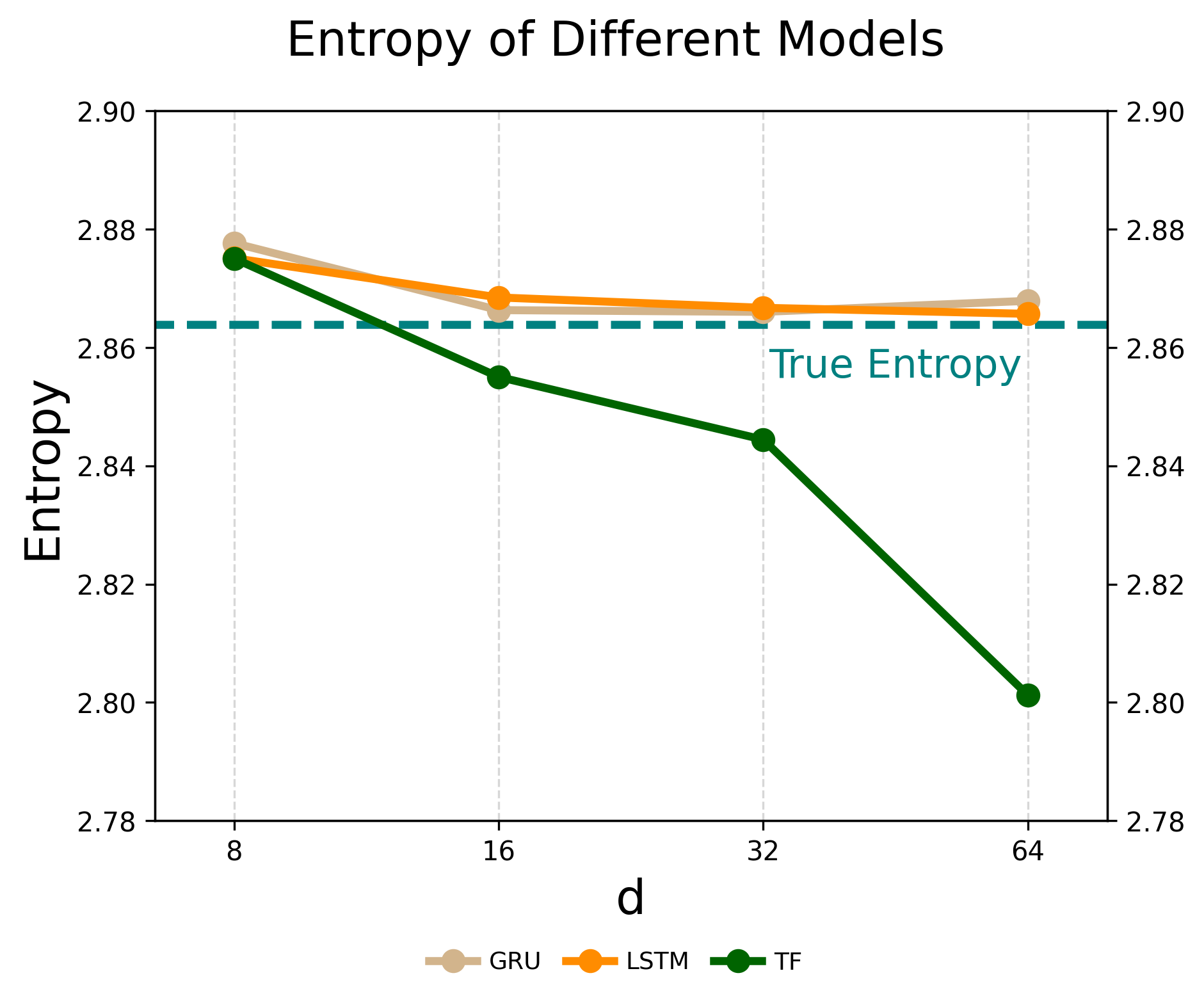}
		%		\subcaption{ }
		%		\label{fig:b}
	\end{subfigure}
	\hspace{0.05cm}
	\begin{subfigure}[t]{0.18\linewidth}
		\centering
		\includegraphics[scale=.17]{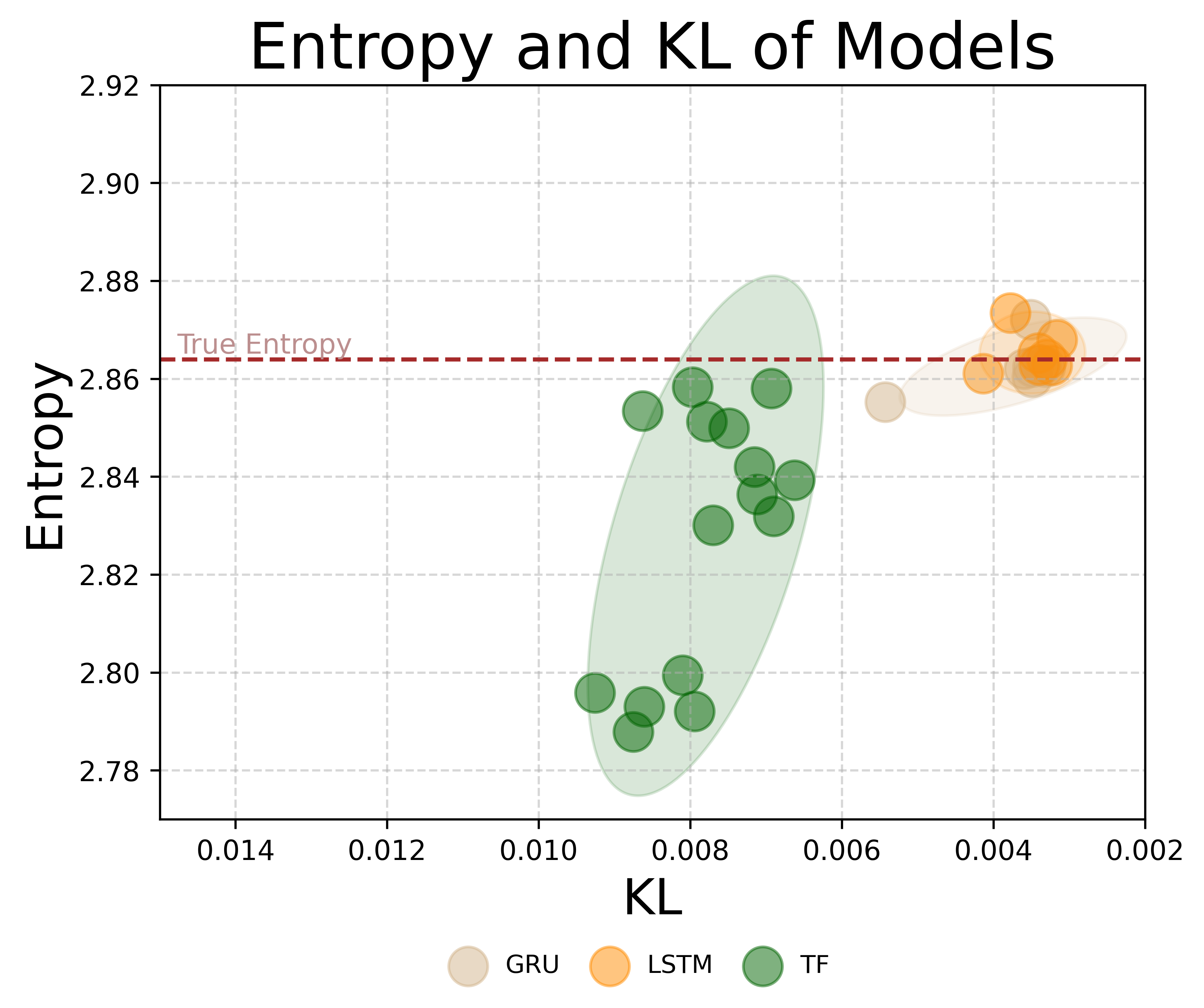}
		%		\subcaption{ }
		%		\label{fig:c}
	\end{subfigure}
	\hspace{0.05cm}
	\begin{subfigure}[t]{0.18\linewidth}
		\centering
		\includegraphics[scale=.17]{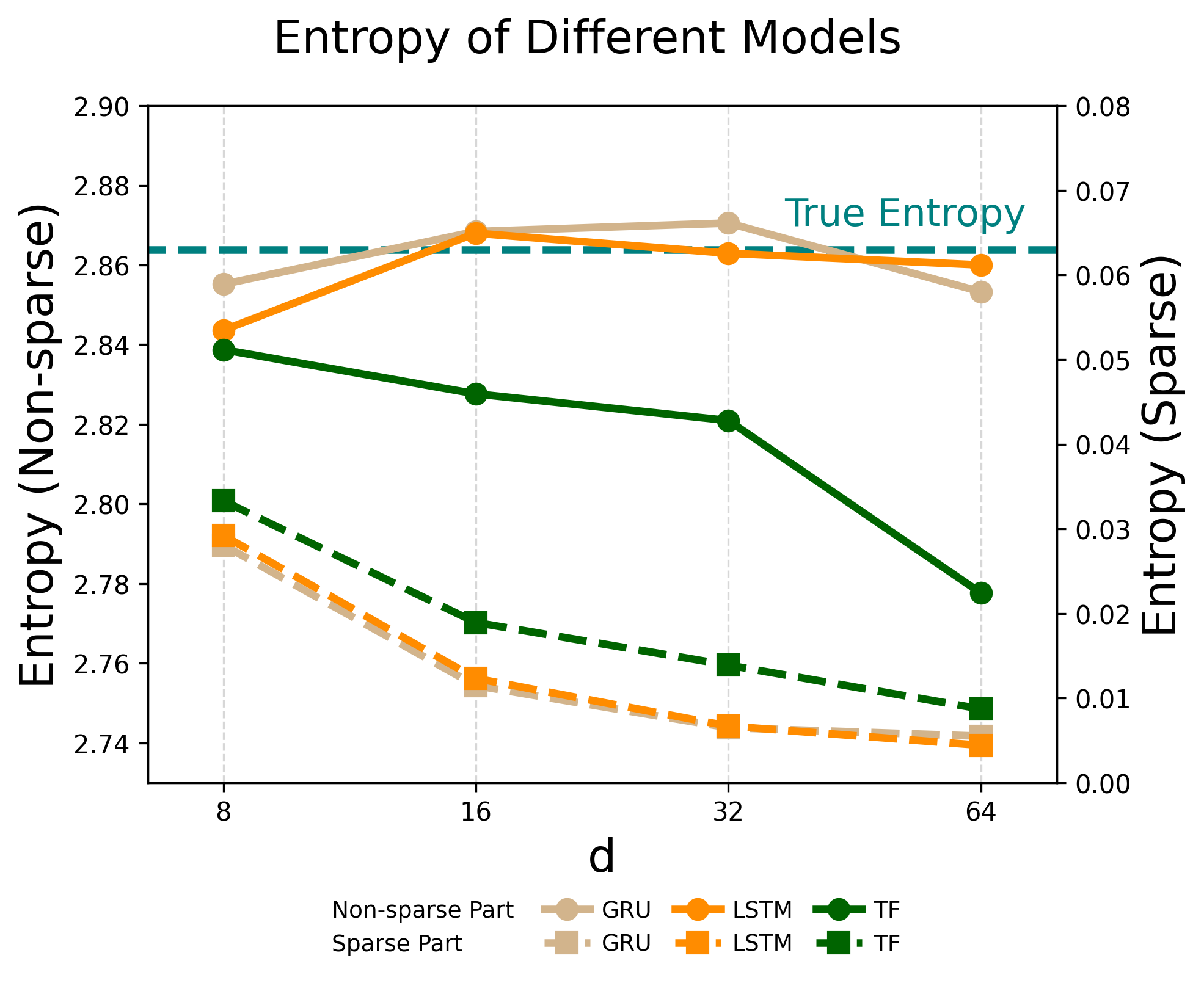}
		%		\subcaption{ }
		%		\label{fig:c}
	\end{subfigure}
	\vspace{-0cm}
	\caption{{\bf Left:} Entropy and KL During Training for Transformers of Different Sizes. {\bf Center Left:} Entropy and KL During Training for Transformer and LSTM when $d=8$ and $d=64$. {\bf Center:} The change in entropy with model size for GRU, LSTM and Transformer, averaged over the last 15 epochs. {\bf Center Right:} The relationship between entropy and KL for different model configurations. {\bf Right:} The change in entropy of the sparse and non-sparse parts with model size.}
	\vspace*{-0cm}
\end{figure*}

\begin{figure*}[!htbp]
	\centering
	\begin{subfigure}[t]{0.22\linewidth}
		\centering
		\includegraphics[scale=.20]{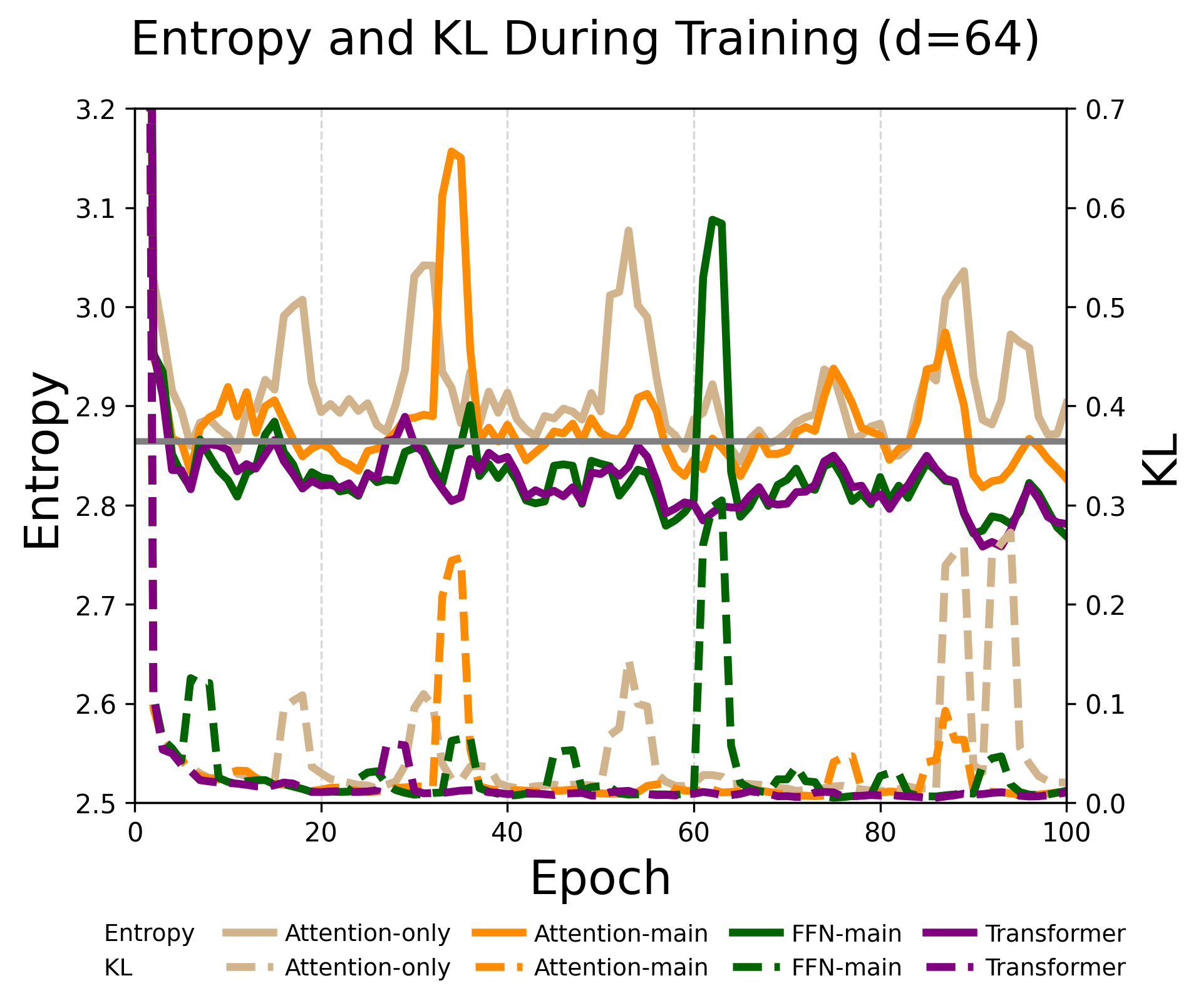}
		%		\subcaption{ }
		%		\label{fig:16-norm}
	\end{subfigure}
	\hspace{0.2cm}
	\begin{subfigure}[t]{0.22\linewidth}
		\centering
		\includegraphics[scale=.205]{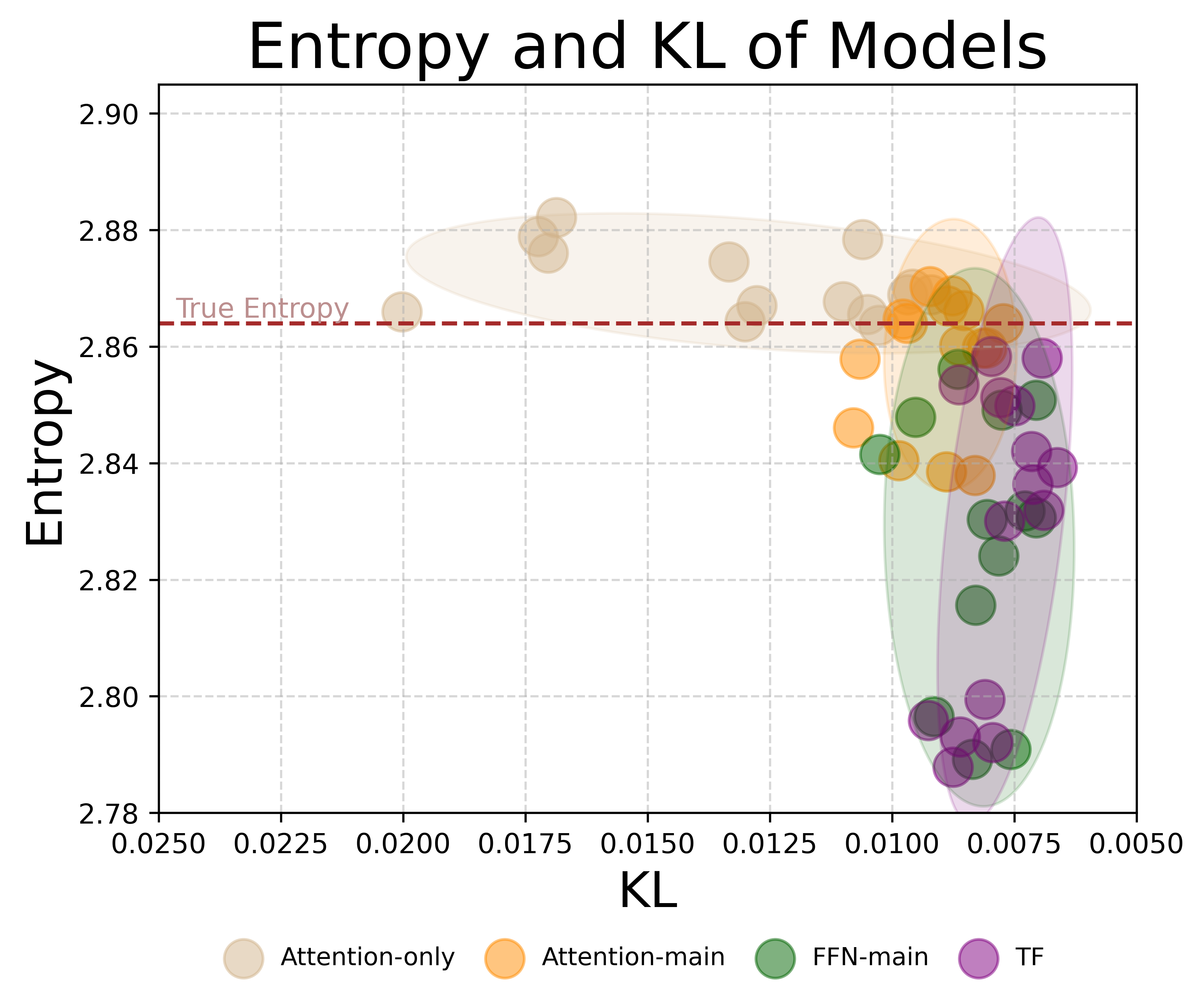}
		%		\subcaption{ }
		%		\label{fig:a}
	\end{subfigure}
	\hspace{0.2cm}
	\begin{subfigure}[t]{0.22\linewidth}
		\centering
		\includegraphics[scale=.205]{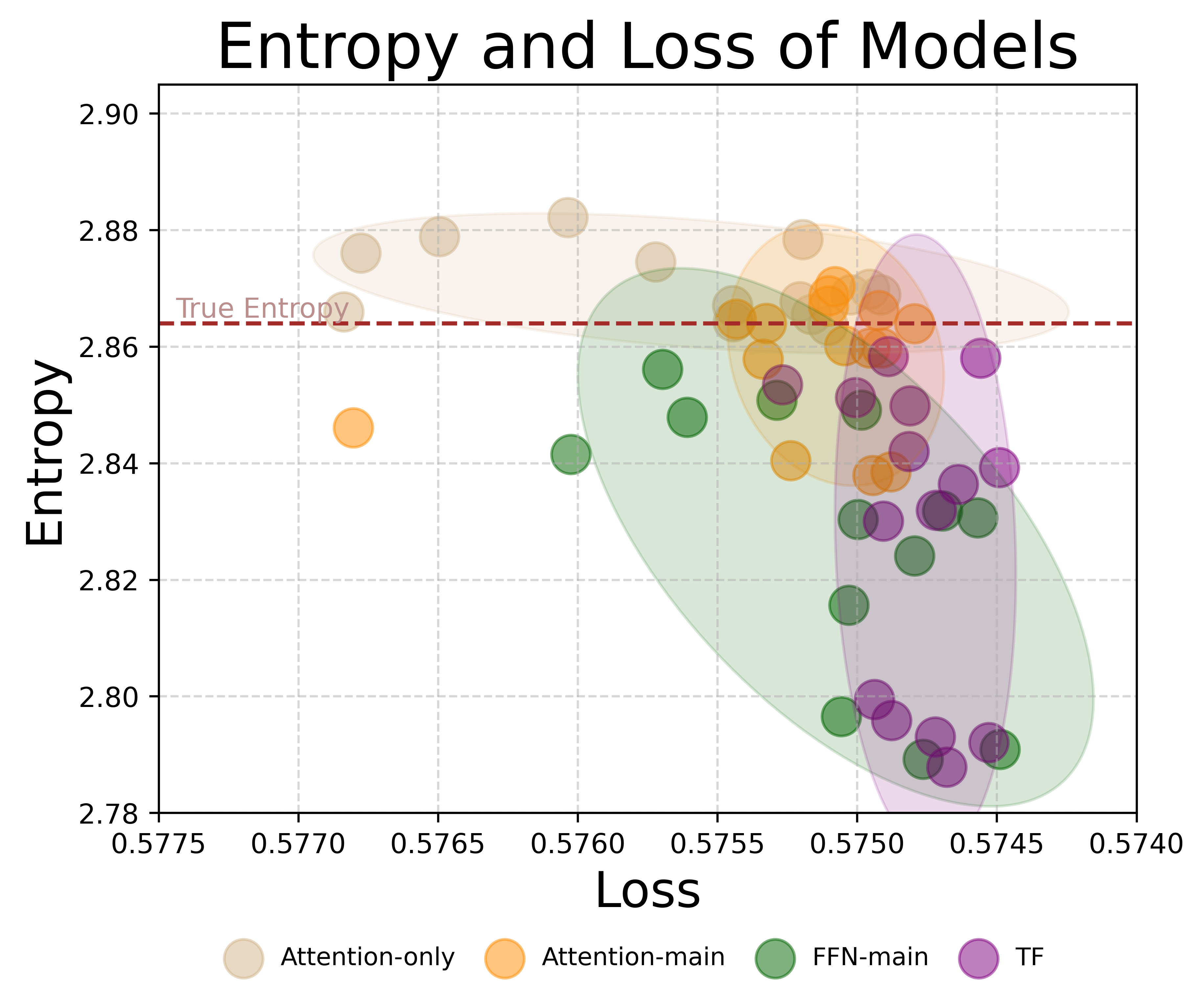}
		%		\subcaption{ }
		%		\label{fig:b}
	\end{subfigure}
	\hspace{0.2cm}
	\begin{subfigure}[t]{0.22\linewidth}
		\centering
		\includegraphics[scale=.205]{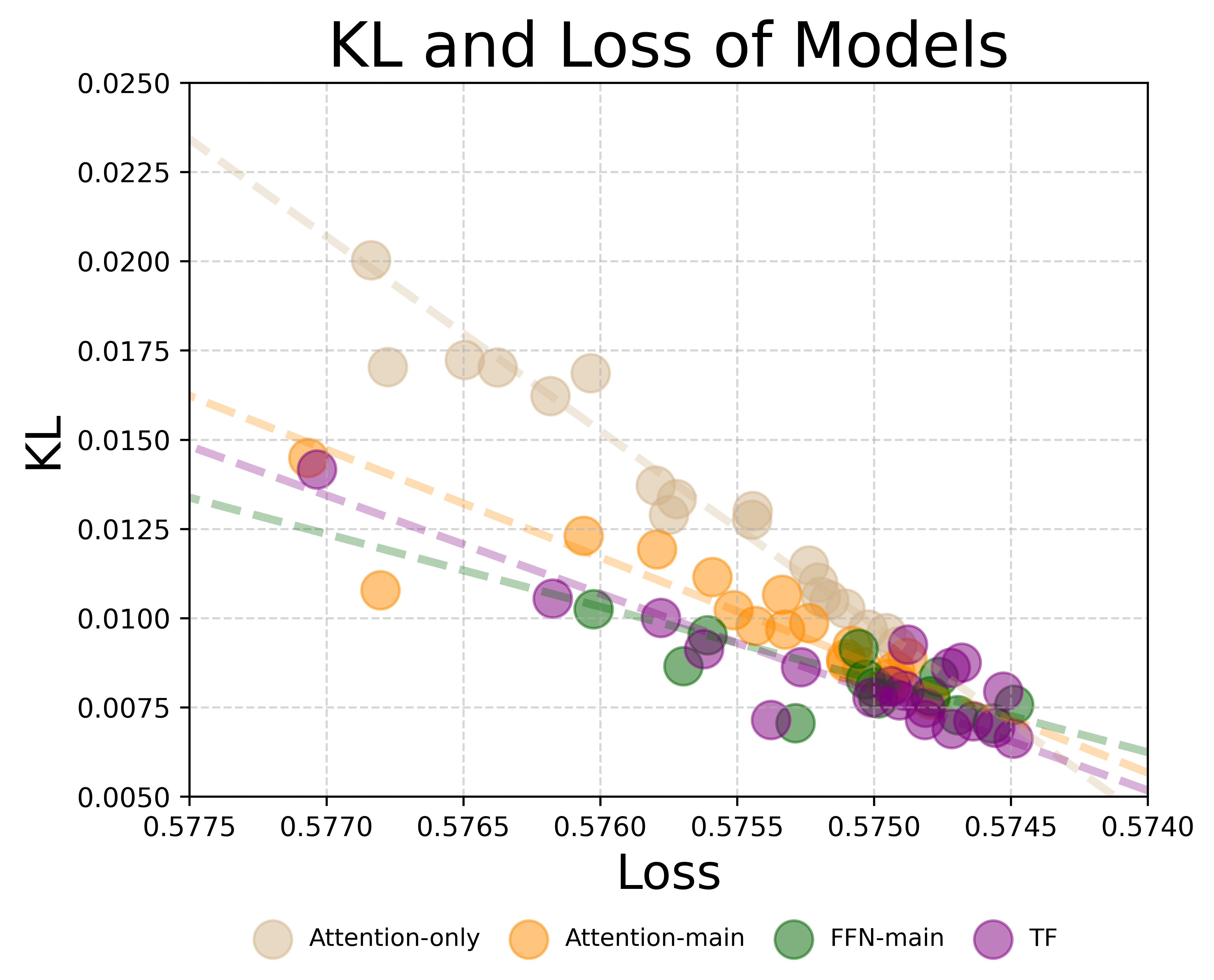}
		%		\subcaption{ }
		%		\label{fig:c}
	\end{subfigure}
	\vspace{-0cm}
	\caption{
		{\bf Left:} Entropy and KL during training for different Transformer variants. 
		{\bf Center Left and Center Right:} Relationship between KL/Loss and Entropy for different variants. FFN-main can achieve lower entropy compared to Attention-only and Attention-main.
		{\bf Right:} Relationship between KL and Loss for different variants.
	}
	\vspace*{-0cm}
\end{figure*}

\begin{figure*}[!htbp]
	\centering
	\begin{subfigure}[t]{0.22\linewidth}
		\centering
		\includegraphics[scale=.21]{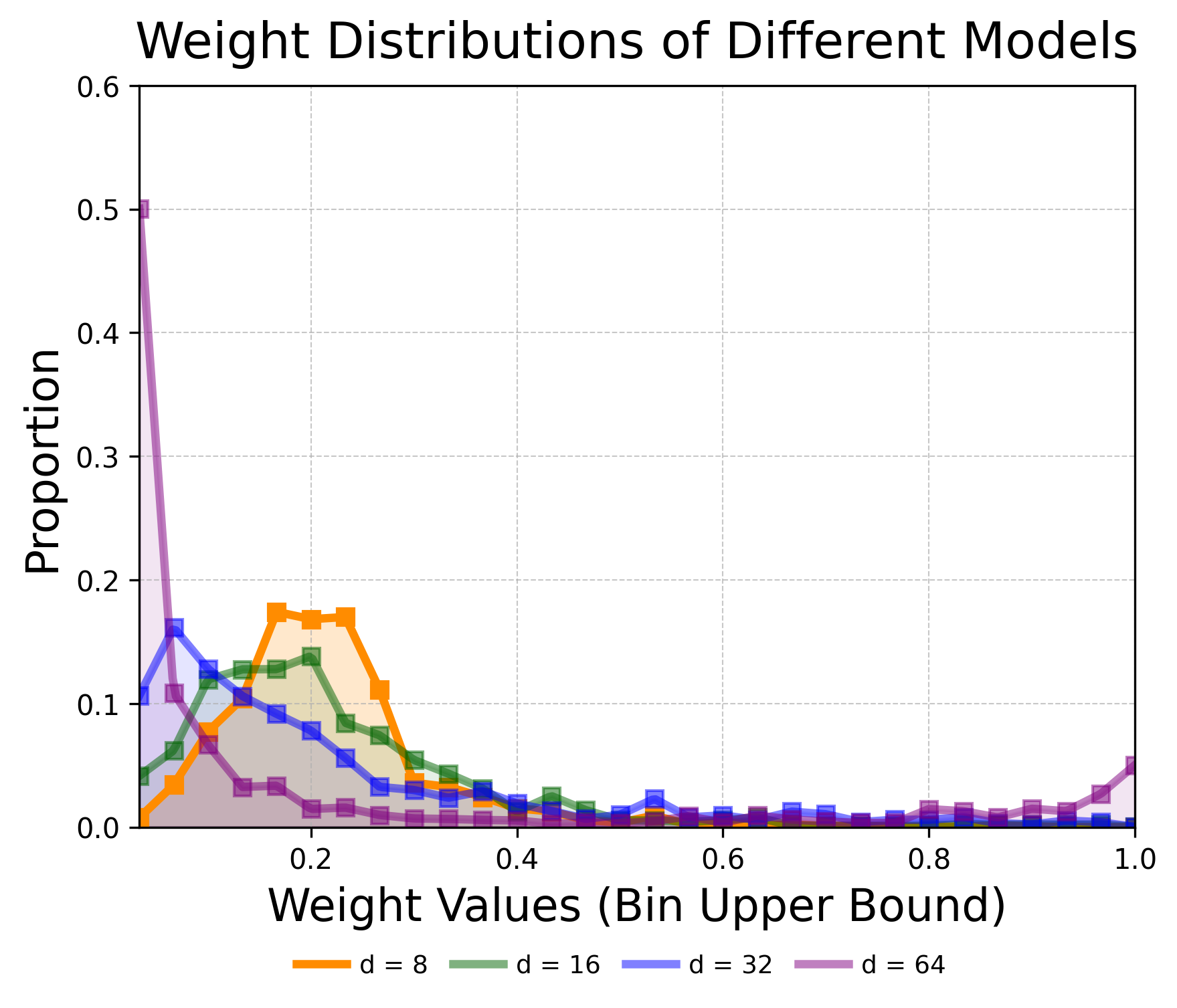}
		%		\subcaption{ }
		%		\label{fig:16-norm}
	\end{subfigure}
	\hspace{0.1cm}
	\begin{subfigure}[t]{0.22\linewidth}
		\centering
		\includegraphics[scale=.21]{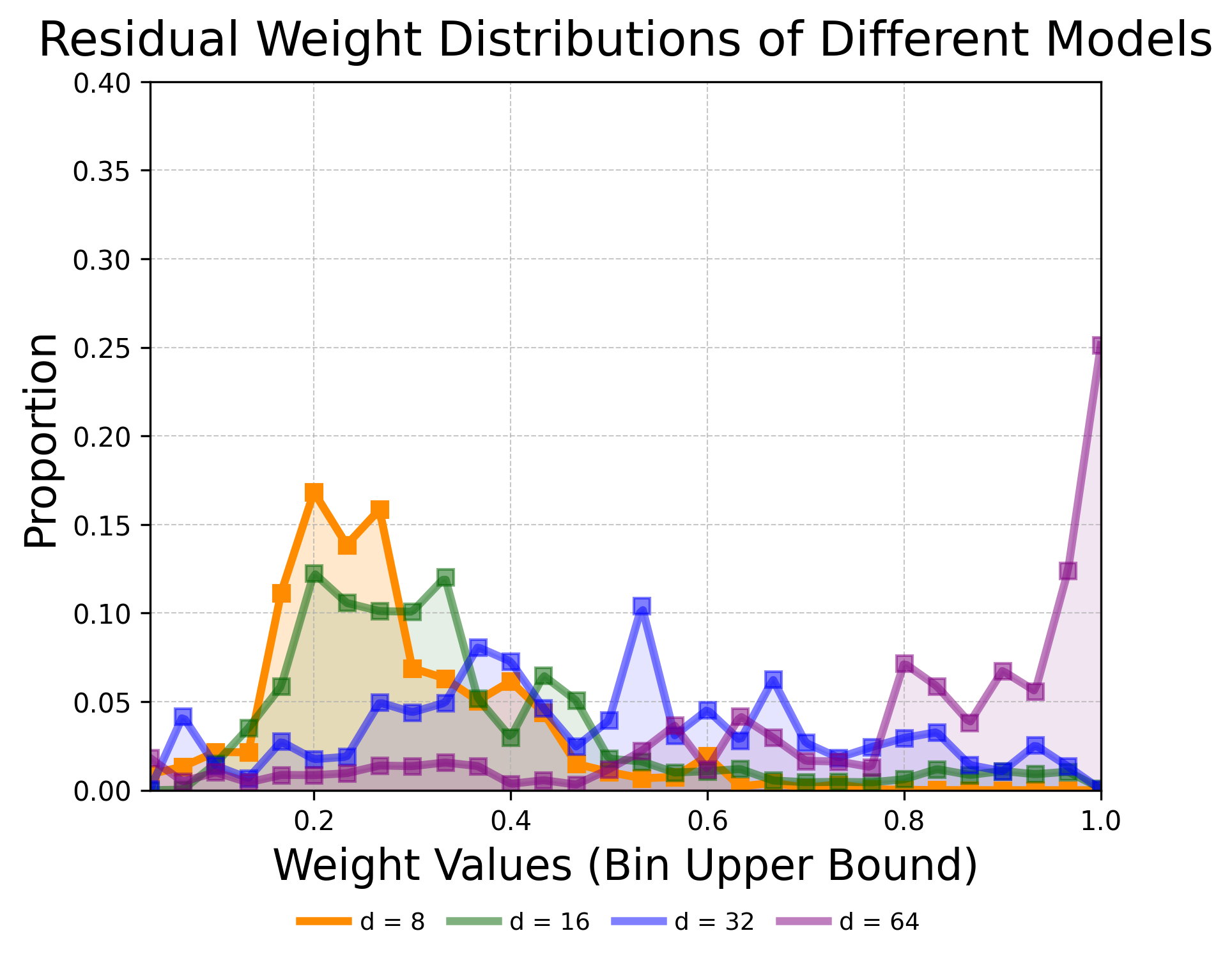}
		%		\subcaption{ }
		%		\label{fig:a}
	\end{subfigure}
	\hspace{0.1cm}
	\begin{subfigure}[t]{0.22\linewidth}
		\centering
		\includegraphics[scale=.21]{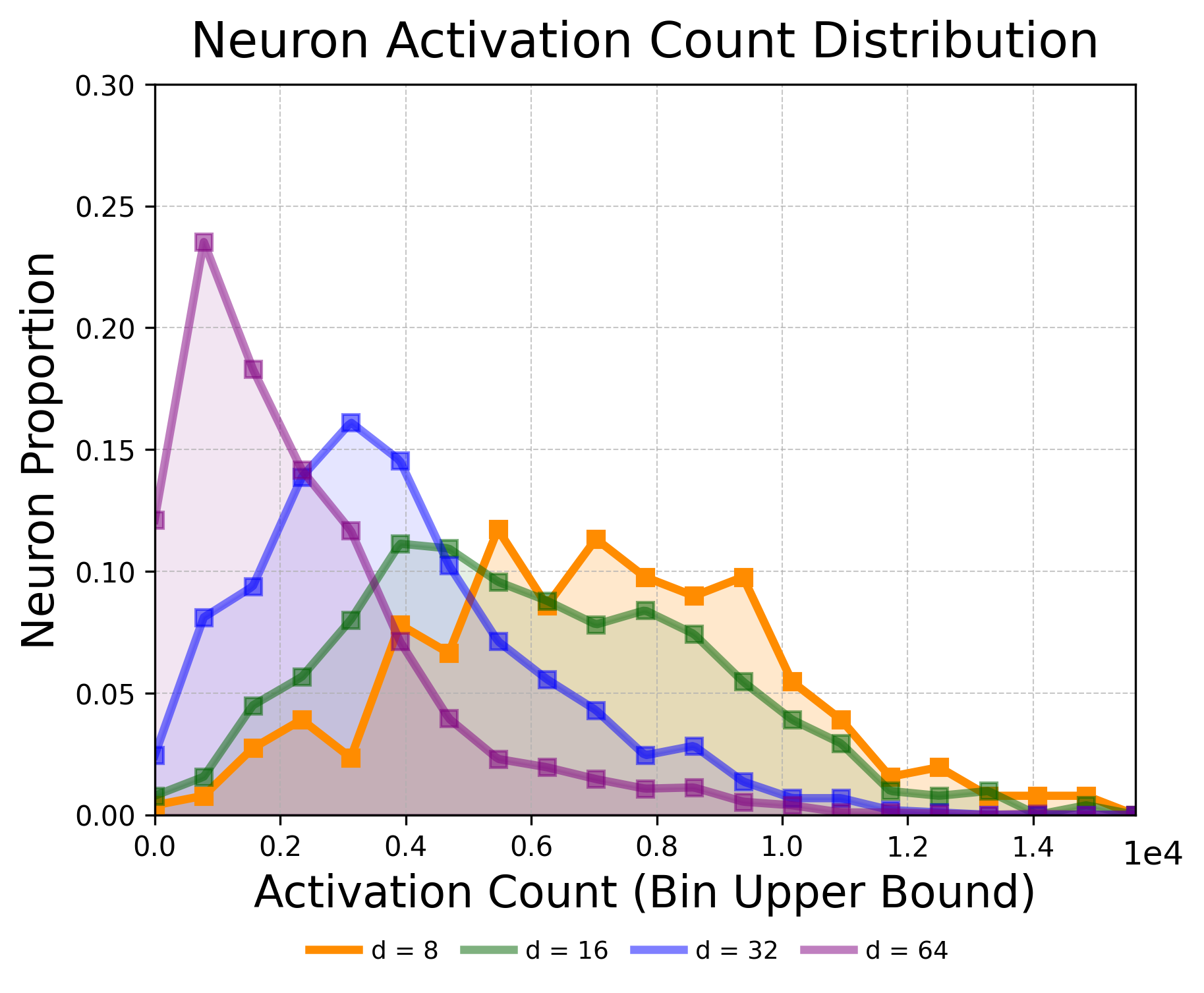}
		%		\subcaption{ }
		%		\label{fig:b}
	\end{subfigure}
	\hspace{0.1cm}
	\begin{subfigure}[t]{0.22\linewidth}
		\centering
		\includegraphics[scale=.21]{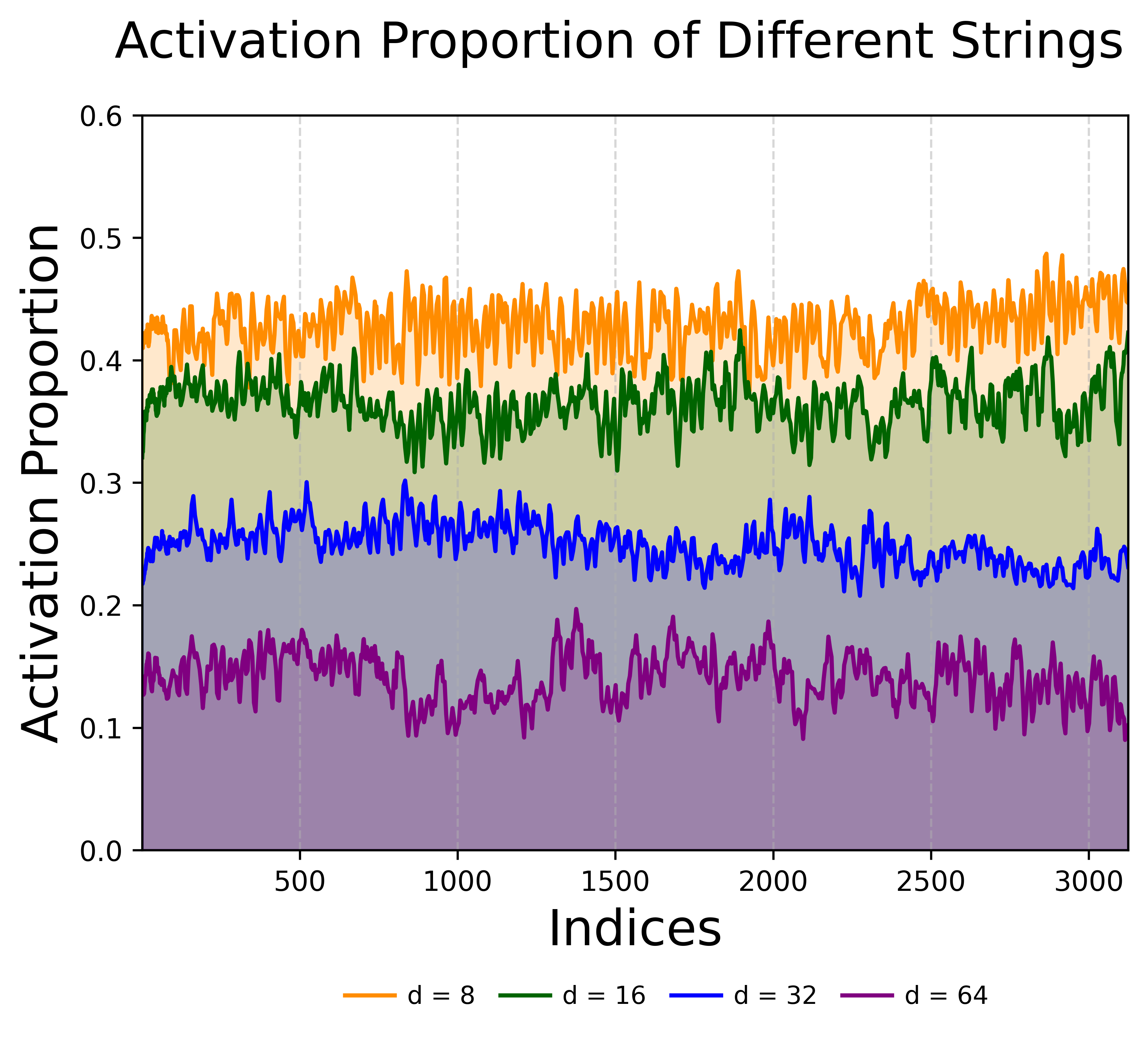}
		%		\subcaption{ }
		%		\label{fig:c}
	\end{subfigure}
	\vspace{-0cm}
	\caption{
		{\bf Left:} The weight distribution of all paths for different model sizes.
		{\bf Center Left:} The weight distribution of only residual paths for different model sizes.
		{\bf Center Right:} The distribution of neuron activation counts for Transformers of different sizes.
		{\bf Right:} The proportion of neurons activated across different input sequences.
	}
	\vspace*{-0cm}
\end{figure*}

\begin{figure*}[!htbp]
	\centering
	\begin{subfigure}[t]{0.18\linewidth}
		\centering
		\includegraphics[scale=.165]{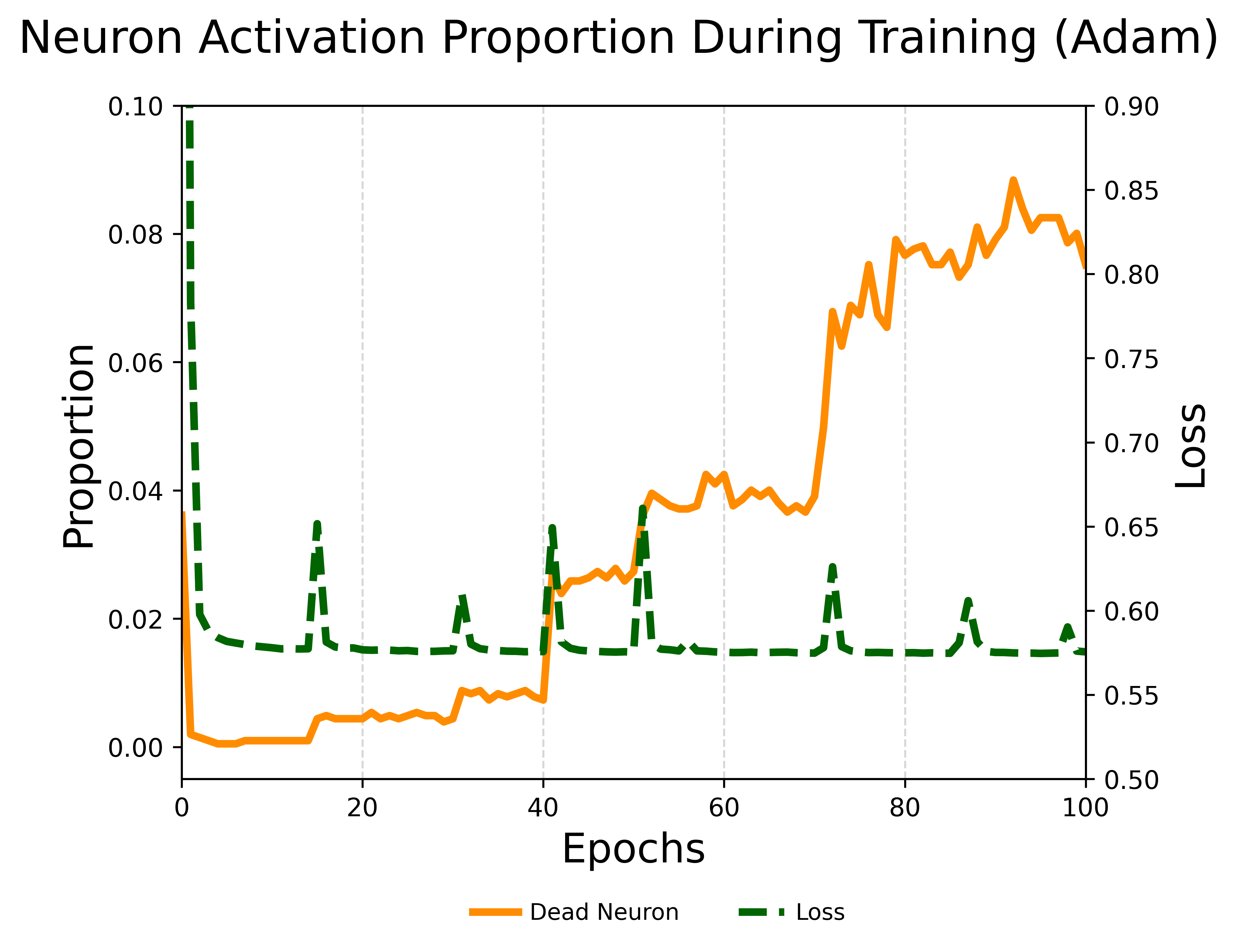}
		%		\subcaption{ }
		%		\label{fig:16-norm}
	\end{subfigure}
	\hspace{0.15cm}
	\begin{subfigure}[t]{0.18\linewidth}
		\centering
		\includegraphics[scale=.165]{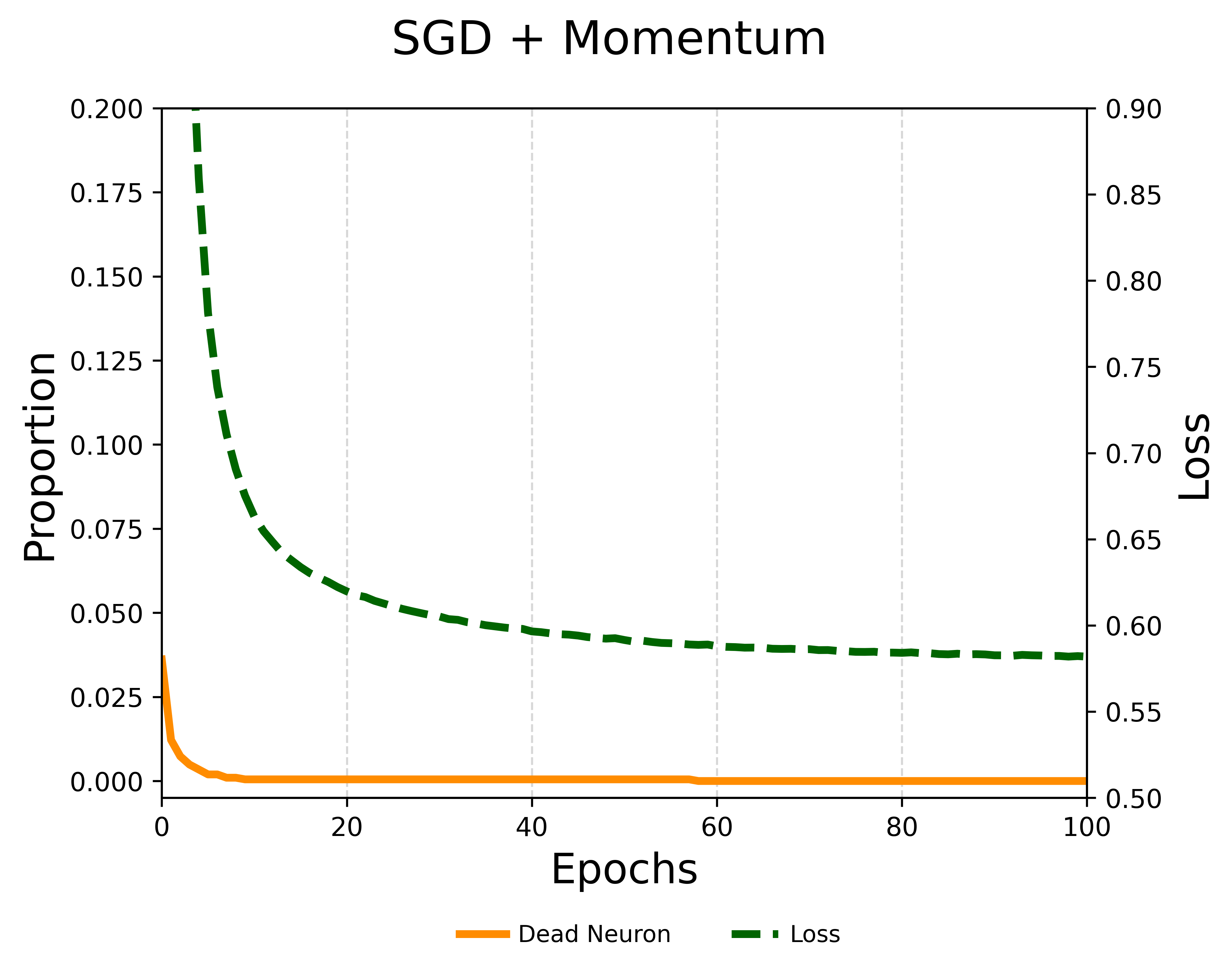}
		%		\subcaption{ }
		%		\label{fig:16-norm}
	\end{subfigure}
	\hspace{0.15cm}
	\begin{subfigure}[t]{0.18\linewidth}
		\centering
		\includegraphics[scale=.165]{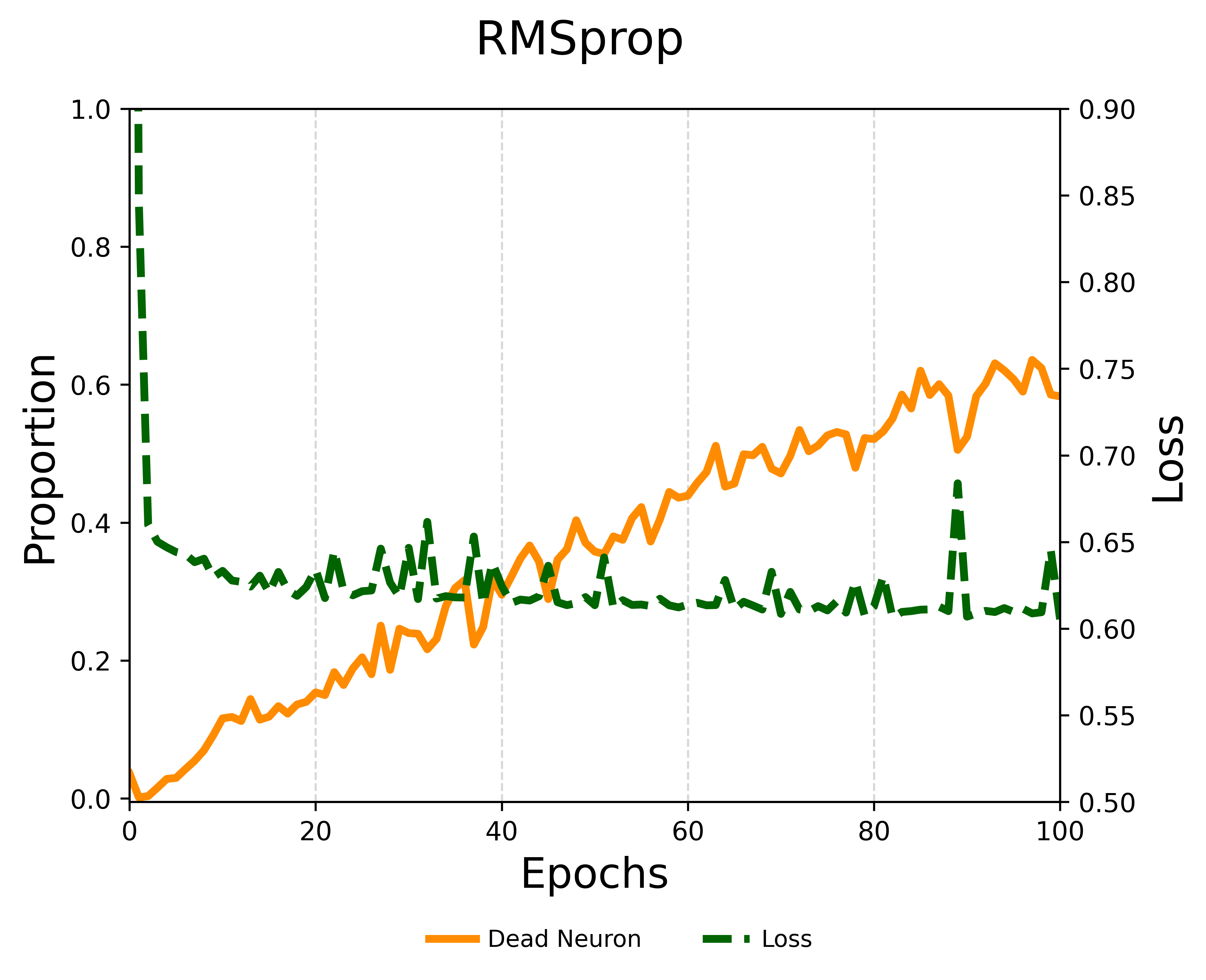}
		%		\subcaption{ }
		%		\label{fig:b}
	\end{subfigure}
	\hspace{0.15cm}
	\begin{subfigure}[t]{0.18\linewidth}
		\centering
		\includegraphics[scale=.165]{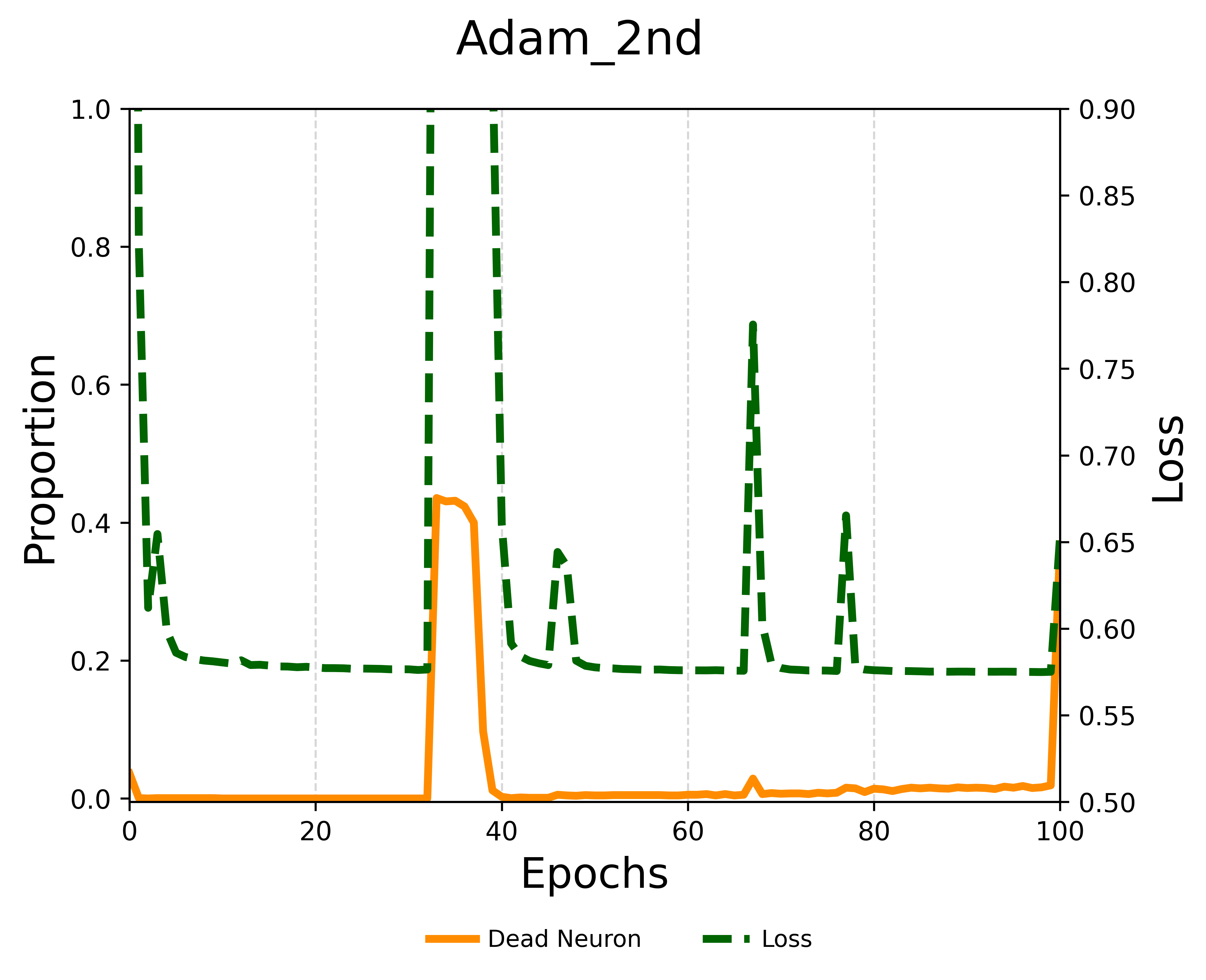}
		%		\subcaption{ }
		%		\label{fig:a}
	\end{subfigure}
	\hspace{0.15cm}
	\begin{subfigure}[t]{0.18\linewidth}
		\centering
		\includegraphics[scale=.165]{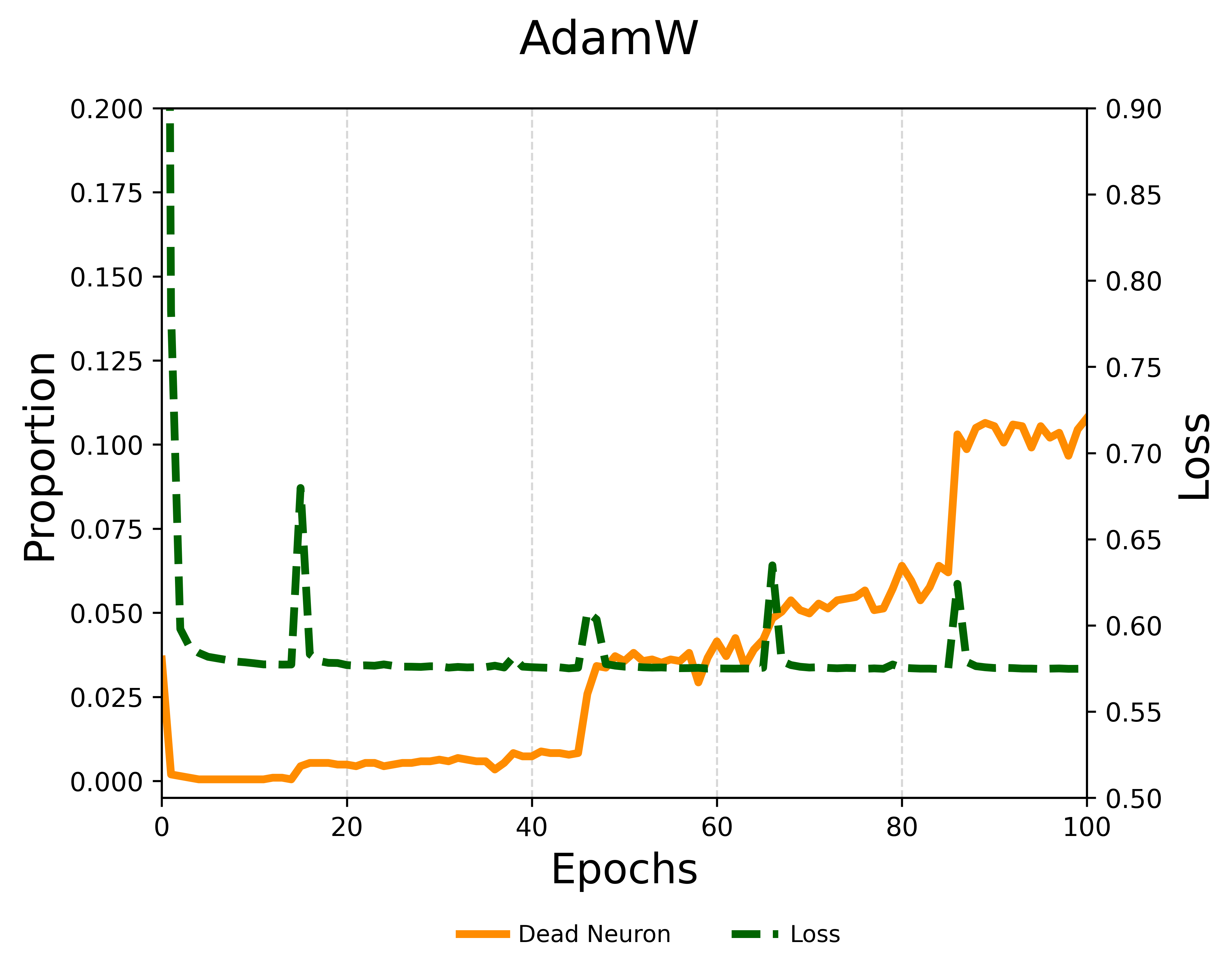}
		%		\subcaption{ }
		%		\label{fig:c}
	\end{subfigure}
	\vspace{-0cm}
	\caption{Loss and the proportion of dead neurons during training under different optimizers. We use {\bf Adam} as the optimizer by default. {\bf SGD with momentum} does not cause loss spikes or an increase in the proportion of dead neurons. {\bf RMSprop} exhibits unstable training and a higher proportion of dead neurons. {\bf Adam primarily relying on second-order gradients} causes more severe loss spikes and greater fluctuations in the dead neurons proportion compared to regular Adam while {\bf AdamW} shows behavior similar to Adam.}
	\vspace*{-0cm}
\end{figure*}

\newpage
\subsection{More results on the distribution with higher entropy}\label{app:dist631}
\begin{figure*}[!htbp]
	\centering
	\begin{subfigure}[t]{0.18\linewidth}
		\centering
		\includegraphics[scale=.165]{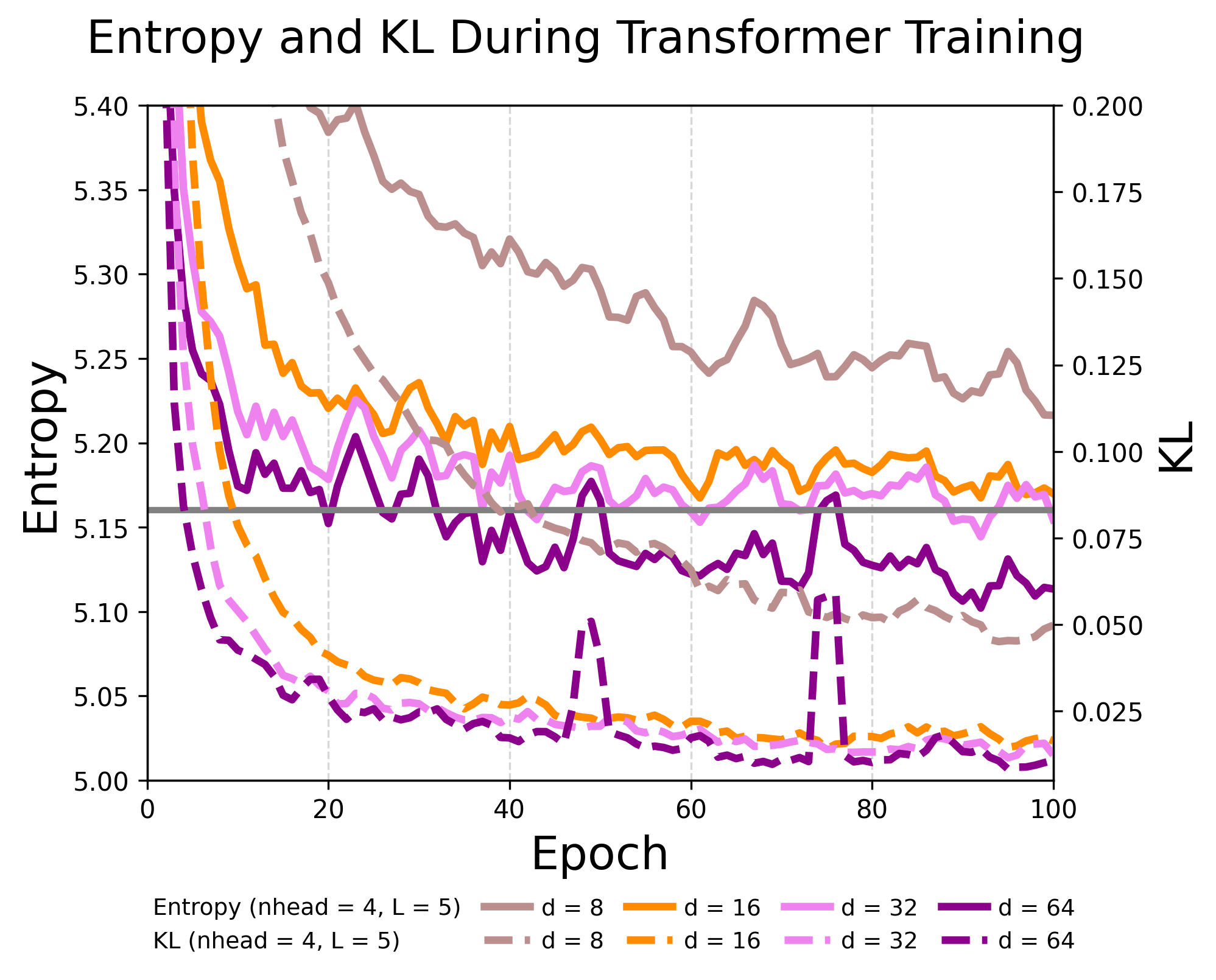}
		%		\subcaption{ }
		%		\label{fig:16-norm}
	\end{subfigure}
	\hspace{0.05cm}
	\begin{subfigure}[t]{0.18\linewidth}
		\centering
		\includegraphics[scale=.165]{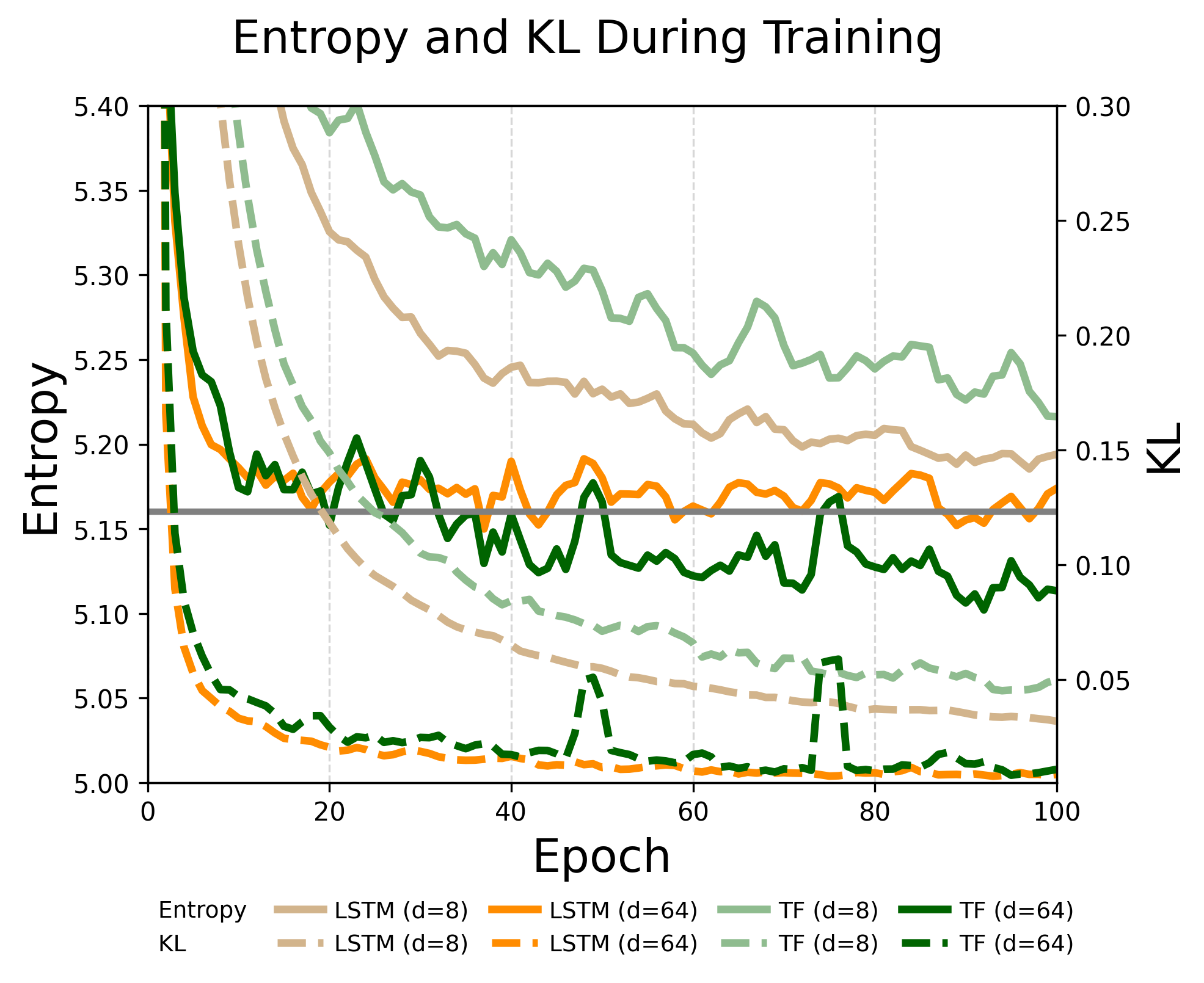}
		%		\subcaption{ }
		%		\label{fig:a}
	\end{subfigure}
	\hspace{0.05cm}
	\begin{subfigure}[t]{0.18\linewidth}
		\centering
		\includegraphics[scale=.17]{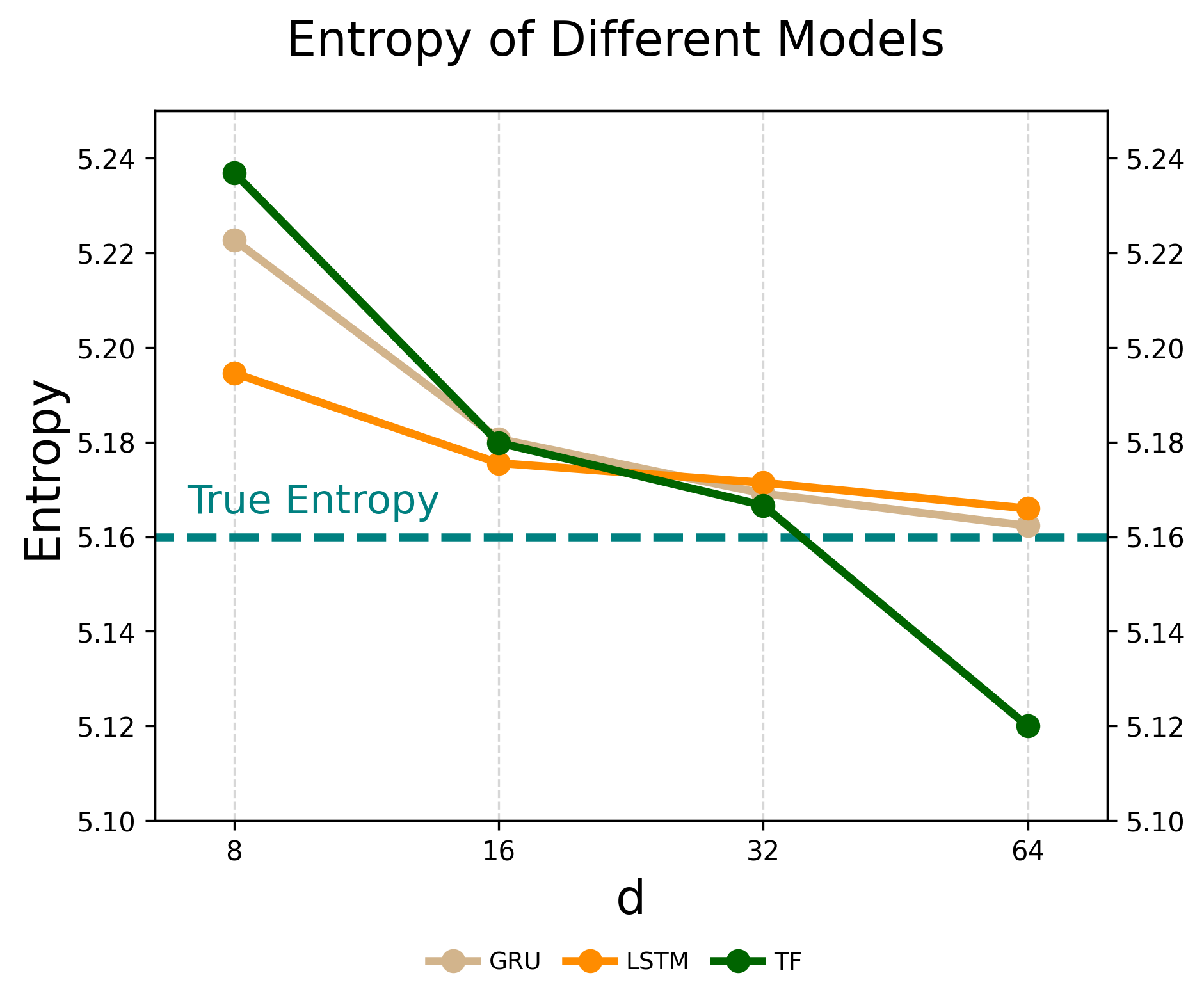}
		%		\subcaption{ }
		%		\label{fig:b}
	\end{subfigure}
	\hspace{0.05cm}
	\begin{subfigure}[t]{0.18\linewidth}
		\centering
		\includegraphics[scale=.17]{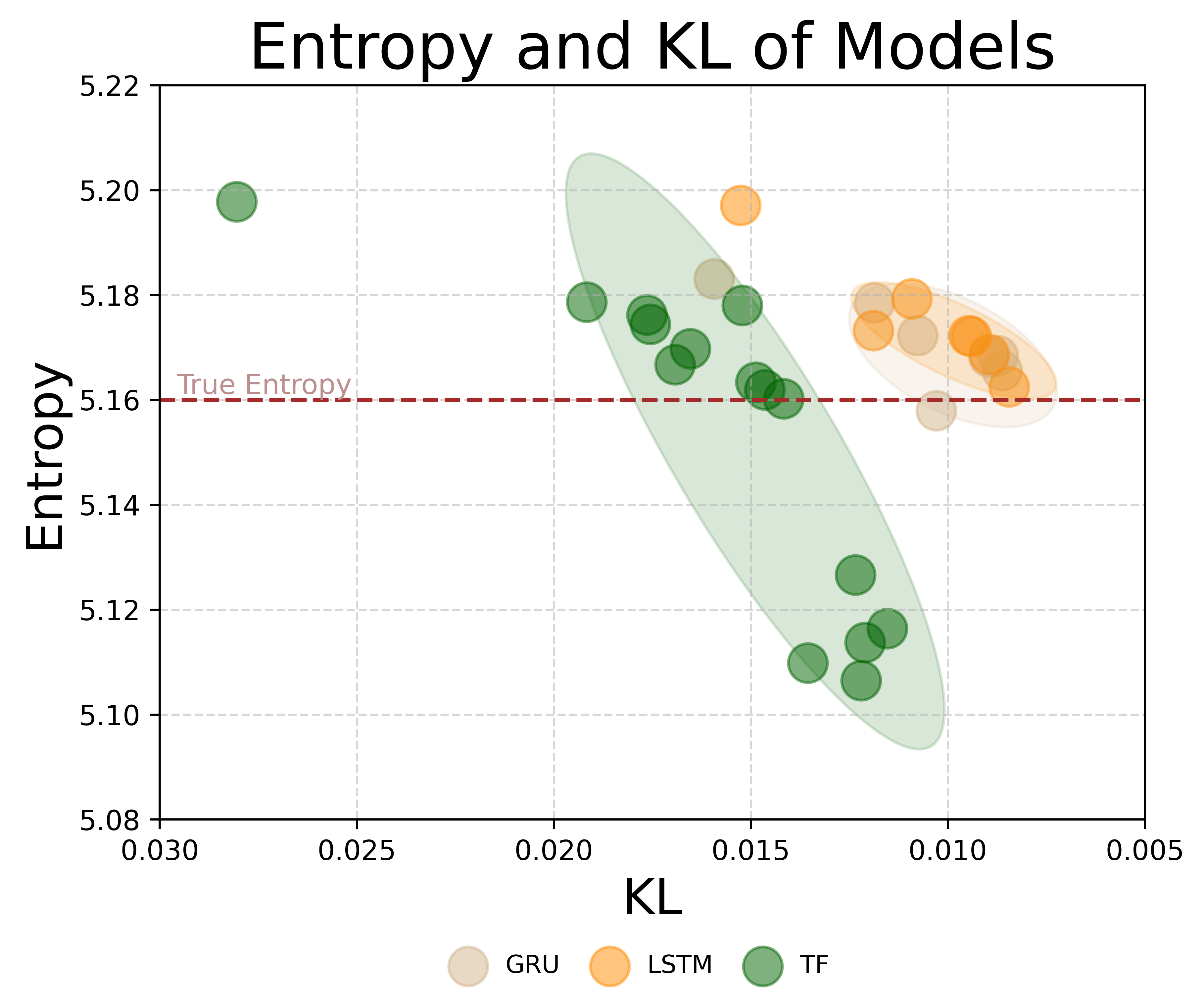}
		%		\subcaption{ }
		%		\label{fig:c}
	\end{subfigure}
	\hspace{0.05cm}
	\begin{subfigure}[t]{0.18\linewidth}
		\centering
		\includegraphics[scale=.17]{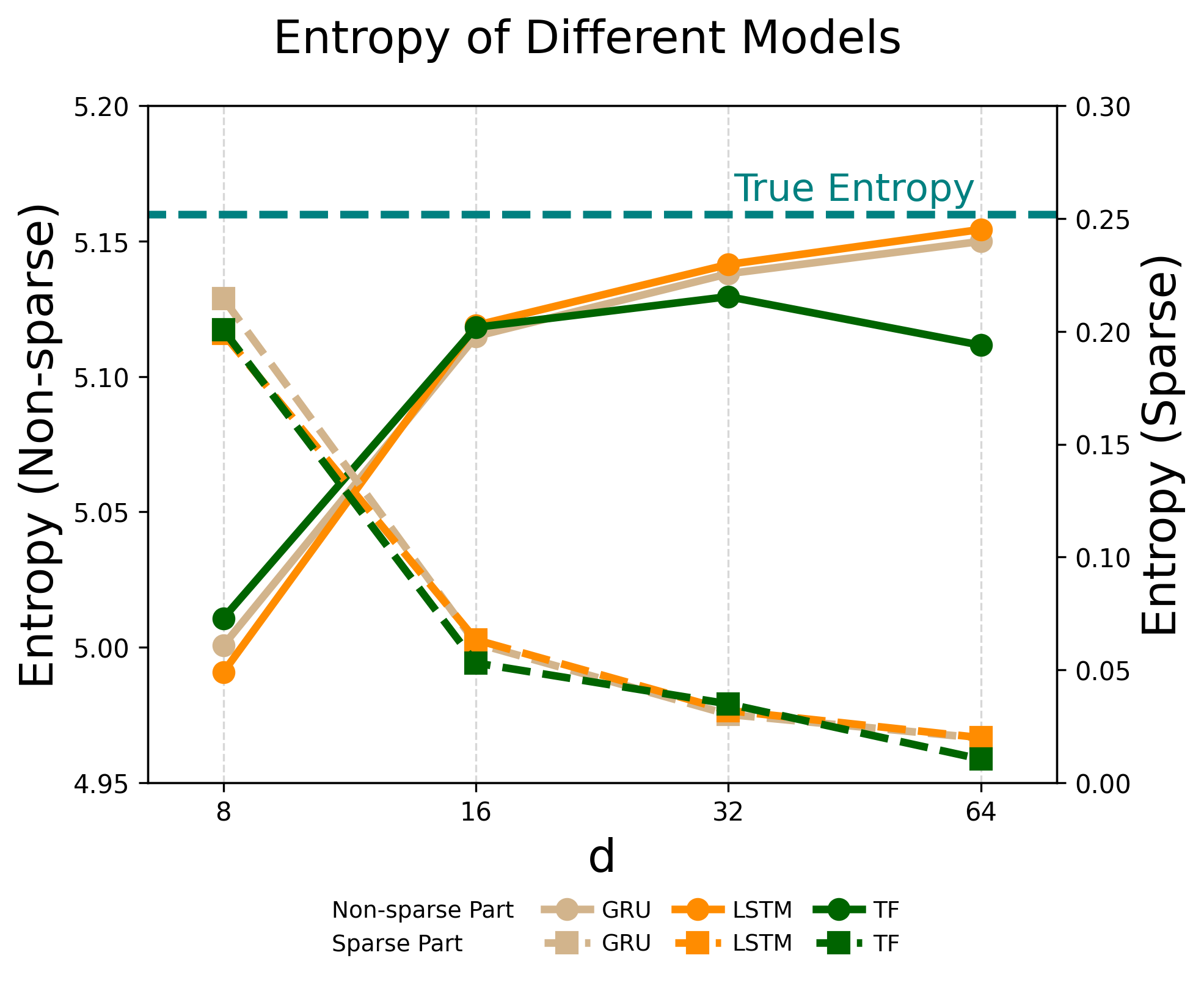}
		%		\subcaption{ }
		%		\label{fig:c}
	\end{subfigure}
	\vspace{-0cm}
	\caption{{\bf Left:} Entropy and KL During Training for Transformers of Different Sizes. {\bf Center Left:} Entropy and KL During Training for Transformer and LSTM when $d=8$ and $d=64$. {\bf Center:} The change in entropy with model size for GRU, LSTM and Transformer, averaged over the last 15 epochs. {\bf Center Right:} The relationship between entropy and KL for different model configurations. For Transformers, {\bf the five scatters at the bottom that are far below the true entropy exactly correspond to cases when $d = 64$}.{\bf Right:} The change in entropy of the sparse and non-sparse parts with model size.}
	\vspace*{-0cm}
\end{figure*}

\begin{figure*}[!htbp]
	\centering
	\begin{subfigure}[t]{0.22\linewidth}
		\centering
		\includegraphics[scale=.20]{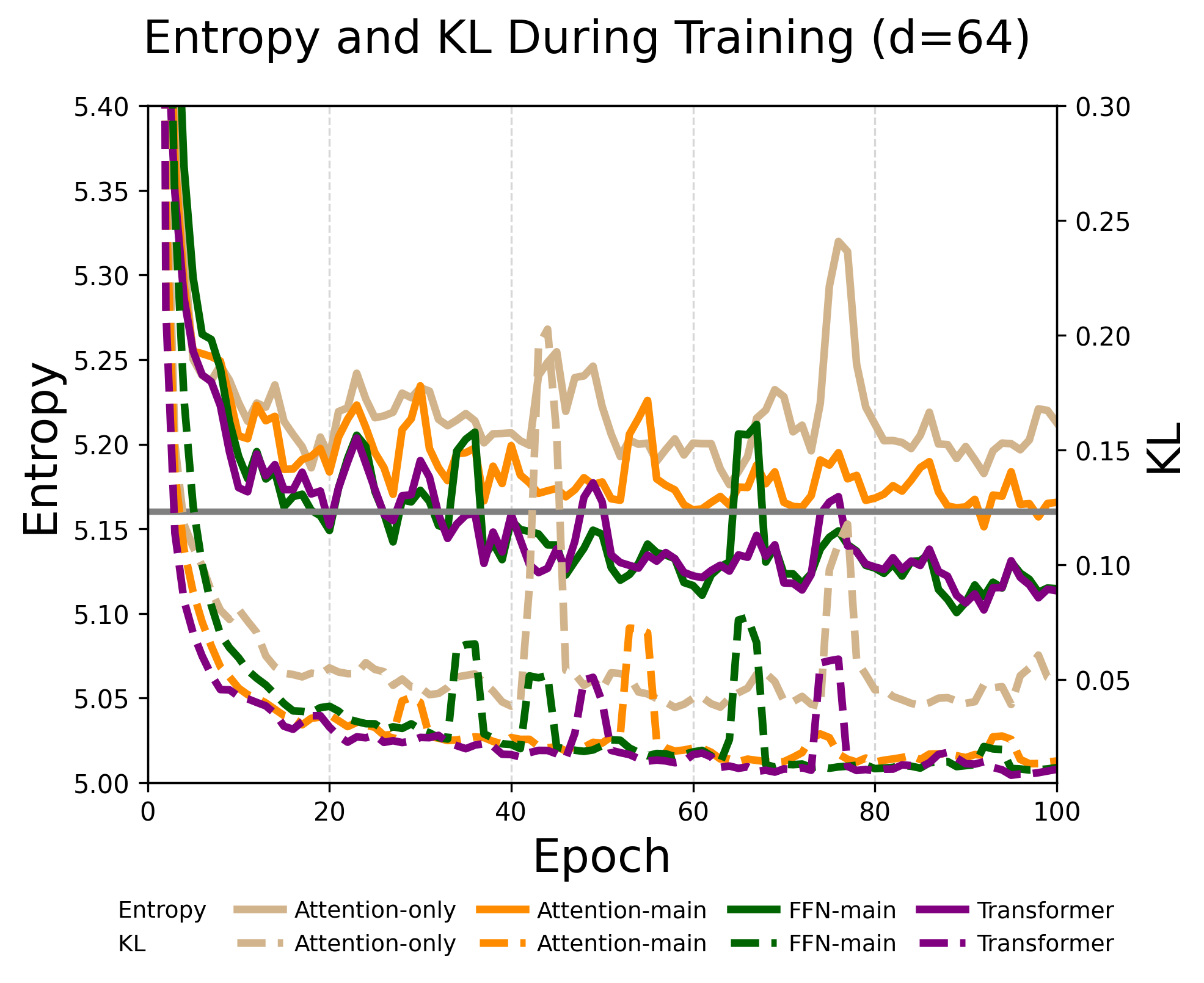}
		%		\subcaption{ }
		%		\label{fig:16-norm}
	\end{subfigure}
	\hspace{0.2cm}
	\begin{subfigure}[t]{0.22\linewidth}
		\centering
		\includegraphics[scale=.205]{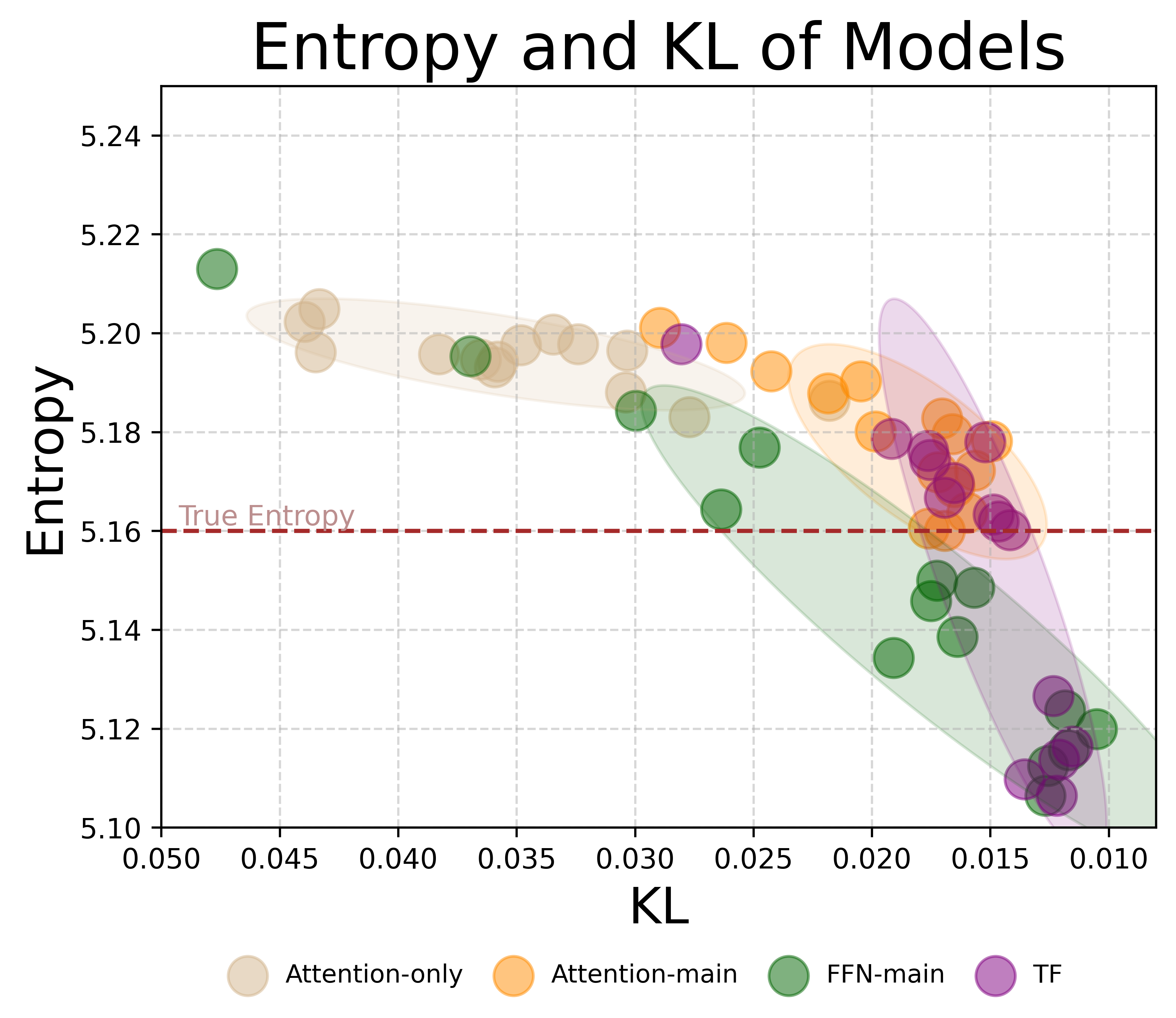}
		%		\subcaption{ }
		%		\label{fig:a}
	\end{subfigure}
	\hspace{0.2cm}
	\begin{subfigure}[t]{0.22\linewidth}
		\centering
		\includegraphics[scale=.205]{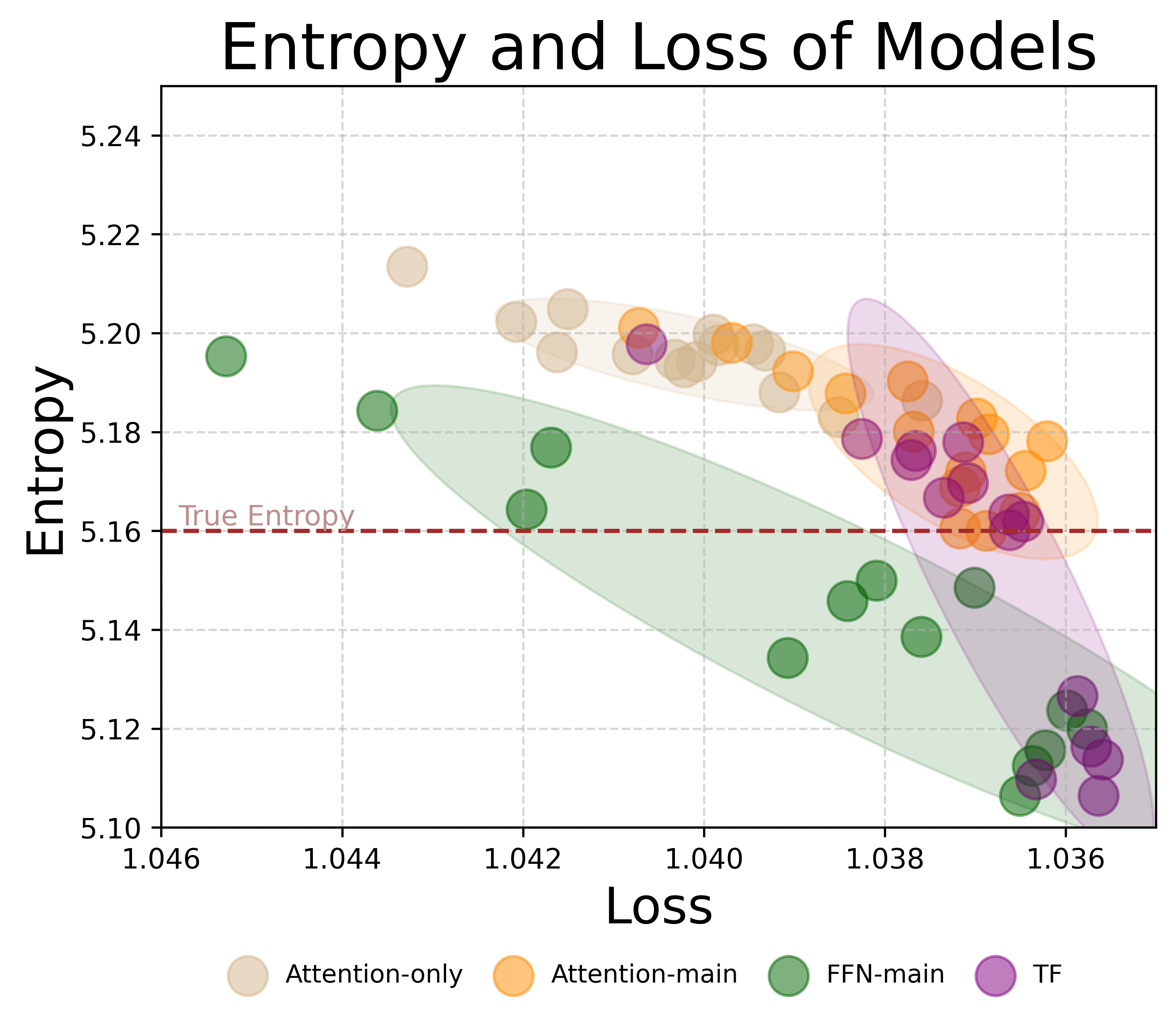}
		%		\subcaption{ }
		%		\label{fig:b}
	\end{subfigure}
	\hspace{0.2cm}
	\begin{subfigure}[t]{0.22\linewidth}
		\centering
		\includegraphics[scale=.205]{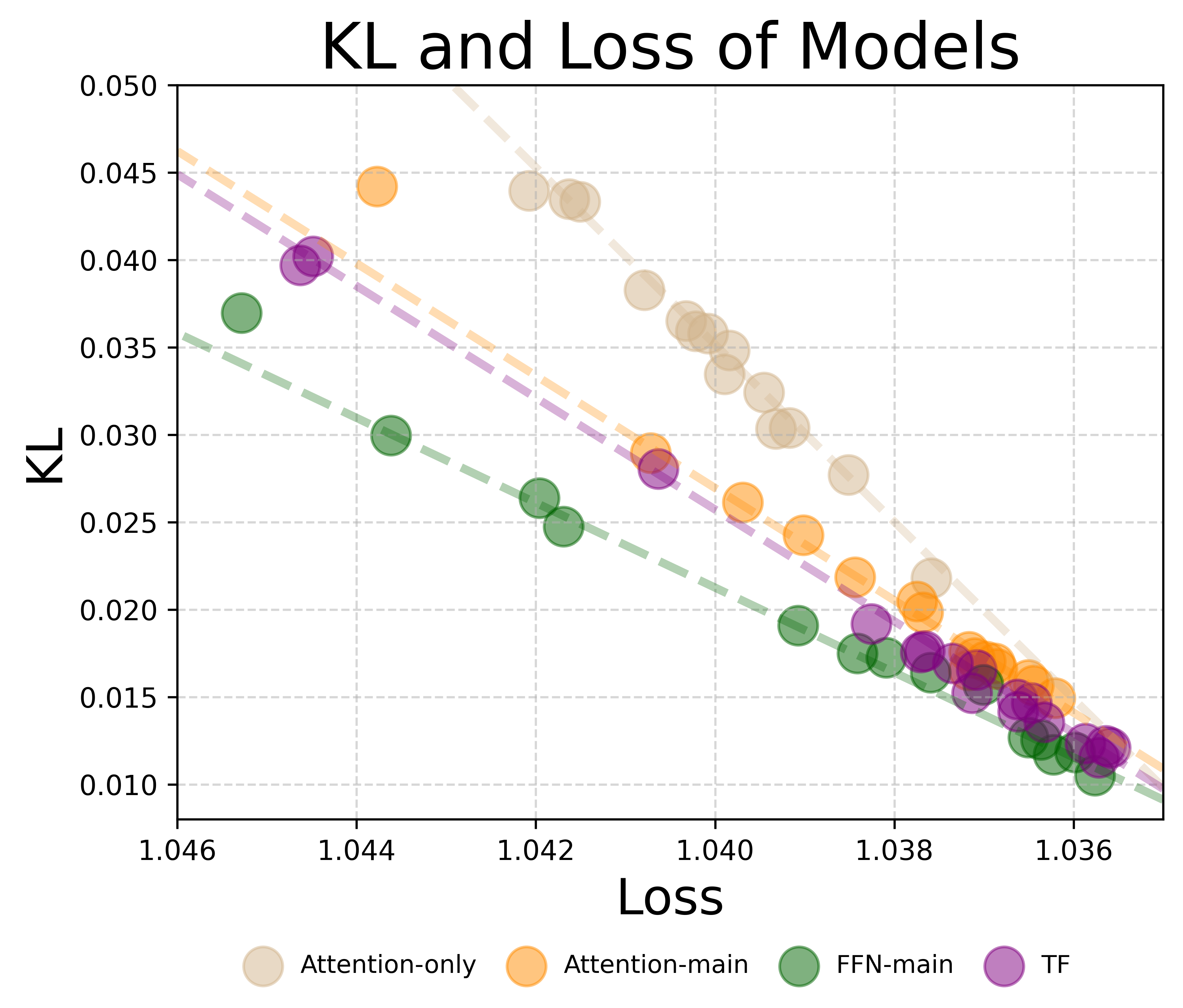}
		%		\subcaption{ }
		%		\label{fig:c}
	\end{subfigure}
	\vspace{-0cm}
	\caption{
		{\bf Left:} Entropy and KL during training for different Transformer variants. 
		{\bf Center Left and Center Right:} Relationship between KL/Loss and Entropy for different variants. FFN-main can achieve lower entropy compared to Attention-only and Attention-main.
		{\bf Right:} Relationship between KL and Loss for different variants.
	}
	\vspace*{-0cm}
\end{figure*}

\begin{figure*}[!htbp]
	\centering
	\begin{subfigure}[t]{0.22\linewidth}
		\centering
		\includegraphics[scale=.21]{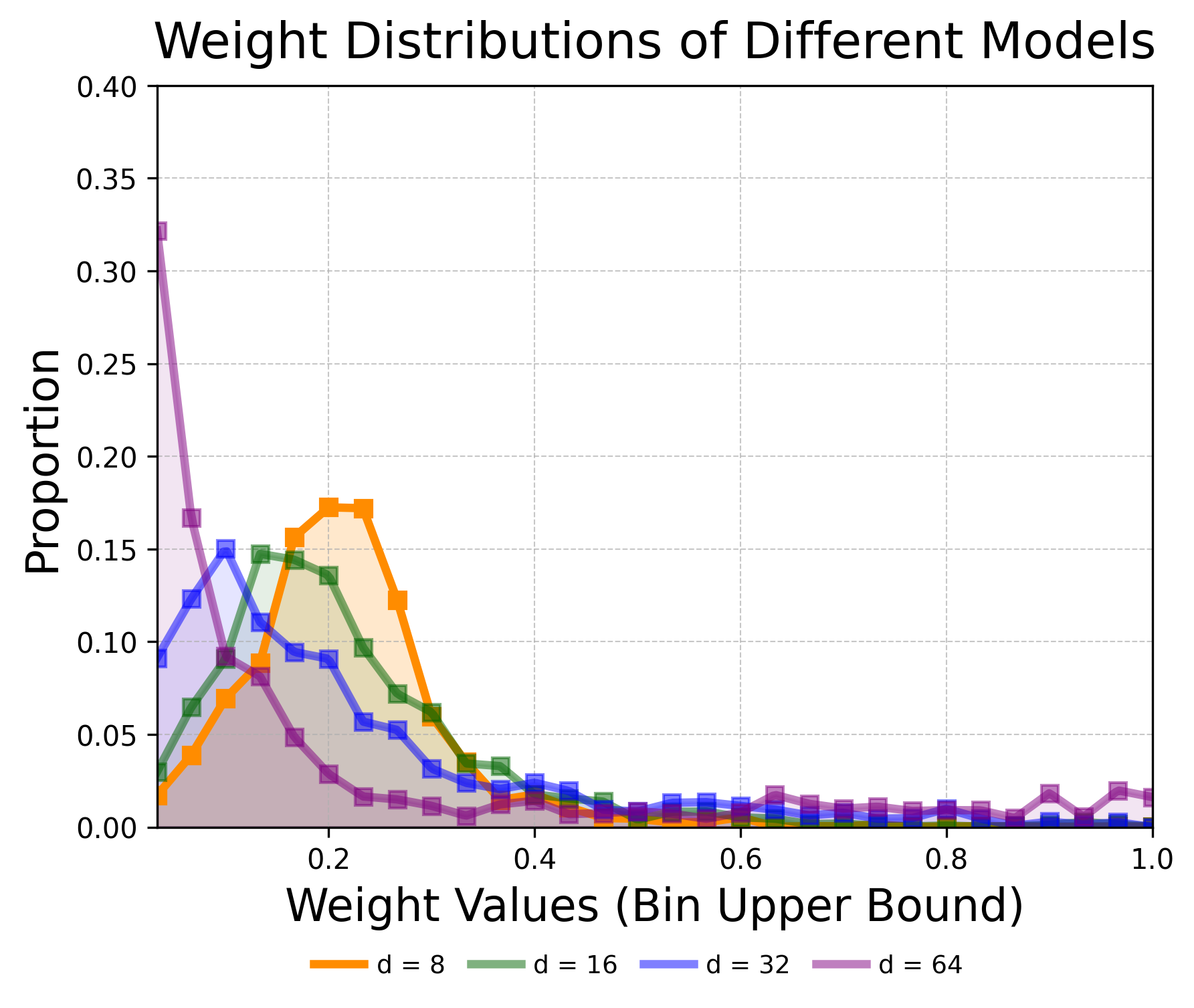}
		%		\subcaption{ }
		%		\label{fig:16-norm}
	\end{subfigure}
	\hspace{0.1cm}
	\begin{subfigure}[t]{0.22\linewidth}
		\centering
		\includegraphics[scale=.21]{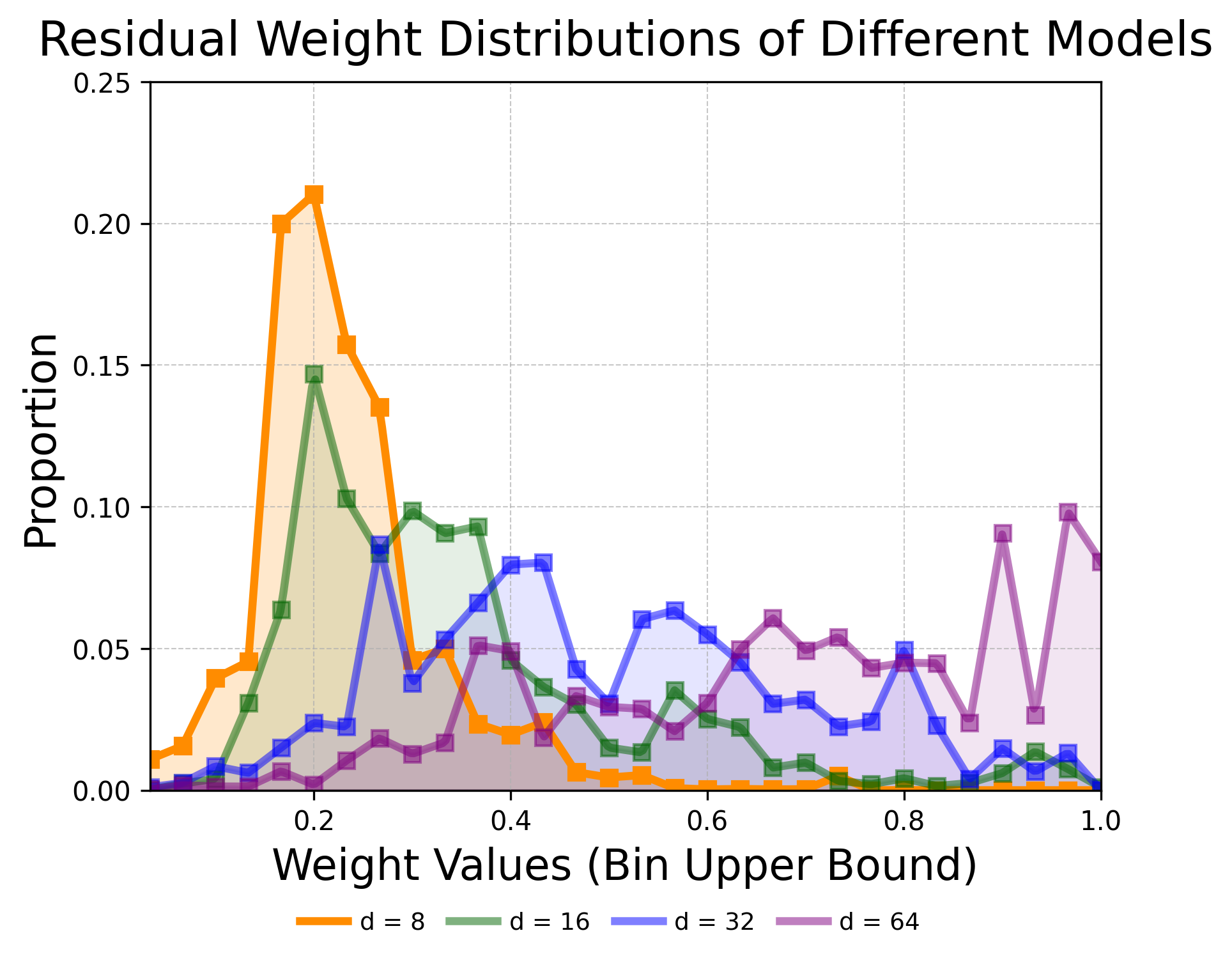}
		%		\subcaption{ }
		%		\label{fig:a}
	\end{subfigure}
	\hspace{0.1cm}
	\begin{subfigure}[t]{0.22\linewidth}
		\centering
		\includegraphics[scale=.21]{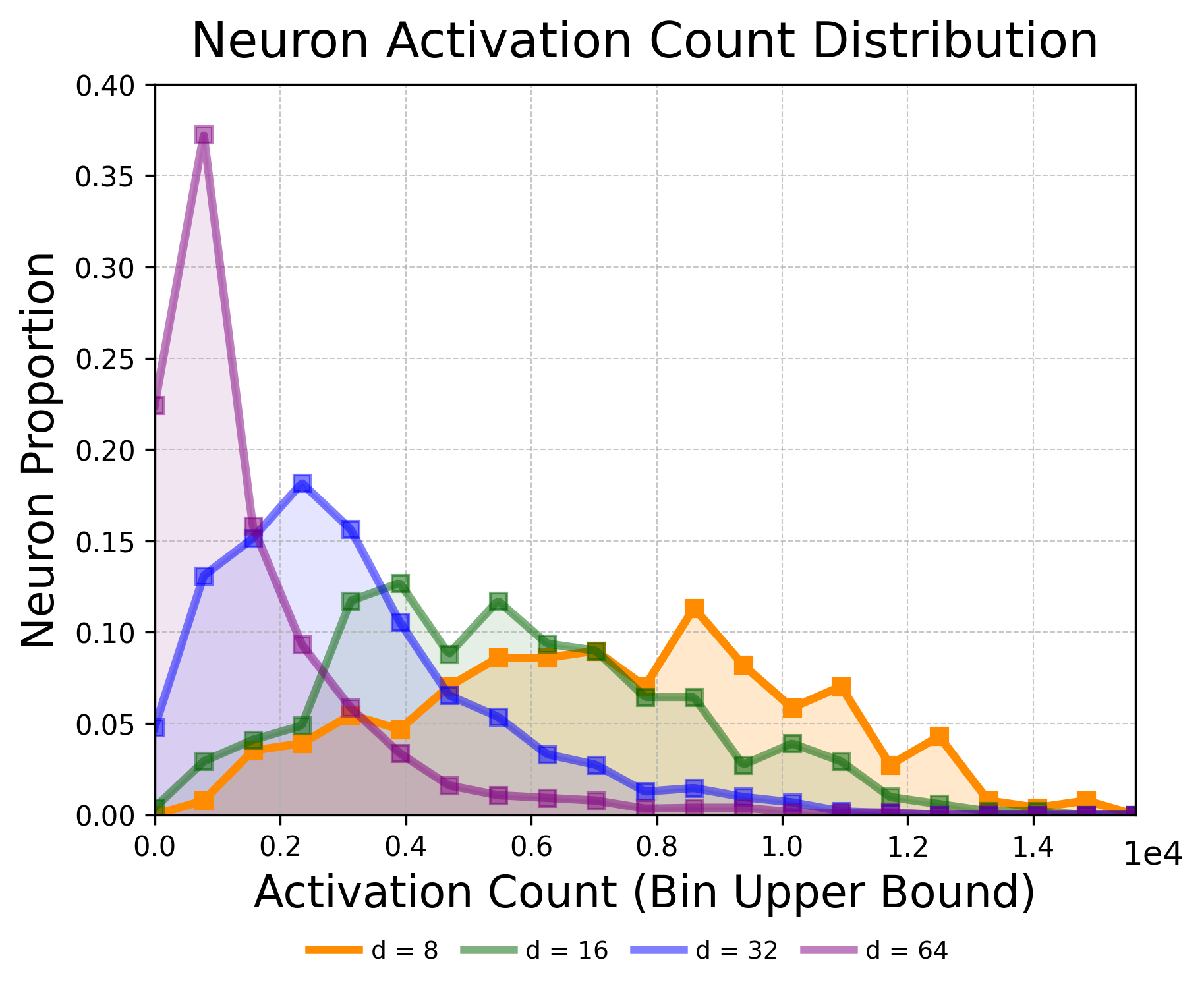}
		%		\subcaption{ }
		%		\label{fig:b}
	\end{subfigure}
	\hspace{0.1cm}
	\begin{subfigure}[t]{0.22\linewidth}
		\centering
		\includegraphics[scale=.21]{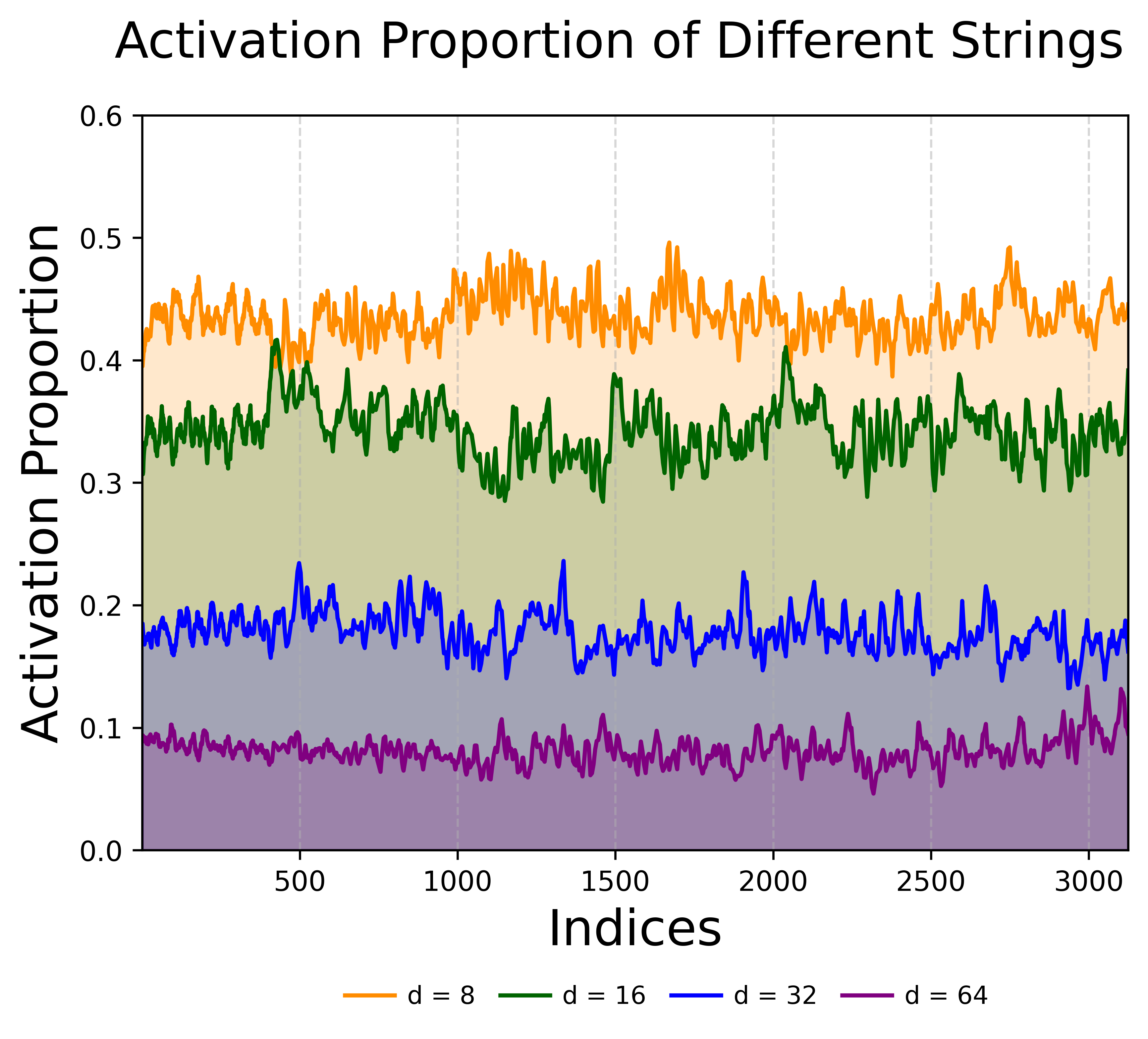}
		%		\subcaption{ }
		%		\label{fig:c}
	\end{subfigure}
	\vspace{-0cm}
	\caption{
		{\bf Left:} The weight distribution of all paths for different model sizes.
		{\bf Center Left:} The weight distribution of only residual paths for different model sizes.
		{\bf Center Right:} The distribution of neuron activation counts for Transformers of different sizes.
		{\bf Right:} The proportion of neurons activated across different input sequences.
	}
	\vspace*{-0cm}
\end{figure*}

\begin{figure*}[!htbp]
	\centering
	\begin{subfigure}[t]{0.18\linewidth}
		\centering
		\includegraphics[scale=.165]{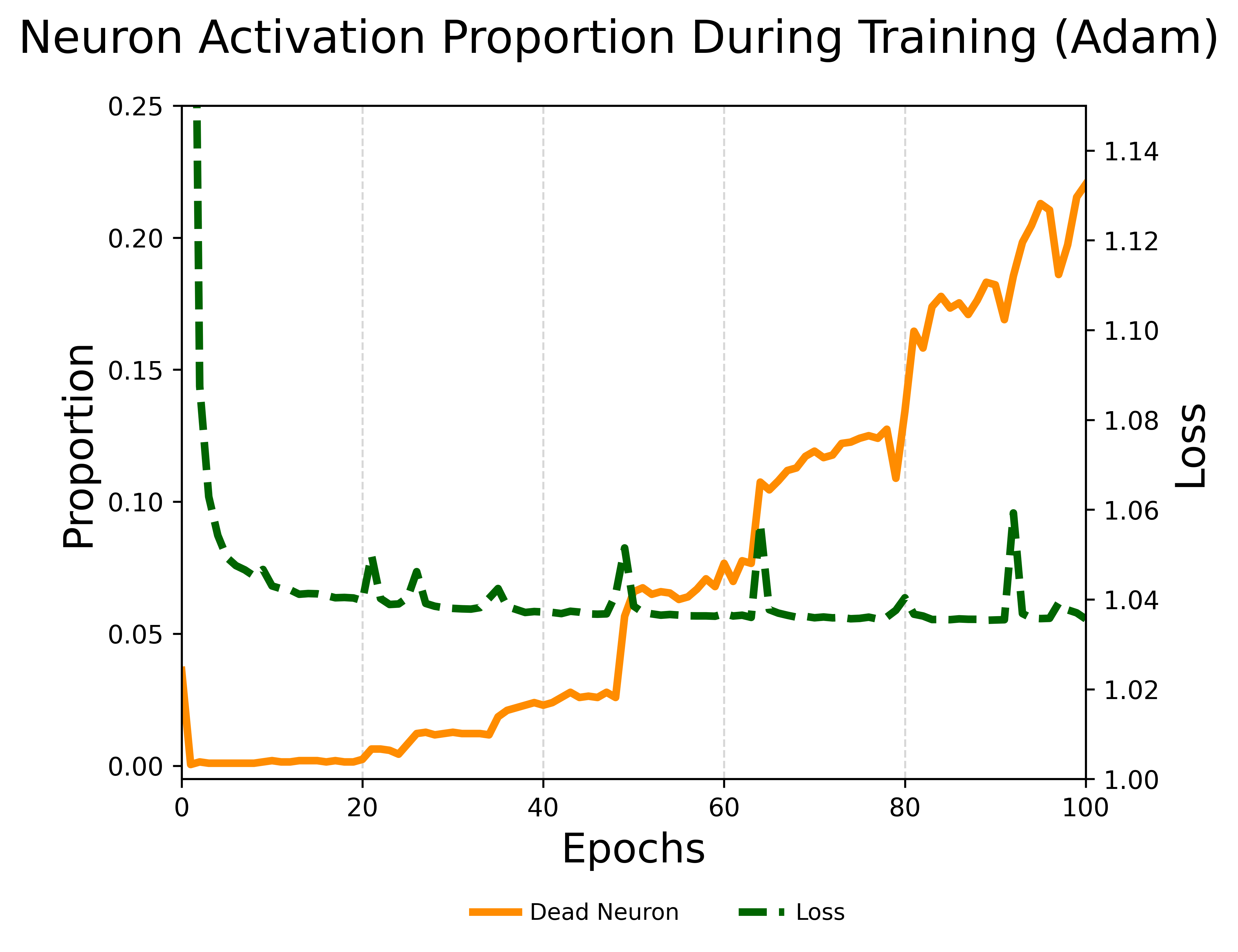}
		%		\subcaption{ }
		%		\label{fig:16-norm}
	\end{subfigure}
	\hspace{0.15cm}
	\begin{subfigure}[t]{0.18\linewidth}
		\centering
		\includegraphics[scale=.165]{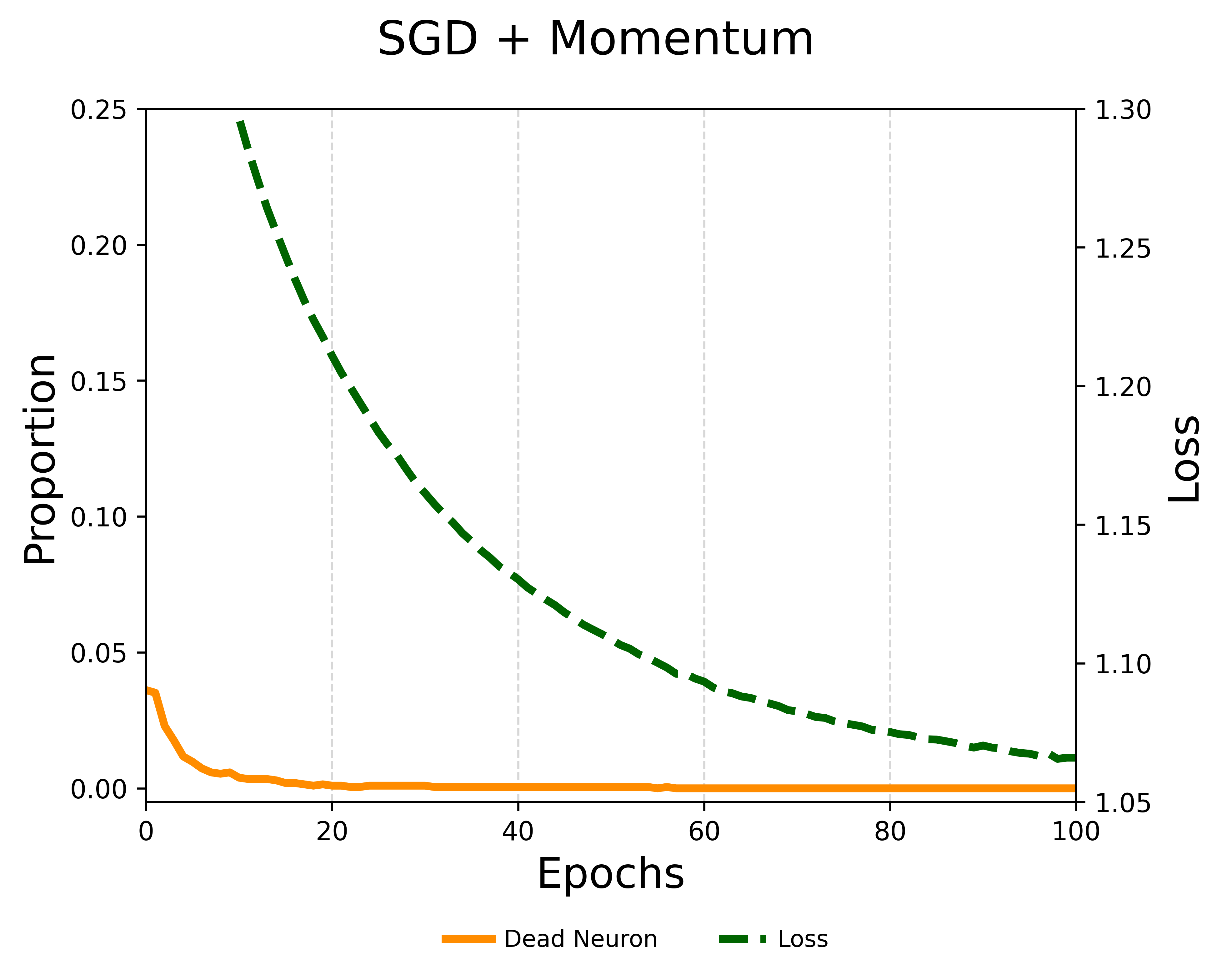}
		%		\subcaption{ }
		%		\label{fig:16-norm}
	\end{subfigure}
	\hspace{0.15cm}
	\begin{subfigure}[t]{0.18\linewidth}
		\centering
		\includegraphics[scale=.165]{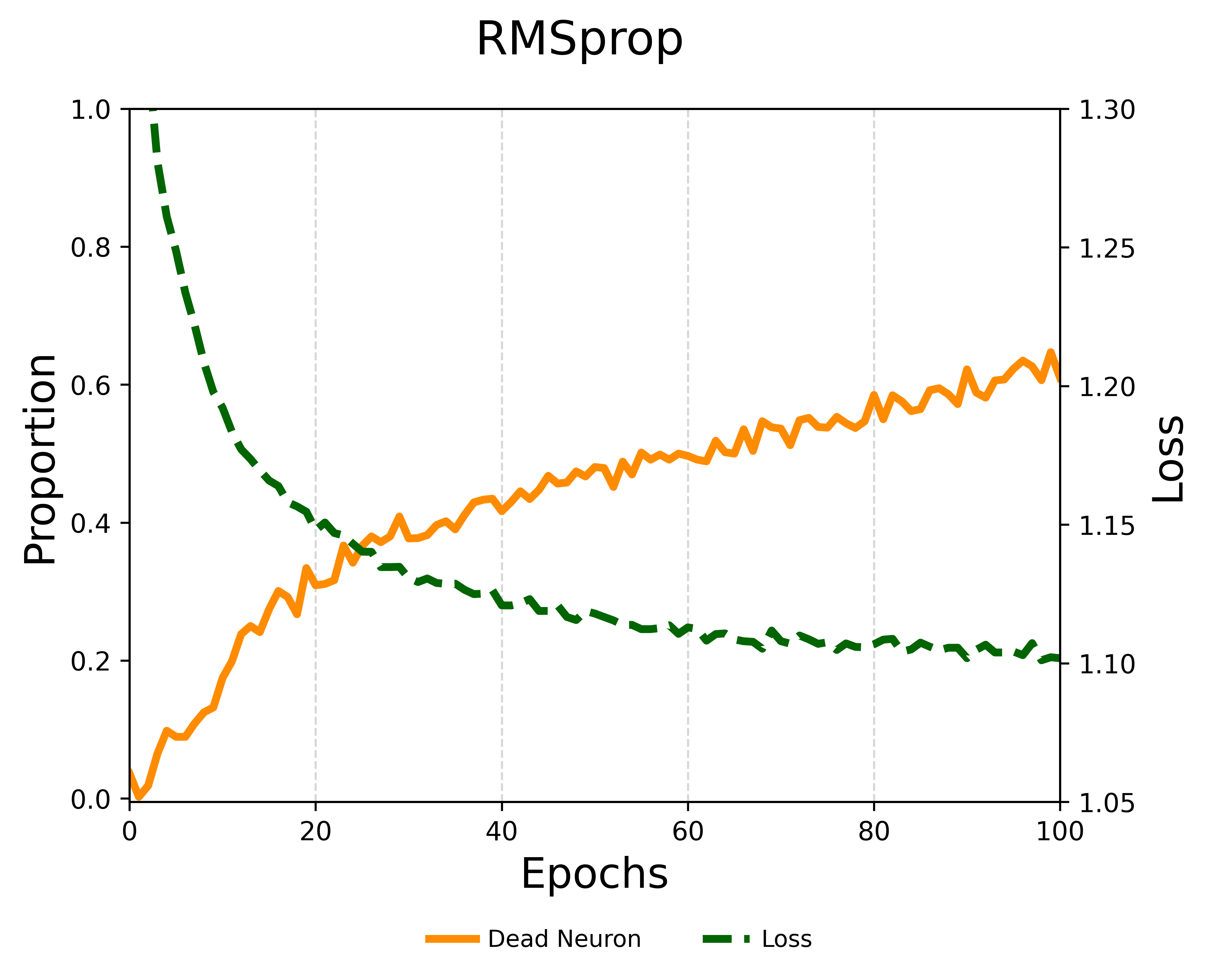}
		%		\subcaption{ }
		%		\label{fig:b}
	\end{subfigure}
	\hspace{0.15cm}
	\begin{subfigure}[t]{0.18\linewidth}
		\centering
		\includegraphics[scale=.165]{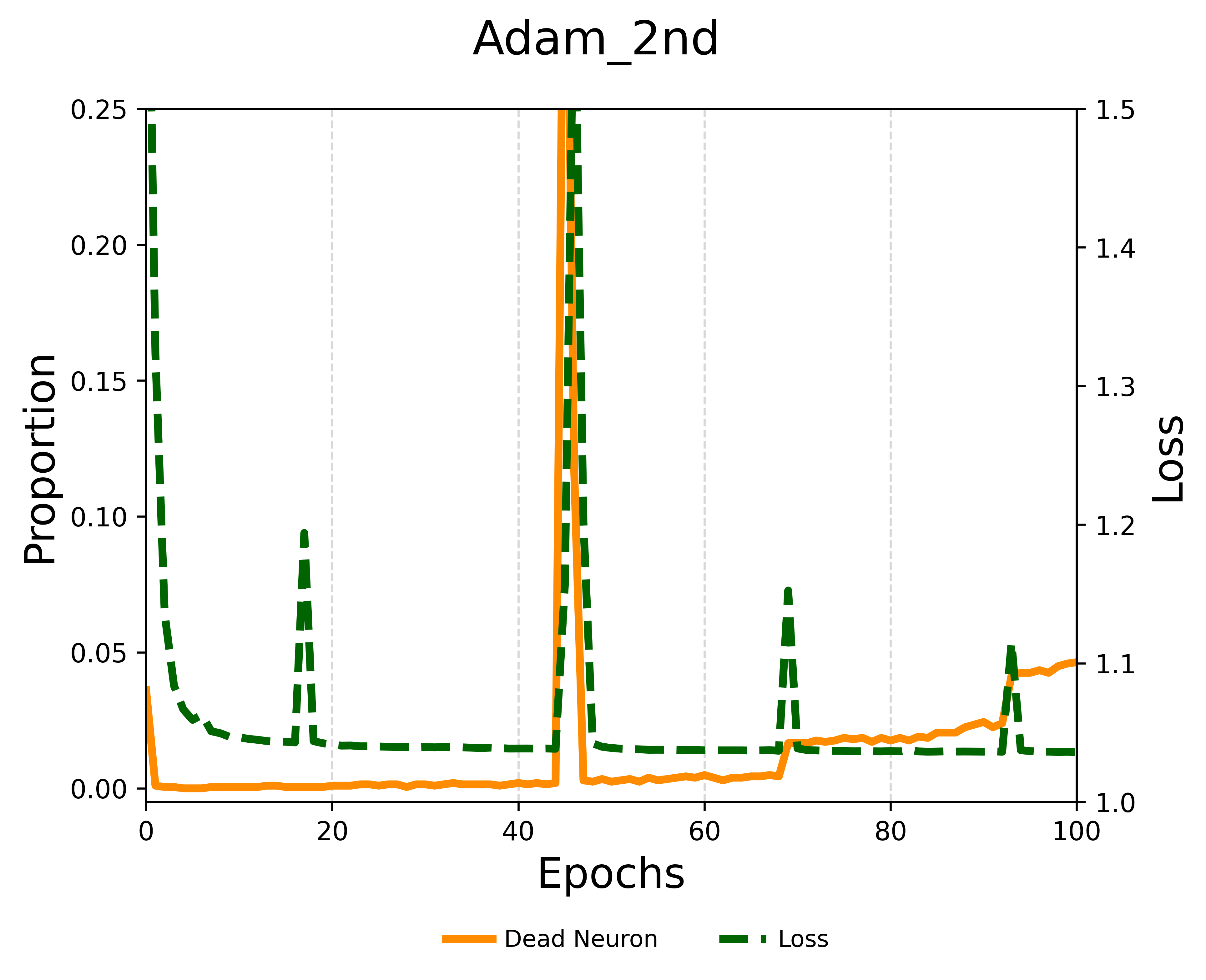}
		%		\subcaption{ }
		%		\label{fig:a}
	\end{subfigure}
	\hspace{0.15cm}
	\begin{subfigure}[t]{0.18\linewidth}
		\centering
		\includegraphics[scale=.165]{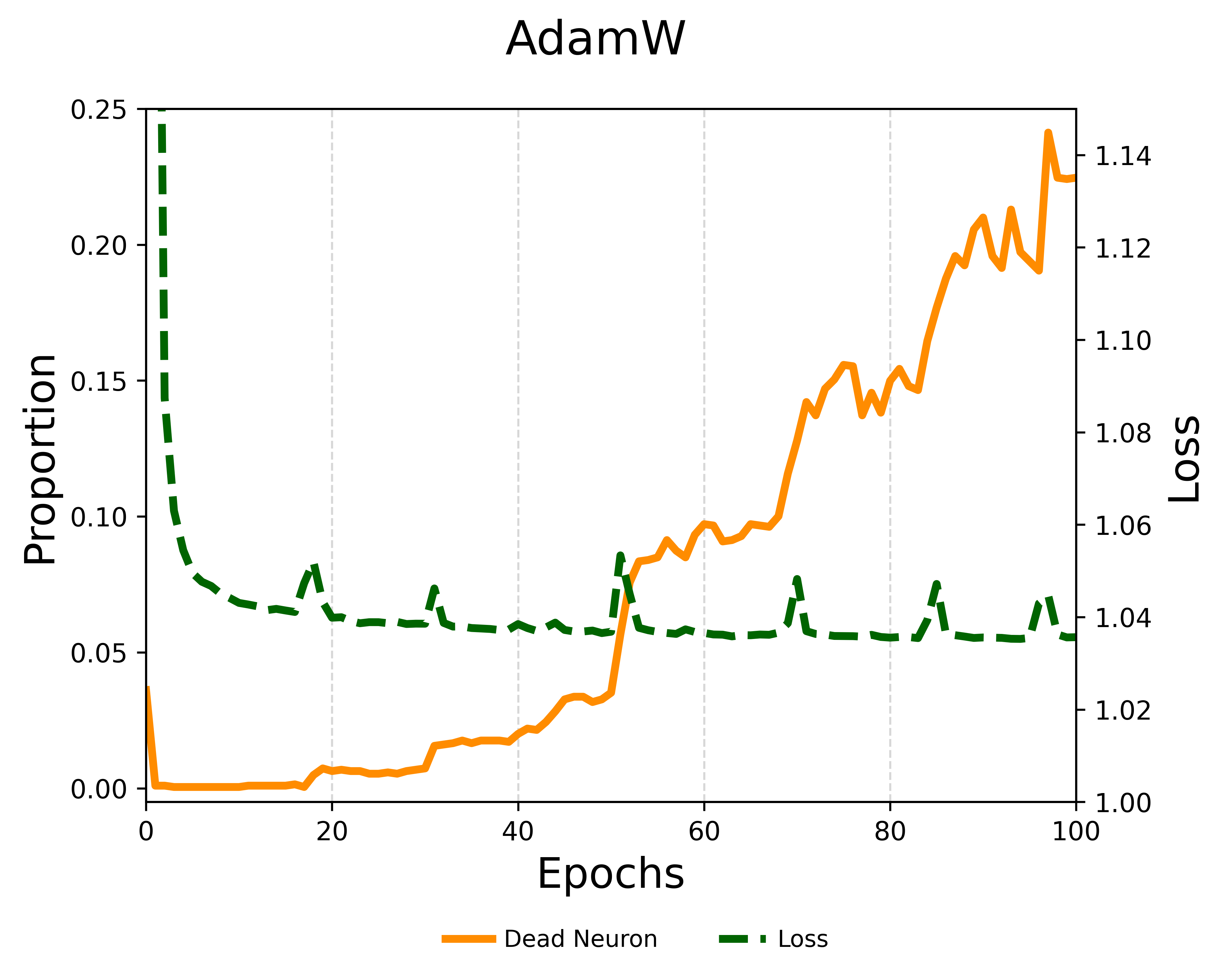}
		%		\subcaption{ }
		%		\label{fig:c}
	\end{subfigure}
	\vspace{-0cm}
	\caption{Loss and the proportion of dead neurons during training under different optimizers. We use {\bf Adam} as the optimizer by default. {\bf SGD with momentum} does not cause loss spikes or an increase in the proportion of dead neurons. {\bf RMSprop} exhibits unstable training and a higher proportion of dead neurons. {\bf Adam primarily relying on second-order gradients} causes more severe loss spikes and greater fluctuations in the dead neurons proportion compared to regular Adam while {\bf AdamW} shows behavior similar to Adam.}
	\vspace*{-0cm}
\end{figure*}

\end{document}